\documentclass[11pt, draftcls, onecolumn]{IEEEtran}
\usepackage{amsmath,amssymb}
\usepackage{bm}
\usepackage{mathtools}
\usepackage{stmaryrd}
\usepackage{graphicx,graphics,color,psfrag}
\usepackage{cite,balance}
\usepackage{subfigure}
\usepackage{booktabs}
\usepackage{caption}   
\allowdisplaybreaks
\usepackage{algorithm}
\usepackage{bm}
\usepackage{array}
\usepackage{url}
\usepackage{scalerel}
\usepackage{tcolorbox}
\usepackage[english]{babel}
\usepackage{algpseudocode}
\usepackage{multirow}
\usepackage{enumerate}
\usepackage{cases}
\usepackage{booktabs}
\newcommand{\mb}[1]{\mathbf{#1}}
\newcommand{\bc}[1]{\mbox{\boldmath $\mathcal{#1}$}}
\usepackage{multirow}
\usepackage{placeins}
\usepackage{hhline}

\usepackage{amsmath}
\usepackage{amssymb}

\usepackage{url}
\usepackage{multirow}

\makeatletter

\let\SF@@footnote\footnote
\def\footnote{\ifx\protect\@typeset@protect
    \expandafter\SF@@footnote
  \else
    \expandafter\SF@gobble@opt
  \fi
}
\expandafter\def\csname SF@gobble@opt \endcsname{\@ifnextchar[
  \SF@gobble@twobracket
  \@gobble
}
\edef\SF@gobble@opt{\noexpand\protect
  \expandafter\noexpand\csname SF@gobble@opt \endcsname}
\def\SF@gobble@twobracket[#1]#2{}

\usepackage{tikz}
\usetikzlibrary{shapes.misc, positioning}
\usetikzlibrary{backgrounds,fit,matrix, calc}
\tikzset{
   box/.style = {minimum height=10pt, minimum width=10pt, draw, rounded corners,rectangle, fill=white!50},
}
\usetikzlibrary{fadings}

\tikzset{
   boxconv/.style = {minimum height=2cm, minimum width=2cm, draw, line width=0.4mm, fill opacity=0.9, rounded corners,rectangle, fill=white!50},
}

\tikzset{
   boxconv_inactive/.style = {minimum height=2cm, minimum width=2cm,line width=0.3mm, draw, line width=0.1mm , fill opacity=0.9, rounded corners,rectangle, gray, fill=white!50},
}
\tikzset{
   input/.style = {minimum height=3cm, minimum width=3cm, draw, , fill opacity=0.9, rectangle, fill=white!50},
}

\tikzset{
   boxpooled/.style = {minimum height=1.5cm, minimum width=1.5cm, draw, line width=0.4mm, fill opacity=0.9, rounded corners,rectangle, fill=white!50},
}

\tikzset{
   boxpooled_inactive/.style = {minimum height=1.5cm, minimum width=1.5cm, draw, line width=0.4mm, fill opacity=0.9, rounded corners,rectangle, gray, fill=white!50},
}

\usetikzlibrary{fadings}\tikzset{
    boxwta/.style={%
        draw=black, thick,
        rectangle,
        rounded corners,
        minimum height=3cm,
        minimum width=3cm
    }
}

\usetikzlibrary{fadings}\tikzset{
    box1/.style={%
        draw=black, thick,
        rectangle,
        minimum height=2cm,
        minimum width=2cm
    }
}

\usetikzlibrary{fadings}\tikzset{
    box2/.style={%
        draw=black, thick,
        rectangle,
        minimum height=1.cm,
        minimum width=1.cm
    }
}

\usetikzlibrary{fadings}\tikzset{
    box3/.style={%
        draw=black, thick,
        rectangle,
        minimum height=.8cm,
        minimum width=.8cm
    }
}
\usetikzlibrary{decorations.markings}

\usepackage{xcolor} 

\begin{document}

\title{Rethinking Bayesian Learning for Data Analysis: The Art of Prior and Inference in Sparsity-Aware Modeling}

\author{Lei Cheng$^\sharp$, Feng Yin$^\S$\footnote{Lei Cheng and Feng Yin contribute equally.}, Sergios Theodoridis$^\star$, Sotirios Chatzis$^\dagger$, and Tsung-Hui Chang$^\S$\\ 
\vspace{0.3 cm}
{\small $^\sharp$ College of Information Science and Electronic Engineering, Zhejiang University, Hangzhou, China}\\
{\small $^\S$  School of Science and Engineering, The Chinese University of Hong Kong, Shenzhen, China}\\
{\small $^\star$ Department of Informatics and Telecommunications,  National and Kapodistrian University of Athens, Greece \\ and Department of Electronics Systems, Aalborg University, Denmark} \\
{\small $^\dagger$ Department of Electrical $\&$ Computer
Engineering and Informatics, Cyprus University of Technology, Cyprus}
{\small Correspondence Author: Feng Yin, yinfeng@cuhk.edu.cn}}

\maketitle

\begin{abstract}
Sparse modeling for signal processing and machine learning, in general, has been at the focus of scientific research for over two decades. Among others, supervised sparsity-aware learning comprises two major paths paved by: a) discriminative methods that establish direct input-output mapping based on a regularized cost function optimization, and b) generative methods that learn the underlying distributions. The latter, more widely known as Bayesian methods, enable uncertainty evaluation with respect to the performed predictions. Furthermore, they can better exploit related prior information and also, in principle, can naturally introduce robustness into the model, due to their unique capacity to marginalize out uncertainties related to the parameter estimates. Moreover, hyper-parameters (tuning parameters) associated with the adopted priors, which correspond to cost function regularizers, can be learnt via the training data and  not via costly cross-validation techniques, which is, in general, the case with the discriminative methods. To implement sparsity-aware learning, the crucial point lies in the choice of the function regularizer for discriminative methods and the choice of the prior distribution for Bayesian learning. Over the last decade or so, due to the intense research on deep learning, emphasis has been put on discriminative techniques. However, a come back of Bayesian methods is taking place that sheds new light on the design of deep neural networks, which also establish firm links with Bayesian models, such as Gaussian processes, and, also, inspire new paths for unsupervised learning, such as Bayesian tensor decomposition.

The goal of this article is two-fold. First, to review, in a unified way, some recent advances in incorporating sparsity-promoting priors into three highly popular data modeling/analysis tools, namely deep neural networks, Gaussian processes, and tensor decomposition. Second, to review their associated inference techniques from different aspects, including: evidence maximization via optimization and variational inference methods. Challenges such as small data dilemma, automatic model structure search, and natural prediction uncertainty evaluation are also discussed. Typical signal processing and machine learning tasks are considered, such as time series prediction, adversarial learning, social group clustering, and image completion. Simulation results corroborate the effectiveness of the Bayesian path in addressing the aforementioned challenges and its outstanding capability of matching data patterns automatically.
\end{abstract}

\section{Introduction}
\label{sec:introduction}
Over the past three decades or so, machine learning has been gradually established as the umbrella name to cover methods whose goal is to extract valuable information and knowledge from data, and then use it to make predictions \cite{theodoridis2020machine}. Machine learning  has been extensively applied to a wide range of disciplines, such as signal processing, data mining, communications, finance, bio-medicine, robotics, to name but a few. The majority of the machine learning  methods first rely on adopting a parametric model to describe the data at hand, and then an inference/estimation technique to derive estimates that describe the unknown model parameters. In the discriminative methods, point estimates of the involved parameters are obtained via cost function optimization. In contrast, by practicing the Bayesian philosophy, one can infer the underlying statistical distributions that describe the unknown parameters given the observed data,  and, thus, provide a generative mechanism that models the random process that generates the data.

For the newcomers to machine learning, the discriminative (also referred to as cost function optimization) perspective might be more straightforward. It first formulates a task that quantifies the overall deviation between the observed target data and the model predictions, and then solves it for the point parameter estimates  via an optimization algorithm. On the contrary, the generative (Bayesian) perspective, which aims to reveal the generative process and the statistical properties of the observed data, sounds more complicated due to some ``jargon" terms  such as prior, likelihood, posterior, and evidence. Nevertheless, machine learning  under the Bayesian perspective is gaining in popularity recently due to the \emph{comparative advantages} that spring from the nature of the statistical modeling and the extra information returned by the posterior distributions. This article aims at demystifying the philosophy that underlies the Bayesian techniques, and then review, in a unified way, recent advances of \emph{Bayesian sparsity-aware learning} for three analysis tools of high current interest. In the Bayesian framework, model sparsity is implemented via sparsity-promoting priors that lead to automatic model determination by optimally sparsifying an, originally, over-parameterized model. The goal is  to optimally  predict the order of the system that corresponds to the best trade-off between accuracy and complexity, with the aim to combat overfitting, in line with the general concept of regularization. However, in Bayesian learning, all the associated  (hyper-)parameters, which control the degree of regularization,  can be optimally obtained via the training set during the learning phase. It is hoped that this article can help the newcomers grasp the essence of Bayesian learning, and at the same time, provide the experts with an update of some recent advances developed for different data modeling and analysis tasks.  

In particular, we will focus on Bayesian sparsity-aware learning for three popular data modeling and analysis tools, namely the deep neural networks (DNNs), Gaussian processes (GPs), and tensor decomposition, that have promoted intelligent signal processing applications. Some typical examples are as follows. 

In the supervised learning front with over-parameterized DNNs, novel data-driven mechanisms have been proposed in \cite{panousis2019nonparametric,  panousis21a, panousis2022stochastic, louizos2017bayesian, ghosh2019model} to intelligently prune redundant neuron connections without human assistance. In a similar vein, in \cite{yin2020linear, paananen2019variable,kim2018scaling}, sparsity-promoting priors have been used in the context of the GPs that give rise to optimal and interpretable kernels that are capable of identifying a sparse subset of effective frequency components automatically. In the unsupervised learning front, some advanced works on tensor decomposition, e.g., \cite{cheng2022towards, cheng2020learning, zhou2019bayesian, cheng2016probabilistic, zhao2015bayesian, zhang2018variational}, have shown that sparsity-promoting priors are able to unravel the few underlying interpretable components in a completely tuning-free fashion. Such techniques have found various signal processing  applications, including data classification \cite{panousis2019nonparametric, louizos2017bayesian, ghosh2019model}, adversarial learning \cite{panousis21a,panousis2022stochastic}, time-series prediction \cite{yin2020linear,dai2020interpretable,paananen2019variable,kim2018scaling},  blind source separation \cite{cheng2016probabilistic, cheng2018scaling, cheng2022towards}, image completion \cite{zhou2019bayesian, zhao2015bayesian, zhang2018variational},  and wireless communications \cite{cheng2021towards}.
\begin{figure}[!t]
\centering
\includegraphics[width= 5.5 in]{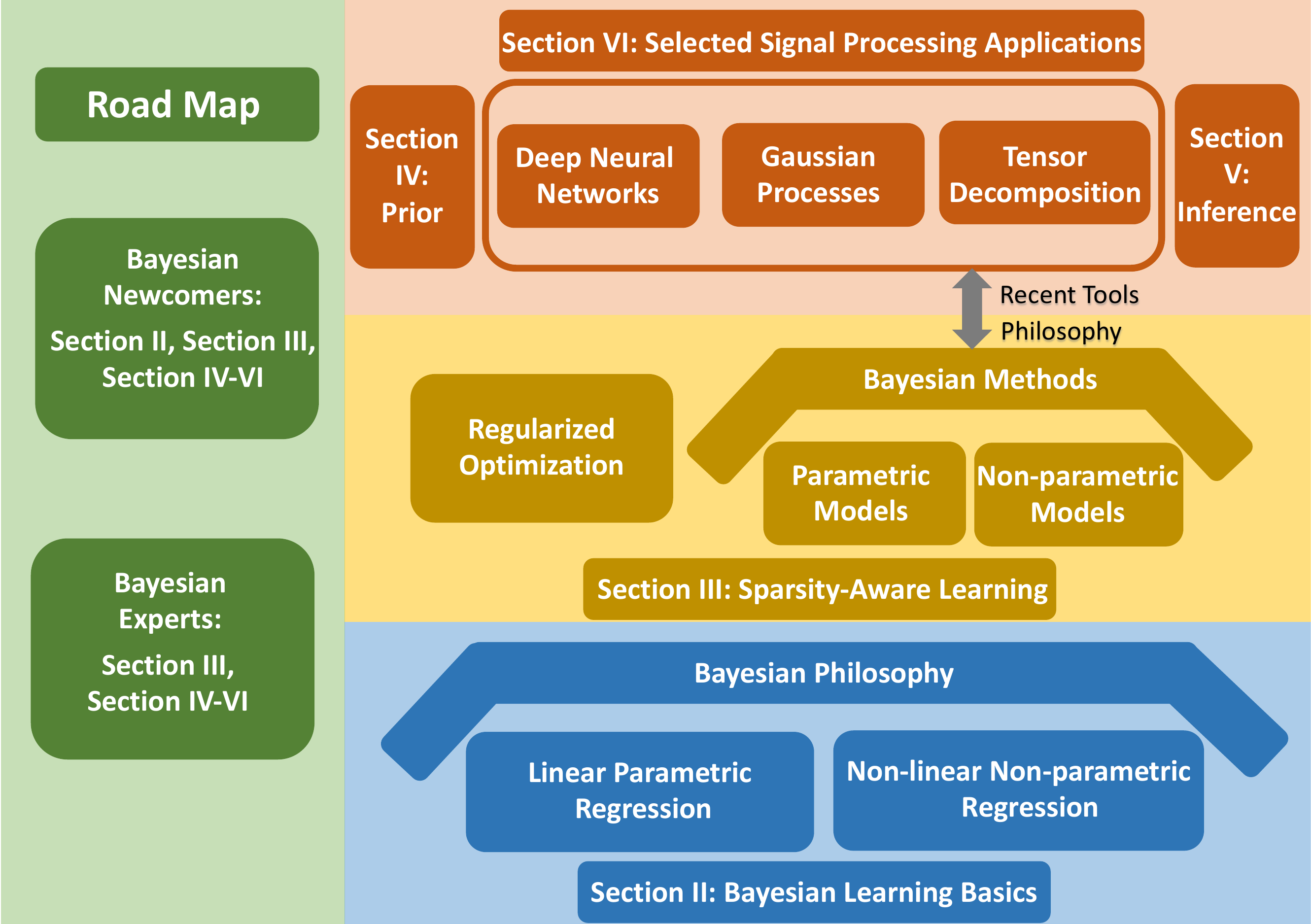}
\caption{The organization of this article and road map for the readers.}
\label{map}
\end{figure}

The aforementioned references address two state-of-the-art challenges: \emph{1) The art of prior}: how should the fundamental sparsity-promoting priors be chosen and tailored to fit modern data-driven models with complex structures? 2) \emph{The art of inference}: how can recent optimization theory and stochastic approximation techniques be leveraged to design fast, accurate, and scalable inference algorithms? This tutorial-style article aims to give a unified treatment on the underlying common ideas and techniques to offer concrete answers to the above questions. It is yet the goal of this article to provide a comprehensive review of such sparsity-promoting techniques. On the one hand, we will introduce some newly proposed sparsity-promoting priors, as well as various salient ones that, although being powerful, had never been used before in our target models. On the other hand, we will showcase some recent developments of the associated inference algorithms. For readers with different backgrounds and familiarity with Bayesian statistics, we provide a roadmap in Fig.~\ref{map} to facilitate their reading. 

The remaining sections of this article are organized as follows. In Section \ref{sec:BayesianLearningBasics}, we introduce some Bayesian learning basics, aiming to let the readers easily follow the main concepts, jargon terms, and math notations. In Section \ref{sec:sal}, we first review two different paths (the regularized optimization and Bayesian paths), and further introduce some sparsity-promoting priors along the Bayesian path. In Section \ref{sec:prior-with-recent-tools}, we demonstrate how to integrate the introduced sparsity-promoting priors into three prevailing data analysis tools, i.e., the DNNs, GPs, and tensor decomposition. For the reviewed sparsity-aware learning models, we further introduce their associated inference methods in Section \ref{sec:inference}. Various signal processing applications of high current interests enabled by the aforementioned models are exemplified in Section \ref{sec:application}. Finally, we conclude the article and bring up some potential future research directions in Section \ref{sec:conclusion}.

\section{Bayesian Learning Basics}
\label{sec:BayesianLearningBasics}
In this section, we first provide some touches on the philosophy of Bayesian learning in Section~\ref{subsec:BayesianPhi}, and use \emph{Bayesian linear regression} as an example to elucidate different symbol notations, terminology and unique features of Bayesian learning in Section~\ref{subsec:blr-exmaple}. Then, we discuss extensions to the non-linear and non-parametric cases, shedding light on the connections between simple linear regression and advanced Gaussian process regression in Section~\ref{subsec:gp}. 

\subsection{Bayesian Philosophy Basics}
\label{subsec:BayesianPhi}
\subsubsection{Bayes' Theorem} Let $\mathcal D$ be the observed (training) dataset and $\mathcal M$ be the underlying  model that is assumed to generate the data. For simplicity, we start our treatment with models that are parameterized in terms of a set of unknown parameters $\boldsymbol \theta \in \mathbb{R}^{L \times 1}$, where $\mathbb{R}$ is the set of real numbers.  By the definition of parametric models, the dimension $L$ is pre-selected and fixed \cite{theodoridis2020machine}. According to the Bayesian philosophy, these parameters are treated  as random variables. Their randomness does not imply a random nature of these parameters, but essentially encodes our uncertainty with respect to their true (yet unknown) values, see related discussions in, e.g.,  \cite{theodoridis2020machine}.  First, in Bayesian modeling, we assume that the  set of unknown random parameters is described by a prior distribution, i.e, $\boldsymbol \theta \sim p_{ \mathcal M}(\boldsymbol \theta; \boldsymbol{\eta}_{p})$, which encodes our \emph{prior} belief in $\boldsymbol \theta$; that is, it encodes our uncertainty prior to receiving the dataset $\mathcal{D}$. As we are going to see soon, this corresponds to regularizing the learning task, since it will bias the solution that we seek towards certain regions in the parameter space. The prior $p_{ \mathcal M}(\boldsymbol \theta; \boldsymbol{\eta}_{p})$ is specified via a set of deterministic yet unknown \emph{hyper-parameters} (tuning parameters) stacked together in a vector and denoted by $\boldsymbol{\eta}_{p}$. The second quantity that is assumed to be known is the conditional distribution that describes the data given the values of the parameters,  $\boldsymbol \theta$, which for the specific observed dataset $\mathcal{D}$ is known as the {\it likelihood} $p_{\mathcal M}(\mathcal D | \boldsymbol \theta)$.\footnote{Throughout this paper, we use ``$;$'', i.e., $p(x;\eta)$ if $\eta$ is a deterministic parameter to be optimized or pre-selected by the user; and we use ``$|$", i.e., $p(x | \eta)$ if $\eta$ is a random variable or a hyper-parameter treated as a random variable; that is, if the distribution is conditional on another random variable.}

Having selected the likelihood and the prior distribution function, the goal of Bayesian inference is to infer (estimate) the \emph{posterior} distribution of the parameters given the observations, i.e., $p_{\mathcal M}(\boldsymbol \theta | \mathcal D;  \boldsymbol \eta)$, that comprises the update of the prior assumption encoded in $p_{ \mathcal M}(\boldsymbol \theta; \boldsymbol \eta_{p})$ after digesting the dataset $\mathcal D$. This process can be elegantly described by the celebrated \emph{Bayes' theorem}, e.g., the one given in \cite{theodoridis2020machine}:
\begin{align}
p_{\mathcal M}(\boldsymbol \theta | \mathcal D;  \boldsymbol \eta) = \frac{p_{\mathcal M}(\mathcal D | \boldsymbol \theta) p_{\mathcal M}(\boldsymbol \theta;  \boldsymbol{\eta}_{p})}{p_{\mathcal M}(\mathcal D; \boldsymbol \eta)}.
\label{eq1}
\end{align}
Note that $\boldsymbol \eta$ includes both the hyper-parameters associated with the prior, $\boldsymbol{\eta}_{p}$, and some extra hyper-parameters involved in the likelihood function $p_{\mathcal M}(\mathcal D | \boldsymbol \theta)$, which are omitted for notation brevity. 

The Bayes' theorem solves for the {\it inverse problem} that is associated with any machine learning  task. The {\it forward problem} is an easy one. Given the model $\mathcal M$ and the values of the associated parameters $\boldsymbol{\theta}$, one can easily generate the output observations $\mathcal D$ from the conditional distribution $p_{\mathcal M}(\mathcal D | \boldsymbol \theta)$.  The task of machine learning is the opposite and a more difficult one. Given the observed data $\mathcal D$, the task is  to estimate/infer the model $\mathcal M$. This is known as the {\it inverse problem}, and Bayes' theorem applied to the machine learning  task does exactly that. It relates the inverse problem (posterior) to the forward one (likelihood). All one needs for this ``update" is to assume a prior and also to obtain an estimate of the distribution associated with the data, which comprises the denominator in \eqref{eq1}. The latter term and the related information are neglected in the discriminative models, hence important information is not taken into account, see the discussion in, e.g., \cite{theodoridis2020machine,Bishop2006}.

Occasionally, we may need a point estimate of the model parameters as the intermediate result, and there are two commonly used estimates that can be computed from the posterior distribution, $p_{\mathcal M}(\boldsymbol \theta | \mathcal D;  \boldsymbol \eta)$. Assuming that $\boldsymbol{\theta}$ is known or an estimate  is available, the first one is known as the maximum-\textit{a-posteriori} (MAP) estimate and the other one as the minimum-mean-squared-error (MMSE) estimate, concretely \cite{theodoridis2020machine}, 
\begin{align}
\hat{\boldsymbol{\theta}}_{\textrm{MAP}} &= \arg \max_{\boldsymbol{\theta}} \,\, p_{\mathcal M}(\boldsymbol \theta | \mathcal D;  \boldsymbol \eta), \\
\hat{\boldsymbol{\theta}}_{\textrm{MMSE}} &= \int \boldsymbol{\theta} \cdot p_{\mathcal M}(\boldsymbol \theta | \mathcal D;  \boldsymbol \eta) \,  d \boldsymbol{\theta}. 
\end{align}

\subsubsection{Evidence Maximization for Hyper-parameter Learning} In the prior distribution, $p_{\mathcal M}(\boldsymbol \theta;  \boldsymbol{\eta}_{p})$, the hyper-parameters, $\boldsymbol{\eta}_{p}$, could be either pre-selected according to the side information at hand, or learnt from the observed dataset $\mathcal D$. In Bayesian learning, the latter path is followed favorably. One popular alternative is to select the full set of hyper-parameters $\boldsymbol \eta$ to be the most compatible with the observed dataset $\mathcal D$, which can be naturally formulated as the following so-called {\it evidence maximization}:
\begin{align}
\max_{\boldsymbol \eta}  \, \log p_{\mathcal M}(\mathcal D; \boldsymbol \eta),
\label{evidence_max}
\end{align}
where
\begin{align}
p_{\mathcal M}(\mathcal D; \boldsymbol \eta) = \int p_{\mathcal M}(\mathcal D | \boldsymbol \theta) p_{\mathcal M}(\boldsymbol \theta;  \boldsymbol{\eta}_{p})  d\boldsymbol \theta
\label{evidence_def}
\end{align}
is known as the model {\it evidence}, since it measures the plausibility of the dataset $\mathcal D$ given the hyper-parameters $\boldsymbol \eta$. Note that the evidence depends on the model itself and not on any specific value of the parameters $\boldsymbol \theta$, which have been integrated out (marginalized). \textit{This is a crucial difference compared with the discriminative methods.} As it can be shown, the evidence maximization problem  \eqref{evidence_max} involves a trade-off between accuracy (the achieved likelihood value) and model complexity, in line with Occam's razor rule\cite{gull1988bayesian}, \cite{theodoridis2020machine}. This allows computation of the model hyper-parameters $\boldsymbol \eta$ directly from the observed dataset $\mathcal D$. At this point, recall that one of the major difficulties associated with machine learning, and the inverse problems in general, is {\it overfitting}. That is, if the model is too complex with respect to the number of training data samples, then the estimated models learn the specificities of the given training data and cannot generalize well when dealing with new unseen (test) data. 

The use of regularization in the discriminative methods and priors in the Bayesian ones try to achieve the best trade-off between accuracy (fitting to the observed data) and generalization that heavily depends on the complexity of the model, see, e.g., \cite{theodoridis2020machine, Bishop2006} for further discussions. Furthermore, note that in the Bayesian context, \emph{model complexity} is interpreted from a broader view, since it depends not only on the number of parameters but also on the shape  (e.g., variance and skewness) of the involved distributions of $\boldsymbol \theta$, see e.g., \cite{theodoridis2020machine, Bishop2006} for in-depth discussions. For example, under a broad enough Gaussian prior for the model parameters, $\boldsymbol \theta$, and some limiting properties, it can be shown that the evidence in (\ref{evidence_def}) results in the well-known Bayesian information criterion (BIC) for model selection\cite{schwarz1978estimating},\cite{theodoridis2020machine}, which has the form:
\begin{equation}
\log p_{\mathcal{M}}(\mathcal{D};\bm{\eta})=\log p_{\mathcal{M}}(\mathcal{D} | \hat{\bm{\theta}}_{\text{MAP}} )-\frac{L}{2}\log N,
\label{BIC}
\end{equation}
where the first term on the right-hand side is the accuracy (best likelihood fit) term and the second is the complexity term that ``competes''  in a trade-off fashion while maximizing the evidence, see, e.g., \cite{mackay1992bayesian}, \cite{theodoridis2020machine} for further discussion.  In \eqref{BIC}, $L$ denotes the number of unknown parameters in $\boldsymbol \theta$ and $N$ is the size of the training data. A more recent interpretation of this trade-off, in the context of over-parameterized DNNs, is provided in \cite{wilson2020bayesian}, where the prior is viewed as the {\it inductive bias} that favors certain datasets.

\subsubsection{Marginalization for Prediction} The learnt posterior $p_{\mathcal M}(\boldsymbol \theta | \mathcal D;  \boldsymbol \eta)$ provides uncertainty information about $\boldsymbol \theta$, i.e., the plausibility of each possible $\boldsymbol \theta$ to be endorsed by the observed dataset $\mathcal D$, and it can be used to forecast an unseen dataset, $\mathcal D_{\text{new}}$, via \emph{marginalization}:
\begin{align}
p_{\mathcal M}(\mathcal D_{\text{new}} | \mathcal D; \boldsymbol{\eta}) =  \int  p_{\mathcal M}(\mathcal D_{\text{new}} | \boldsymbol \theta) p_{\mathcal M}(\boldsymbol \theta | \mathcal D;  \boldsymbol \eta)   d \boldsymbol \theta.
\label{eq2}
\end{align}
From \eqref{eq2}, Bayesian prediction can be interpreted as the weighted average of the predicted probability  $p_{\mathcal M}(\mathcal D_{\text{new}} | \boldsymbol \theta)$ among all possible model configurations, each of which is specified by different model parameters, $\boldsymbol \theta$, and weighted by the respective posterior $p_{\mathcal M}(\boldsymbol \theta | \mathcal D;  \boldsymbol \eta)$.\footnote{In this paper,  the unseen dataset  $\mathcal D_{\text{new}}$ is assumed to be statistically independent of the training dataset $\mathcal D$. Therefore, $p_{\mathcal M}(\mathcal D_{\text{new}} | \boldsymbol \theta) = p_{\mathcal M}(\mathcal D_{\text{new}} | \mathcal D, \boldsymbol \theta) $.} In other words, prediction {\it does not} depend on a {\it specific} point estimate of the unknown parameters, which equips Bayesian methods with great potential for more robust predictions against the estimation error of $\boldsymbol \theta$, see, e.g., \cite{theodoridis2020machine,Bishop2006}.

In summary, in light of the Bayes' theorem, the four quantities (i.e., prior $p_{\mathcal M}(\boldsymbol \theta;  \boldsymbol{\eta}_{p})$, likelihood $p_{\mathcal M}(\mathcal D | \boldsymbol \theta)$, posterior $p_{\mathcal M}(\boldsymbol \theta | \mathcal D;  \boldsymbol \eta)$ and evidence $p_{\mathcal M}(\mathcal D; \boldsymbol \eta)$) give a new perspective  on the \emph{inverse problem}. The resulting method combines the strength of the selected priors and the likelihood of the observed data to provide a corresponding posterior. The success of such an inference process strongly relies on the following three steps. First, incorporating a prior for each unknown model parameter/function enables one to naturally encode a desired structure into Bayesian learning. As it will be demonstrated in the rest of this article, a prior can be imposed to both parametric models with a fixed number of unknown parameters and non-parametric models that comprise unknown functions and/or an unknown set of parameters whose number is not fixed but it varies with the size of the dataset. Second, through evidence maximization, one can optimize the set of hyper-parameters that is associated with the selected Bayesian learning model to obtain enhanced generalization performance. Finally, marginalization ensures robust prediction and generalization performance by averaging over an ensemble of predictions using all possible parameter/function estimates weighted by the corresponding posterior probability. These three aspects will be discussed in detail in the following sections.

\subsection{Bayesian Linear Parametric Regression: A Pedagogic Example}
\label{subsec:blr-exmaple}
Before moving to our next topics on more advanced Bayesian data analysis, we introduce the \emph{Bayesian linear regression} model as an example to further elaborate the terminology and concepts discussed previously. It also serves as the cornerstone for the two recent supervised learning tools, namely the Bayesian neural networks and GP models to be elaborated in the following subsections.
%

\subsubsection{Linear Regression} In statistics, the term ``regression'' refers to seeking the relationship between a \emph{dependent} random variable, $y$, which is usually considered as the response of a system, and the associated \emph{input/independent} variables, $\boldsymbol x = [x_{1}, x_{2}, \cdots, x_{L}]^T$. When the system is modeled as a linear combiner with an additive disturbance or noise term $v_n$, the relationship between $y_n$ and $\boldsymbol x_n$ of the $n$-th data sample can be expressed as:
\begin{align}
y_n = \boldsymbol \theta^T \boldsymbol x_n + v_n, \quad \forall n \in \{1,2,\cdots,N\},
\label{linear_reg}
\end{align}
which specifies the linear regression task. For simplicity, we assume that the additive noise terms $\{v_n\}$ are independently and identically distributed (i.i.d.)  Gaussian with zero mean and variance $\beta^{-1}$, i.e., $ \{v_n\} \overset{\text{i.i.d.}}{\sim} \mathcal N (v_n ; 0, \beta^{-1})$, where $\beta$ (i.e., inverse of  the variance) is called ``precision'' in statistics and machine learning. The task of linear regression is to learn the weight parameters  $\boldsymbol \theta = [\theta_1, \theta_2, \cdots, \theta_L]^T$ from the training/observed dataset $\mathcal D \triangleq \{\boldsymbol X, \boldsymbol y\} $, where the input matrix $\boldsymbol X \triangleq \left[ \boldsymbol x_1, \boldsymbol x_2, \cdots, \boldsymbol x_N \right]^{T} \in \mathbb R^{N \times L}$, and the output vector $ \boldsymbol y \triangleq  [y_1, y_2, \cdots, y_N]^T$.

\subsubsection{Bayesian Learning} For the linear regression task, we take a Bayesian perspective by treating the unknown parameters  $\boldsymbol \theta $ as a random vector. As introduced in Section \ref{subsec:BayesianPhi}, the inverse problem can be solved via the Bayes' theorem after specifying the following four quantities. 

\noindent $\blacksquare$ \underline{Likelihood.} The easiest one to derive is the likelihood function, which describes the forward problem of linear regression. Owing to the Gaussian and independence properties of the noise terms $\{v_n\}$, the following Gaussian likelihood function can be easily obtained:
\begin{align}
p_{\mathcal M}(\mathcal{D} | \boldsymbol \theta) = \prod_{n=1}^N \mathcal {N} (y_n ; \boldsymbol \theta^T \boldsymbol{x}_n, \beta^{-1}).
\label{linear_reg_likelihood} 
\end{align}   

\noindent $\blacksquare$ \underline{Prior.} Then, we specify a prior on the unknown parameters $\boldsymbol \theta$.  For mathematical tractability, we adopt an i.i.d. Gaussian distribution as the prior:
\begin{align}
p_{\mathcal M} ( \boldsymbol \theta; \boldsymbol{\eta}_{p} ) = \prod_{l=1}^L \mathcal{N}(\theta_l ;  0, \alpha_l^{-1}), 
\label{linear_reg_prior} 
\end{align}
where $\alpha_l$ is the precision associated with each $\theta_l$, and $\boldsymbol{\eta}_{p} = \boldsymbol{\alpha} \triangleq [\alpha_1, \alpha_2, \cdots, \alpha_L]^T$ represents the hyper-parameters associated with the prior.
 
\noindent $\blacksquare$ \underline{Evidence.} After substituting the prior \eqref{linear_reg_prior} and the likelihood \eqref{linear_reg_likelihood} into \eqref{evidence_def}, and performing the integration, we can derive the following Gaussian evidence:
\begin{align}
p_{\mathcal M}( \mathcal{D} ; \boldsymbol \eta ) = \mathcal N ( \boldsymbol y ; \boldsymbol 0, \beta^{-1} \boldsymbol I + \boldsymbol X \boldsymbol A^{-1} \boldsymbol X^T ),
\label{evidence_lr}
\end{align}
where the diagonal matrix $\boldsymbol A \triangleq  \mathrm{diag}\{ \boldsymbol{\alpha} \}$ and  $\boldsymbol I$ denotes the identity matrix. Here, we have $\boldsymbol{\eta} = [\boldsymbol{\eta}_{p}^T, \beta]^T$.

\noindent $\blacksquare$ \underline{Posterior.} Inserting the prior \eqref{linear_reg_prior}, the likelihood \eqref{linear_reg_likelihood} and the evidence \eqref{evidence_def} into the Bayes' theorem \eqref{eq1}, the posterior can be shown to be the Gaussian distribution:
\begin{align}
p_{\mathcal M}(\boldsymbol \theta | \mathcal{D}; \boldsymbol \eta) = \mathcal N (\boldsymbol \theta ; \boldsymbol \mu, \boldsymbol \Sigma),
\label{post_lr}
\end{align}
where
\begin{subequations}
\begin{align}
&\boldsymbol \mu = \beta \boldsymbol \Sigma \boldsymbol X^T \boldsymbol y, \label{post_lr_mean}   \\
&\boldsymbol \Sigma = (\boldsymbol A + \beta \boldsymbol X^T \boldsymbol X)^{-1}.
\label{post_lr_mean_cov} 
\end{align}
\end{subequations}

Once again, taking the above linear regression as a concrete example, we further demonstrate the merits of Bayesian learning in general. 

\noindent $\blacksquare$ \underline{Merit 1: Parameter Learning with Uncertainty Quantification.} Using the Bayes' theorem, the posterior in \eqref{post_lr} not only provides a point estimate $\boldsymbol \mu$ in (\ref{post_lr_mean}) for the unknown parameters $\boldsymbol \theta$, but also provides a covariance matrix $\boldsymbol \Sigma$ in (\ref{post_lr_mean_cov}) that describes to which extent the posterior distribution is centered around the point estimate $\boldsymbol \mu$. In other words, it quantifies our uncertainty about the parameter estimate, which cannot be naturally obtained in any discriminative method. For the above example, we have $\hat{\boldsymbol{\theta}}_{\textrm{MAP}} = \hat{\boldsymbol{\theta}}_{\textrm{MMSE}}$ because the posterior distribution $p_{\mathcal M}(\boldsymbol \theta | \mathcal{D}; \boldsymbol \eta)$ follows a unimodal Gaussian distribution. Of course, Frequentist methods can also construct uncertainty region/confidence intervals by taking a few extra steps once the parameter estimates have been obtained. However, the Bayesian method provides, in one go, the posterior distribution of the model parameters, from which both a point estimate as well as the uncertainty region can be optimally derived via the learning optimization step.

\noindent $\blacksquare$ \underline{Merit 2: Robust Prediction via Marginalization.}  After substituting \eqref{post_lr} and \eqref{linear_reg_likelihood} tailored to new observations into \eqref{eq2}, the posterior/predictive distribution for a novel input $\boldsymbol x_{*}$ is:
\begin{align}
p_{\mathcal M}(y_{*} | \{ \boldsymbol X, \boldsymbol y\}) &=  \int   \mathcal {N} (y_{*} ; \boldsymbol \theta^T \boldsymbol x_{*}, \beta^{-1}) \mathcal N (\boldsymbol \theta ; \boldsymbol \mu, \boldsymbol \Sigma )  d \boldsymbol{\theta} \nonumber \\
& = \mathcal N (y_{*} ;  \boldsymbol \mu^T \boldsymbol x_{*}, \beta^{-1} +  \boldsymbol x_{*}^T \boldsymbol \Sigma  \boldsymbol x_{*} ).
\label{pred_dis}
\end{align}
The predicted value of $y_{*}$ can be acquired via  $\boldsymbol \mu^T \boldsymbol x_{*}$, and the posterior variance, $\beta^{-1} +  \boldsymbol x_{*}^T \boldsymbol \Sigma \boldsymbol x_{*} $, quantifies the uncertainty about this point prediction. Rather than providing a point prediction like in the discriminative methods, Bayesian methods advocate averaging all possible predicted values via marginalization, and are thus more robust against erroneous parameter estimates.

\subsection{Bayesian Nonlinear Nonparametric Model: GP Regression Example}
\label{subsec:gp}
In order to improve the data representation power of Bayesian linear parametric models, a lot of efforts have been invested on designing \emph{non-linear} and \emph{non-parametric} models. A direct non-linear generalization of \eqref{linear_reg} is given by 
\begin{align}
y=f(\boldsymbol{x})+v,
\label{12a}
\end{align}
where instead of the linearity of \eqref{linear_reg},  we employ a non-linear functional dependence $f(\boldsymbol{x})$ and let $v$ be the noise term like before. Moreover, the randomness associated with the weight parameters $\boldsymbol{\theta}$ in \eqref{linear_reg} is now embedded into the function $f(\boldsymbol{x})$ itself, which is assumed to be a {\it random process}. That is, the outcome/realization of each random experiment is a \textit{function} instead of a single value/vector. Thus, in this case, we have to deal with priors related to non-linear functions directly, rather than indirectly, i.e., by specifying a family of non-linear parametric functions and placing priors over the associated weight parameters. 

\subsubsection{GP Model} In the sequel, we introduce one representative model that adopts this rationale, namely the GP model for non-linear regression. The GP models constitute a special family of random processes where the outcome of each experiment is a function or a sequence. For instance, in signal processing, this can be a continuous-time signal $f(t)$ as a function of time $t$ or a discrete-time signal $f(n)$ in terms of the sequence index $n$. In this article, we treat the GP model as a data analysis tool whose input that acts as the argument in $f(\cdot)$ is a vector, i.e.,  vector, $\boldsymbol{x} = [x_1, x_2,\cdots, x_L]^T$ \cite{RW06, theodoridis2020machine}. 

For clarity, we give the definition of GP as follows:
\begin{tcolorbox}
\textit{Definition of GP} \cite{RW06, theodoridis2020machine}: A random process, $f(\boldsymbol{x})$, is called a GP if and only if for any finite number of points, $\boldsymbol{x}_{1}, \boldsymbol{x}_{2}, \cdots, \boldsymbol{x}_{N}$, the associated joint probability density function (pdf), $p(f(\boldsymbol{x}_{1}), f(\boldsymbol{x}_{2}), \cdots, f(\boldsymbol{x}_{N}) )$ is Gaussian. 
\end{tcolorbox}

A GP can be considered as an infinitely long vector of jointly Gaussian distributed random variables, so that it can be fully described by its mean function and covariance function, defined as follows:
\begin{equation}
m(\boldsymbol{x}) \triangleq \mathbb{E}[ f(\boldsymbol{x}) ],
\label{eq:gp-prior-mean}
\end{equation}
\begin{equation}
\textrm{cov}(\boldsymbol{x}, \boldsymbol{x}') \triangleq \mathbb{E}[ (f(\boldsymbol{x}) - m(\boldsymbol{x})) (f(\boldsymbol{x}') - m(\boldsymbol{x}'))].
\label{eq:gp-prior-variance}
\end{equation}
A GP is said to be stationary if the mean function, $m(\boldsymbol{x})$, is a constant mean, and moreover its covariance function has the following simplified form: $\textrm{cov}(\boldsymbol{x}, \boldsymbol{x}') = \textrm{cov}(\boldsymbol{\tau})$ with $ \boldsymbol{\tau} \triangleq  \boldsymbol{x} - \boldsymbol{x}'$.

When a GP is adopted for data modeling and analysis, we need to specify the mean function and the covariance function in order to make the model match the underlying data patterns. The mean function is often set to zero especially when there is no prior knowledge available. The data representation power of the non-parametric GP models is determined overwhelmingly by the covariance function, which is also known as the \textit{kernel function} due to the positive semi-definite nature of a covariance function. In the following, we use $k(\boldsymbol{x}, \boldsymbol{x}'; \boldsymbol{\eta}_{p} ) \triangleq  \textrm{cov}(\boldsymbol{x}, \boldsymbol{x}') $ to represent a pre-selected kernel function with an explicit set of tuning kernel hyper-parameters, $\boldsymbol{\eta}_{p}$, for the observed data. Finally, we say that a function realization is drawn from the GP prior, and we write 
\begin{equation}
f(\boldsymbol{x}) \sim \mathcal{GP} \left( m(\boldsymbol{x}), k(\boldsymbol{x}, \boldsymbol{x}'; \boldsymbol{\eta}_{p}) \right).
\label{eq:GP-prior}
\end{equation} 
The consequent GP regression model follows \eqref{12a}, where $f(\boldsymbol{x})$ is represented by a GP model defined in (\ref{eq:GP-prior}) and the noise term $v$ is assumed to be Gaussian distributed with zero mean and variance $\beta^{-1}$, like in the previous simple Bayesian linear regression example. 

\subsubsection{GP Kernel Functions} As mentioned before, the kernel function plays a crucial role in determining a GP model's representation power. To shed more light on the kernel function, especially on how it represents random functions as well as its good physical interpretations, we demonstrate the most widely used (but not necessarily optimal) squared-exponential (SE) kernel. 

\noindent $\blacksquare$ \underline{SE Kernel}: The form of this widely used kernel function is given below:
\begin{equation}
k(\boldsymbol{x}, \boldsymbol{x}'; \boldsymbol{\eta}_{p} ) = \sigma_s^2 \, \textrm{exp}\left( - \frac{|| \boldsymbol{x} - \boldsymbol{x}' ||_{2}^2}{ 2 \ell^2} \right),
\label{eq:SE-kernel}
\end{equation}
where the hyper-parameter $\sigma_s^2$ determines the magnitude of fluctuation of $f(\boldsymbol{x})$ and the other hyper-parameter $\ell$, called \textit{length-scale}, determines the statistical correlation between two points, $f(\boldsymbol{x})$ and $f(\boldsymbol{x}')$, separated by a (Euclidean) distance $d \triangleq || \boldsymbol{x} - \boldsymbol{x}' ||_{2}$. Thus, we have the kernel hyper-parameters, $\boldsymbol{\eta}_{p} = [\sigma_s^2, \ell]^T$, specifically for this kernel.  

\begin{figure}[t]
\begin{minipage}[t]{0.48\textwidth}  \centering
\includegraphics[width=8.5cm]{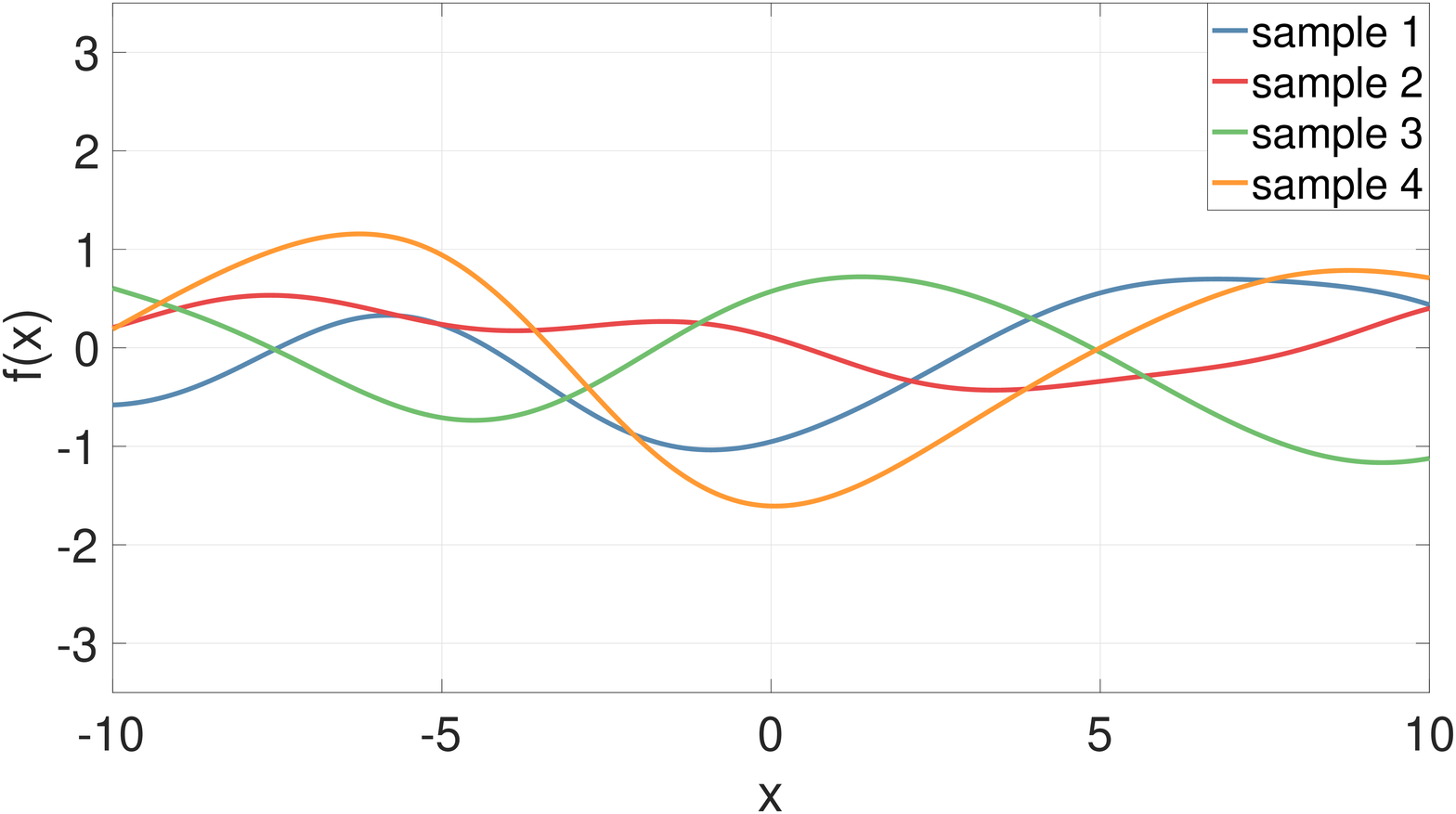}
\caption*{(a)}
\end{minipage}
\begin{minipage}[t]{0.48\textwidth}  \centering
\includegraphics[width=8.5cm]{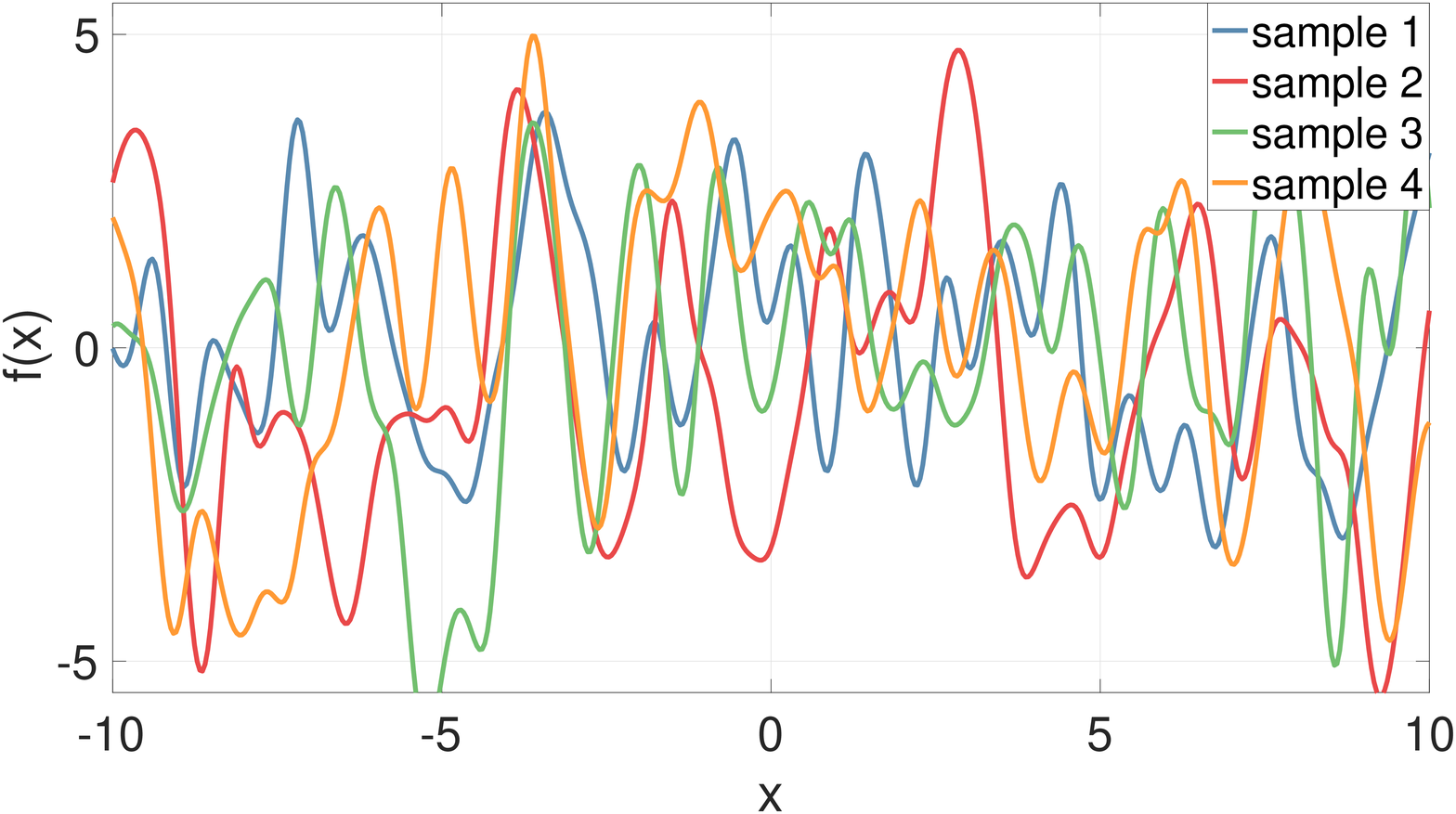}
\caption*{(b)}
\end{minipage}
\caption{Sample functions generated from a GP model using the SE kernel with two different hyper-parameter configurations. (a) GP with hyper-parameters, $\sigma_s^2=1, \ell=5$, generates low-peaked and smooth sample functions; (b) GP with hyper-parameters, $\sigma_s^2=5, \ell=0.5$, generate high-peaked and fast varying sample functions.}
\label{fig:gp_SE_prior_hyper}
\end{figure}

In Fig.~\ref{fig:gp_SE_prior_hyper}, we show some sample functions generated from a GP (for one-dimensional input, $x$) involving the SE kernel with different hyper-parameter configurations. From these illustrations, we can clearly spot the physical meaning of the SE kernel hyper-parameters. There are many other classic kernel functions, such as the Ornstein-Uhlenbeck kernel, rational quadratic kernel, periodic kernel, locally periodic kernel as introduced in, e.g., \cite{RW06}. They can even be combined, for instance in the form of a linearly-weighted sum, to enrich the overall modeling capacity \cite{RW06, Xu19}. 

Designing a competent stationary kernel function for the GP model can also be considered in the frequency domain owing to the famous \textit{Wiener-Khintchine Theorem} \cite{RW06, theodoridis2020machine}. The theorem states that the Fourier transform of a stationary kernel function, $k(\boldsymbol{\tau})$, and the associated \textit{spectral density} of the process, $S(\mathbf{s})$, are Fourier duals:
\begin{equation}
k(\boldsymbol{\tau})=\int S(\mathbf{s}) e^{2 \pi i \mathbf{s}^{T} \boldsymbol{\tau}} d \mathbf{s},  \quad S(\mathbf{s})=\int k(\boldsymbol{\tau}) e^{-2 \pi i \mathbf{s}^{T} \boldsymbol{\tau}} d \boldsymbol{\tau}.
\end{equation}
Here, it is noteworthy to mention that $i$ is the imaginary unit and the operation $\mathbf{s}^{T} \boldsymbol{\tau} $ refers to the inner product of the generalized frequency parameters, $\mathbf{s}$, and the time difference parameters, $\boldsymbol{\tau}$. In Section~\ref{sec:prior-with-recent-tools}, we will introduce some optimal kernel design methods that were first built based on the spectral density in the frequency domain and then transformed back to the original input domain. 

\subsubsection{GP for Regression} In contrast to the Bayesian linear regression model, we set a GP prior directly on the underlying function in the GP regression model, namely, $f(\boldsymbol{x}) \sim \mathcal{GP}(m(\boldsymbol{x}), k(\boldsymbol{x}, \boldsymbol{x}'; \boldsymbol{\eta}_{p} ))$. Given the observed dataset, $\mathcal{D} = \{ \boldsymbol X, \boldsymbol y \}$ as defined before, the main goal of GP-based Bayesian non-linear regression is to compute the evidence, $p(\boldsymbol{y}; \boldsymbol{\eta})$, for optimizing the model hyper-parameters, $\boldsymbol{\eta}$, and to compute the posterior distribution, $p(\boldsymbol{y}_{*}| \boldsymbol{y})$, of $\boldsymbol{y}_{*} =  [y_{*,1}, y_{*,2},\cdots,y_{*,N_{*}}]^T$ evaluated at $n_{*}$ novel test inputs $\boldsymbol{X}_{*} = [\boldsymbol{x}_{*,1}, \boldsymbol{x}_{*,2},\cdots,\boldsymbol{x}_{*, N_{*}}]^T$. 

\noindent $\blacksquare$ \underline{Evidence}: This can be obtained in a straightforward way due to the regression model $\boldsymbol{y} = \boldsymbol{f}(\boldsymbol{X}) + \boldsymbol{v}$, where $\boldsymbol{v}$ is independent of the GP model, $f(\boldsymbol{x}) \sim \mathcal{GP}(m(\boldsymbol{x}), k(\boldsymbol{x}, \boldsymbol{x}'; \boldsymbol{\eta}_{p}))$,
 and we let $\boldsymbol{v} \sim \mathcal{N}(\mathbf{0}, \beta^{-1} \boldsymbol{I})$. As a consequence, it is easy to derive, see e.g., \cite{theodoridis2020machine, RW06}, that 
\begin{equation}
p(\boldsymbol{y}; \boldsymbol{\eta}) = \mathcal{N}(\boldsymbol{y}; \mathbf{0}, \boldsymbol{K}(\boldsymbol{X}, \boldsymbol{X}; \boldsymbol{\eta}_{p}) + \beta^{-1} \boldsymbol{I}),
\label{eq:GP-evidence}
\end{equation}
where $\boldsymbol{\eta} = [\boldsymbol{\eta}_{p}^T, \beta]^T$ and $\boldsymbol{K}(\boldsymbol{X}, \boldsymbol{X}; \boldsymbol{\eta}_{p})$ is the $N \times N$ kernel matrix of $\boldsymbol{f}(\boldsymbol{X}) \triangleq  [f(\boldsymbol{x}_1), f(\boldsymbol{x}_2),\cdots,f(\boldsymbol{x}_n)]^T$ evaluated for the training samples. Note that the kernel matrix is a square matrix whose $(ij)$-th entry is the pairwise covariance between $f(\boldsymbol{x}_i)$ and $f(\boldsymbol{x}_j)$, computed according to (\ref{eq:gp-prior-variance}), for any $\boldsymbol{x}_i$ and $\boldsymbol{x}_j$ in the training dataset. The covariance matrix, $\boldsymbol{X} \boldsymbol{A}^{-1} \boldsymbol{X}$, of the Bayesian linear regression function, $f(\boldsymbol{x})= \boldsymbol{\theta}^T \boldsymbol{x}$, given in (\ref{evidence_lr}) can be regarded as one instance of the kernel matrix, $\boldsymbol{K}(\boldsymbol{X}, \boldsymbol{X}; \boldsymbol{\eta}_{p})$.  The latter can provide increased representation power through choosing more appropriate kernel forms and tuning the associated kernel hyper-parameters. As it will be shown in Section \ref{sec:inference}, we will maximize this evidence function to get an optimal set of the model hyper-parameters. 

\noindent $\blacksquare$ \underline{Posterior Distribution}: It turns out, see e.g. \cite{theodoridis2020machine, Bishop2006}, that the joint distribution of the training output $\boldsymbol{y}$ and the test output $\boldsymbol{y}_{*}$ is a Gaussian, of the following form:
\begin{equation}
\begin{bmatrix} \boldsymbol{y} \\ \boldsymbol{y}_{*} \end{bmatrix} \sim \mathcal{N} \left( \begin{bmatrix} \boldsymbol{y} \\ \boldsymbol{y}_{*} \end{bmatrix};   \boldsymbol{0},  \begin{bmatrix} \boldsymbol{K}(\boldsymbol{X}, \boldsymbol{X}) + \beta^{-1} \boldsymbol{I}, & \!\!\!\!\!\!\!\! \boldsymbol{K}(\boldsymbol{X}, \boldsymbol{X}_{*}) \\ \boldsymbol{K}(\boldsymbol{X}_{*}, \boldsymbol{X}), & \!\!\!\!\!\!\!\! \boldsymbol{K}(\boldsymbol{X}_{*}, \boldsymbol{X}_{*}) + \beta^{-1} \boldsymbol{I} \end{bmatrix} \right),  
\label{eq:joint-prior}
\end{equation}
where $\boldsymbol{K}(\boldsymbol{X}, \boldsymbol{X}_{*})$ stands for the $N \times N_{*}$ kernel matrix between the training inputs and test inputs and $\boldsymbol{K}(\boldsymbol{X}_{*}, \boldsymbol{X}_{*})$ for the $N_{*} \times N_{*}$ kernel matrix among the test inputs. Here, we let $\boldsymbol{K}(\boldsymbol{X}, \boldsymbol{X})$ be a short form of $\boldsymbol{K}(\boldsymbol{X}, \boldsymbol{X}; \boldsymbol \eta)$.

By applying some classic conditional Gaussian results, see e.g., \cite{theodoridis2020machine}, we can derive the posterior distribution from the joint distribution in (\ref{eq:joint-prior}) as:
\begin{align}
p(\boldsymbol{y}_{*} \vert \boldsymbol{y} ) \sim \mathcal{N} \left( \boldsymbol{y}_{*}; \bar{\boldsymbol{m}} , \bar{\boldsymbol{V}}  \right), 
\label{eq:GP-posterior}
\end{align} 
where the posterior mean (vector) and the posterior covariance (matrix) are respectively,
\begin{align}
\label{eq:posterior-mean}
\bar{\boldsymbol{m}} & = \boldsymbol{K}(\boldsymbol{X}_{*}, \boldsymbol{X}) \left[ \boldsymbol{K}(\boldsymbol{X}, \boldsymbol{X}) + \beta^{-1} \boldsymbol{I} \right]^{-1} \boldsymbol{y}, \\
\label{eq:posterior-variance}
\bar{\boldsymbol{V}} & = \boldsymbol{K}(\boldsymbol{X}_{*}, \boldsymbol{X}_{*}) + \beta^{-1} \boldsymbol{I} - \boldsymbol{K}(\boldsymbol{X}_{*}, \boldsymbol{X}) \left[ \boldsymbol{K}(\boldsymbol{X}, \boldsymbol{X}) + \beta^{-1} \boldsymbol{I} \right]^{-1}  \boldsymbol{K}(\boldsymbol{X}, \boldsymbol{X}_{*}).
\end{align}
The above posterior mean gives a point prediction, while the posterior covariance defines the uncertainty region of such prediction. A leading benefit of using the GP models over discriminative methods, such as the kernel ridge regression, lies in the natural uncertainty quantification given by (\ref{eq:posterior-variance}).

\begin{figure}[t]
\begin{minipage}[t]{0.48\textwidth}  \centering
\includegraphics[width=8.5cm]{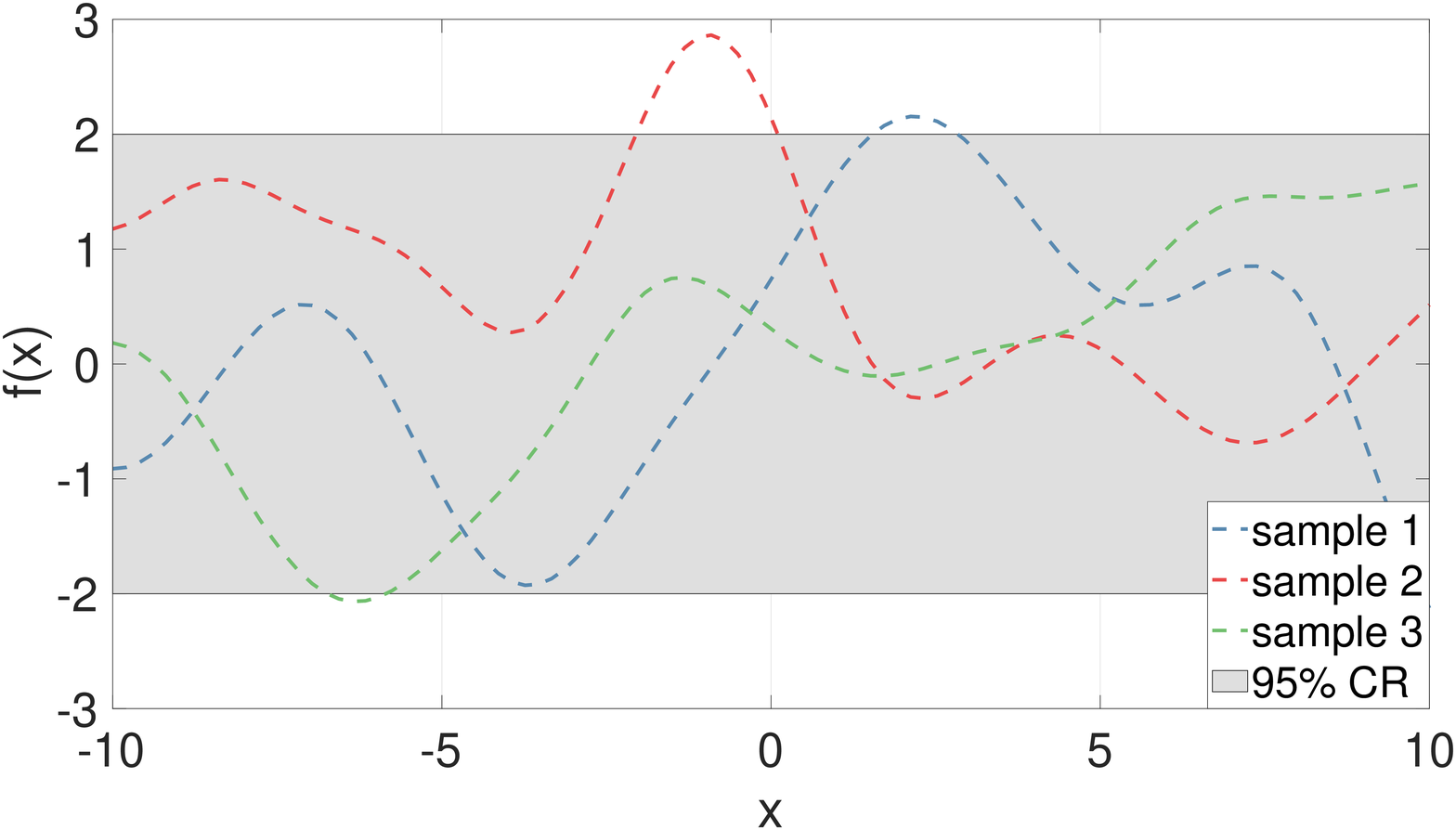}
\caption*{(a)}
\end{minipage}
\begin{minipage}[t]{0.48\textwidth}  \centering
\includegraphics[width=8.5cm]{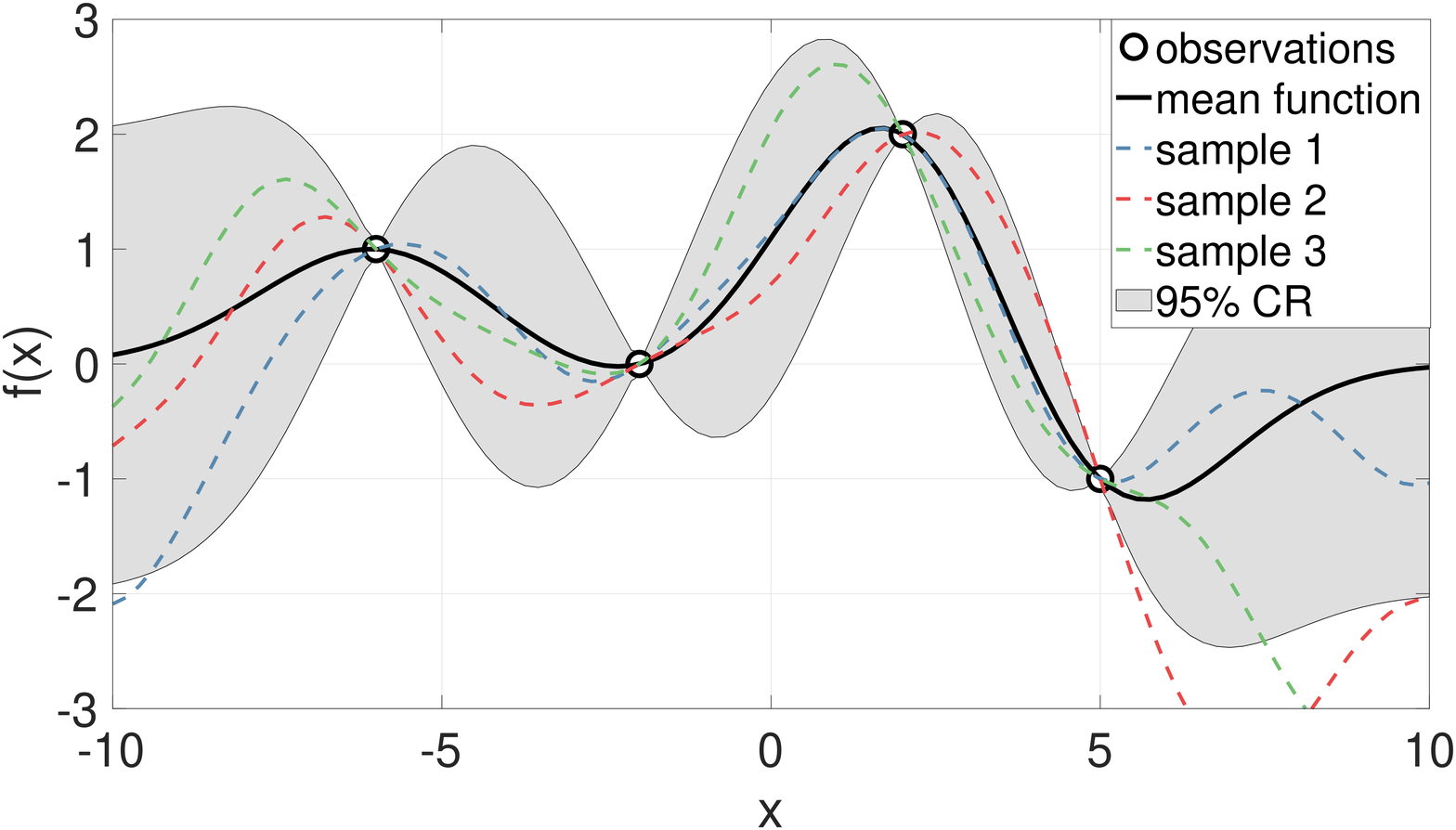}
\caption*{(b)}
\end{minipage}
\caption{Subfigure (a) shows three sample functions drawn randomly from a GP prior (refer to \eqref{eq:GP-prior}) with an SE kernel (refer to \eqref{eq:SE-kernel}). Subfigure (b) shows three sample functions drawn from the GP posterior (refer to \eqref{eq:GP-posterior}, \eqref{eq:posterior-mean}, \eqref{eq:posterior-variance}) computed based on the prior shown in (a) as well as four noisy observations indicated by the black circles. The corresponding posterior mean function is depicted by the dark black curve. The grey shaded area represents the uncertainty region, taken as the $95\%$ confidence region (CR) for both the prior and the posterior herein.}
\label{fig:gp_model}
\end{figure}

A graphical illustration of GP working on a toy regression example is shown in Fig.~\ref{fig:gp_model}. As we can see from the figures, the uncertainty in the GP prior is constantly large, reflecting our crude prior belief in the underlying function. While it has been significantly reduced in the neighborhood of the observed data points in the GP posterior, but still remains comparably large in regions where the observed data points are scarce.

Apart from the representation power of the GP model, it also connects to various other salient machine learning models, including for instance the relevance vector machine, support vector machine \cite{RW06}. Also, it has been shown that a neural network, with one or multiple hidden layers, asymptotically approaches a GP, see e.g. \cite{MacKayGP97, Lee18}.

\section{Sparsity-Aware Learning: Regularization Functions And Prior Distributions}
\label{sec:sal}
In modern big data analysis, there is a trend to employ sophisticated models that involve an excessive number of parameters (sometimes even more than the number of data samples, e.g., in \emph{over-parameterized models}). This makes the learning systems vulnerable to overfitting to the observed data. Thus, the obvious question concerns the right model size given the data sample.  \emph{Sparsity-aware learning} (SAL) that promotes sparsity on the structure of the learnt model comprises a major path in dealing with such models in a data-adaptive fashion. The term {\it sparsity} implies that most of the unknown parameters are pushed to (almost) zero values. This can be achieved either via the combination of a discriminative method and appropriate regularizers, or via the Bayesian path by adopting sparsity-promoting priors. The major difference between the two paths lies in the way ``sparsity'' is interpreted and embedded into the models, as it is explained in the following subsections. 

In the sequel, we will first introduce the first path that leads to SAL via regularized optimization methods in Section~\ref{subsec:SAL-RegOpt}, followed by the SAL via Bayesian methods for parametric models in Section~\ref{subsec:SAL-bayes-parametric} and non-parametric models in Section~\ref{subsec:SAL-bayes-nonparametric}. Note that the aim of this article is not on comparing the two paths, but rather to rethink the Bayesian philosophy.

\subsection{SAL via Regularized Optimization Methods} 
\label{subsec:SAL-RegOpt}
Following the regularized optimization way, ``sparsity'' information is embedded through \emph{regularization} functions. Using the linear regression task as an example, the regularized parameter optimization problem is formulated as: 
\begin{align}
\min_{\boldsymbol \theta} ~ \underbrace{\frac{1}{2} \sum_{n=1}^N  \left( y_n - \boldsymbol \theta^T \boldsymbol x_n \right)^2}_{\text{data fitting cost}} ~ +   \underbrace{\lambda}_{\text{regularization parameter}}  \times \underbrace{r(\boldsymbol \theta)}_{\text{regularization function}},
\label{ls_reg}
\end{align}
where the regularization function $r(\boldsymbol \theta)$ steers the solution towards a preferred sparse structure, and the regularization parameter $\lambda$ is to balance the trade-off between the data fitting cost and the regularization function for sparse structure embedding. In SAL, it is assumed that the unknown parameters $\boldsymbol \theta$ have a majority of zero entries, and thus the adopted regularization function $r(\boldsymbol \theta)$  should help the optimization process unveil such zeros. Such regularization functions include the family of $l_p$ norm functions with $0 \leq p \leq 1$, among which the $\ell_1$ norm is most popular, since it retains the computationally attractive property of convexity. Furthermore, strong theoretical results have been derived, see e.g., \cite{eldar2012compressed, theodoridis2020machine}. In recent years, SAL advances via regularized cost optimization prevail in the context of machine learning using data analysis tools. The literature is very rich and fairly well documented with many sparsity-promoting regularization functions. Although the resulting regularized cost function might be non-convex and/or non-smooth, efficient learning algorithms exist and have been built on solid theoretical foundations in optimization theory, see e.g., \cite{elad2010sparse}.

\subsection{SAL via Bayesian Methods For Parametric Models}
\label{subsec:SAL-bayes-parametric}
\begin{figure}[!t]
\begin{minipage}[t]{0.48\textwidth}  \centering
\includegraphics[width=8cm]{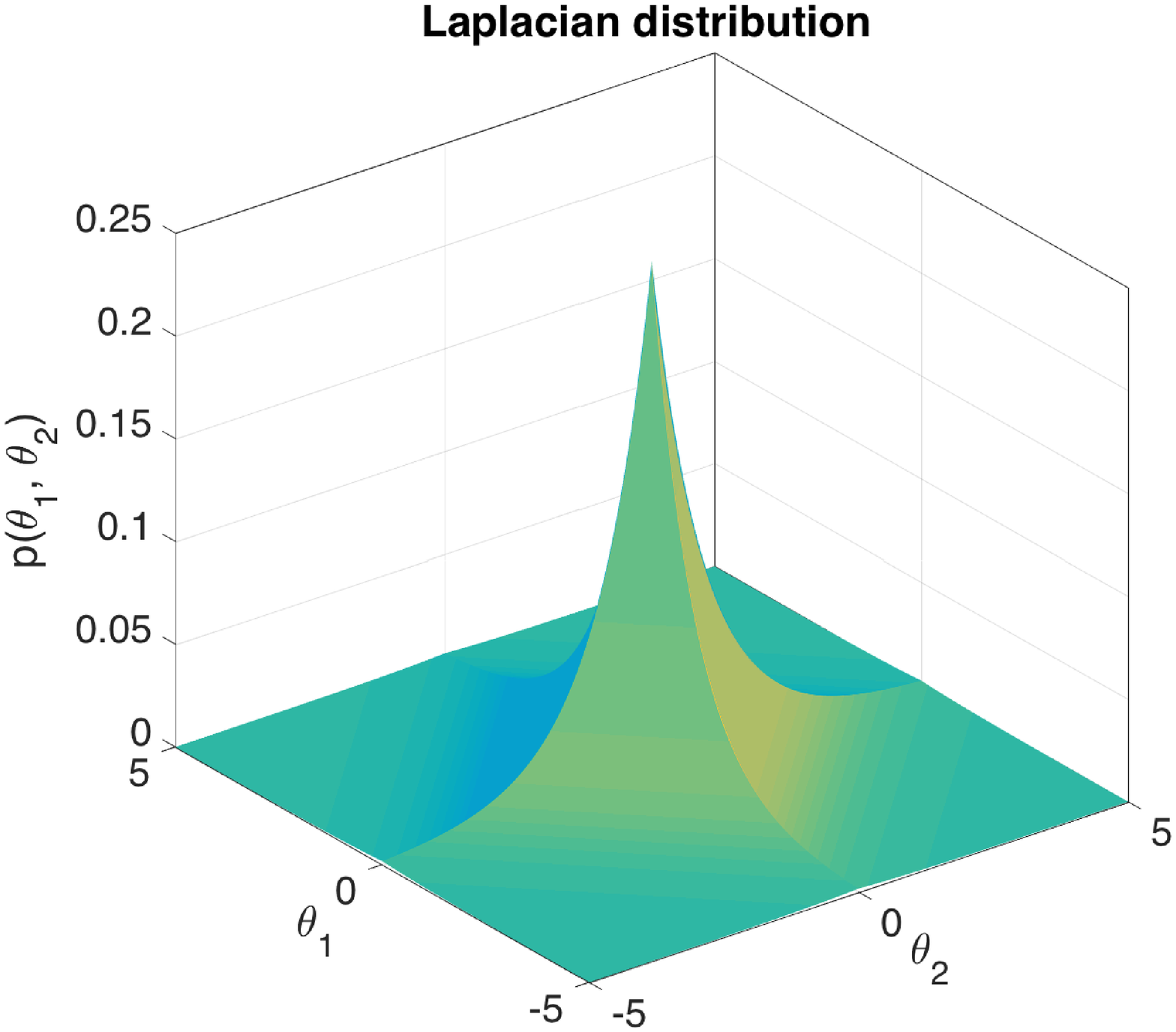}
\caption*{(a)}
\end{minipage}
\begin{minipage}[t]{0.48\textwidth}  \centering
\includegraphics[width=8cm]{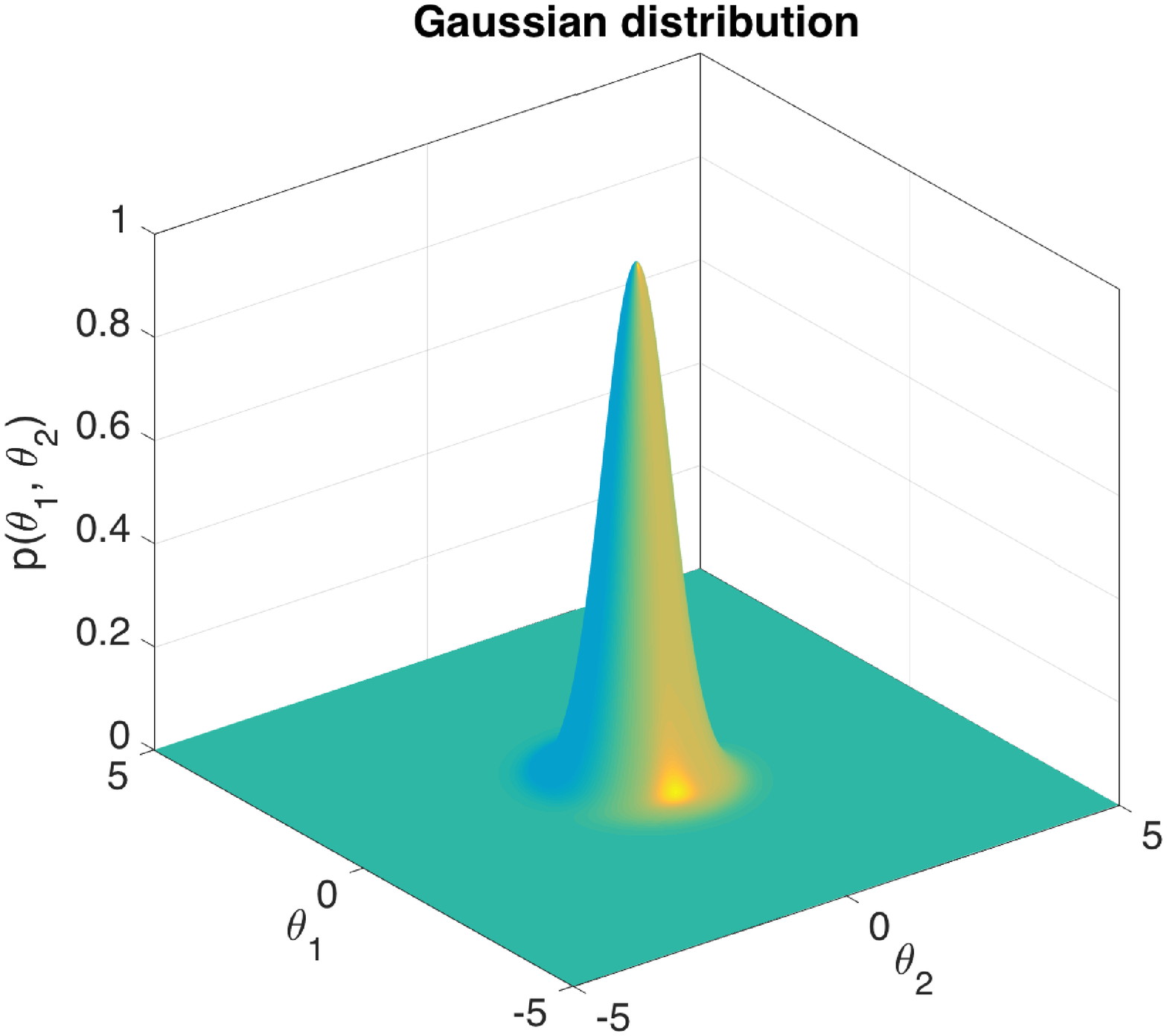}
\caption*{(b)}
\end{minipage}
\caption{Joint probability distribution of the model parameters in two-dimensional space. Subfigure (a) shows the Laplacian distribution  and subfigure (b) shows the Gaussian distribution. The heavy-tail Laplacian distribution peaks sharply around zero and falls slowly along the axes, thus promoting sparse solutions in a probabilistic manner. On the contrary, the Gaussian distribution decays more rapidly along both dimensions compared to the Laplacian distribution.}
\label{fig:heavy_tail_model-1}
\end{figure}

Before we entail into a more formal presentation of a family of probability density functions (pdfs) that promote sparsity, let us first view sparsity from a slightly different angle. It is well known that there is a bridge between the estimate obtained from problem \eqref{ls_reg} and the MAP estimate (see Section~\ref{subsec:BayesianPhi}). For example, it is not difficult to see that if $r(\boldsymbol \theta)$ is the squared Euclidean norm (i.e., the $\ell_2$ norm that gives rise to the so-called ridge regression), the resulting estimate corresponds to the MAP one when assuming the noise to be i.i.d. Gaussian and the prior on $\bm \theta$ to be also of a Gaussian form. If, on the other hand,  $r(\boldsymbol \theta)$ takes the $\ell_1$ norm, this corresponds to imposing a Laplacian prior, instead of a Gaussian one, on $\boldsymbol \theta$. For comparison, Fig.~\ref{fig:heavy_tail_model-1} presents the \emph{Laplacian and Gaussian priors} for $\bm \theta \in \mathbb R^{2}$. It is readily seen that the Laplacian distribution is heavy-tailed compared to the Gaussian one. In other words, the probability that the parameters will take non-zero values, for the zero-mean Gaussian, goes to zero very fast. Most of the probability mass concentrates around zero. This is bad news for sparsity, since we want most of the values to be (close to) zero, but still some of the parameters to have large values. In contrast, observe that in the Laplacian, although most of the probability mass is close to zero, yet there is still high enough probability for non-zero values. More importantly, this probability mass is concentrated along the axes, where one of the parameters is zero. This is how Laplacian prior promotes sparsity. Thus, to practice ``Bayesianism'', one explicit path is to construct priors with heavy tails to promote sparsity. In the sequel, we introduce an important family of such sparsity-promoting priors. 

\subsubsection{The Gaussian Scale Mixture (GSM) Prior}  
\begin{figure}[!t]
\begin{minipage}[h]{0.48\textwidth}  \centering
\includegraphics[width=8cm]{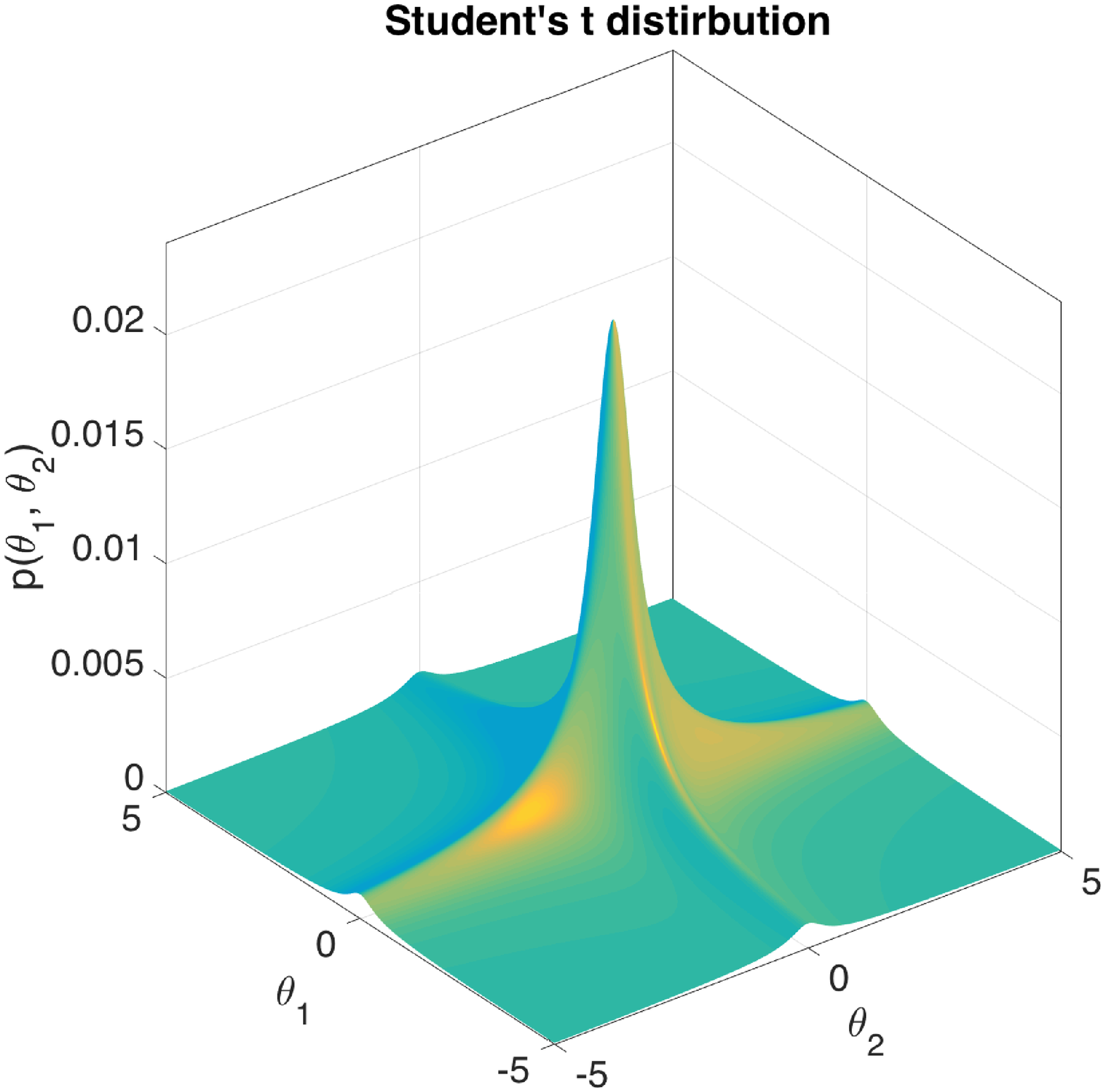}
\caption*{(a)}
\end{minipage}
\begin{minipage}[h]{0.48\textwidth}  \centering
\includegraphics[width=8cm]{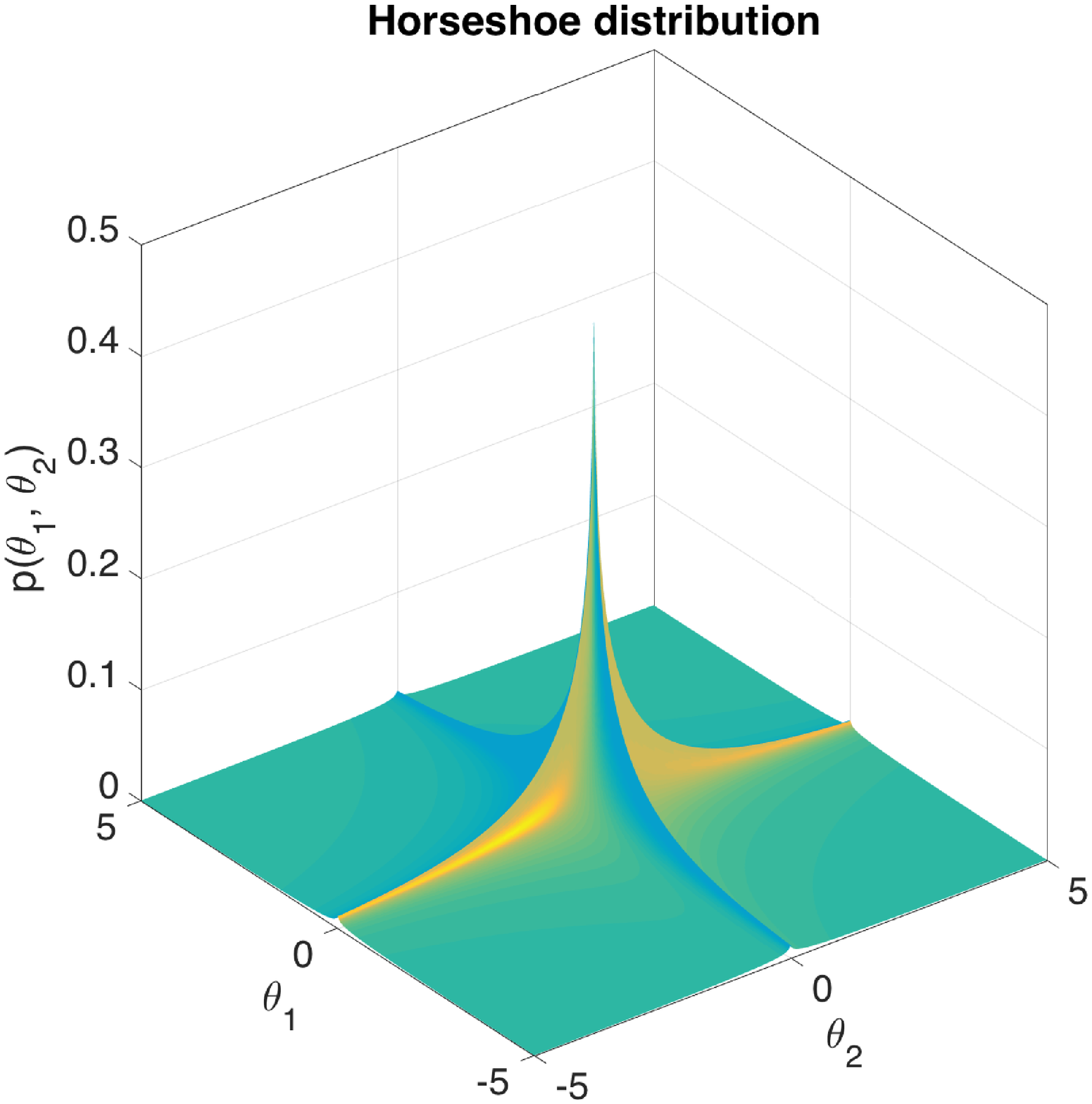}
\caption*{(b)}
\end{minipage}
\caption{Representative GSM prior distributions in two-dimensional space. Subfigure (a) and Subfigure (b) show the Student's $t$ distribution and the horseshoe distribution, respectively. It can be seen that these two distributions show different heavy-tail profiles and are both sparsity-promoting.}
\label{fig:heavy_tail_model-GSM}
\end{figure}

The kick-off point of the GSM prior, see e.g., \cite{andrews1974scale}, is to assume that: a) the parameters, $\theta_l,~ l=1, 2, \cdots, L$, are mutually statistically independent; b) each one of them follows a Gaussian prior with zero mean and c) the respective variances, $\zeta_l, ~l=1,2, \cdots, L$, are also random variables, each one following a prior  $p(\zeta_l; \boldsymbol{\eta}_{p} )$, where $\boldsymbol{\eta}_{p}$ is a set of tuning hyper-parameters associated with the prior. Thus, the GSM prior for each $\theta_l$ is expressed as 
\begin{align}
p( \theta_l; \boldsymbol{\eta}_{p}) = \int \mathcal N ( \theta_l ; 0,  \zeta_l) p(\zeta_l; \boldsymbol{\eta}_{p} ) d\zeta_l.
\label{eq4}
\end{align}
By varying the functional forms of $p(\zeta_l; \boldsymbol{\eta}_{p} )$, the marginalization (i.e., integrating out the dependence on $\zeta_l$) performed in light of \eqref{eq4} induces different prior distributions of $\boldsymbol{\theta}$. For example, if $ p(\zeta_l; \boldsymbol{\eta}_{p})$ is an inverse Gamma distribution, \eqref{eq4} induces a Student's $t$ distribution \cite{andrews1974scale}; if $ p(\zeta_l; \boldsymbol{\eta}_{p})$ is a Gamma distribution, \eqref{eq4} induces a Laplacian distribution \cite{andrews1974scale}. For clarity, Table~\ref{tab2} summarizes different heavy-tail distributions, including Normal-Jefferys, generalized hyperbolic, and horseshoe distributions, among others. To illustrate graphically the sparsity-promoting property endowed by their heavy-tail nature, in addition to the Laplacian distribution plotted in Fig. \ref{fig:heavy_tail_model-1}, we further depict two representative GSM prior distributions, namely the Student's $t$ distribution and the horseshoe distribution, in Fig. \ref{fig:heavy_tail_model-GSM}.  In Section \ref{subsec:SA-BNN} and \ref{subsec:SA-tensor}, we will show the use of GSM prior in modeling Bayesian neural networks and low-rank tensor decomposition models, respectively. 
\begin{table}[!t]
\centering 
\caption{Examples of GSM prior. Abbreviations: Ga = Gamma, IG = inverse Gamma, GIG = generalized inverse Gaussian, $C^+$ = Half Cauchy.}
\begin{tabular}{|c||c|l|}
\hline
GSM prior $p(\theta_l)$         & Mixing distribution  $p(\zeta_l)$                                                                                                       \\ \hline \hline
Student's $t$      & Inverse Gamma:  $p(\zeta_l; \boldsymbol{\eta}_{p} = [a, b]) = \text{IG}( \zeta_l ; a, b)$                                                                                                            \\ \hline
Normal-Jefferys     & Log-uniform:  $p(\zeta_l;  \boldsymbol{\eta}_{p} = [~ ]) \propto \frac{1}{|\zeta_l |}$                                                                                                            \\ \hline
Laplacian              & Gamma: $p(\zeta_l;  \boldsymbol{\eta}_{p} = [a, b] ) = \text{Ga}(\zeta_l ;  a, b)$                                                                                                                  \\ \hline
Generalized hyperbolic & \begin{tabular}[c]{@{}c@{}}Generalized inverse Gaussian:\\ $p(\zeta_l ;  \boldsymbol{\eta}_{p} = [a, b, \lambda]) = \text{GIG}(\zeta_l ; a, b, \lambda)$  \end{tabular}             \\ \hline
Horseshoe              & \begin{tabular}[c]{@{}c@{}}  $\zeta_l = \tau_l \upsilon_l, \boldsymbol{\eta}_{p} = [a, b]  $\\ Half Cauchy: $p(\tau_l) = C^+(0,a)$ \\ $~~~~~~~~~~~~~~~~p(\upsilon_l)= C^+(0,b)$\end{tabular}   \\ \hline
\end{tabular}
\label{tab2}
\end{table}

Besides the aforementioned families of sparsity-promoting priors, another path that has been followed to impose sparsity exploits the underlying property of the evidence function to provide a trade-off between the fitting accuracy and the model complexity, at its maximum value, as it has already been discussed in Section \ref{subsec:BayesianPhi}. To this end, one imposes an individual Gaussian prior $\mathcal{N}(0,\zeta_l)$ on each one of the unknown parameters, which are assumed to be mutually independent, and then treats the respective variances, $\zeta_l,~l=1,2, \cdots,L$, as hyper-parameters that are obtained via the evidence function optimization. Due to the accuracy-complexity trade-off, the variances  of the parameters that need to be pushed to zero (i.e., do not contribute much to the accuracy-likelihood term) get very large values and their corresponding means get values close to zero, see e.g., \cite{Bishop2006, wipf2004sparse}, where a theoretical justification is provided. The key point here is that allowing the parameters to vary independently, with different variances, unveils specific relevance of every individual parameter to the observed data,  and the ``irrelevant'' ones are pushed to zero with high probability. Such methods are also known as \emph{Automatic Relevance Determination} (ARD). In Section \ref{subsec:SA-GP}, we demonstrate the use of ARD philosophy for designing recent sparse kernels. 

\emph{Remark 1:} In practice, the choice of a specific prior depends on the trade-off between the expressive power of the prior and the difficulty of inference. As shown in Table~\ref{tab2}, advanced sparsity-promoting priors, e.g., the generalized hyperbolic prior and the horseshoe prior, come with more complicated mathematical expressions. These endue the priors flexibility  to adapt to different levels of sparsity, while also pose difficulty in deriving efficient inference algorithm.  Typically, when the noise power is known to be small, and/or the side information about the sparsity level is available,  sparsity-promoting priors with simple mathematical forms, e.g., the Student's $t$-prior, are recommended. Otherwise, one might consider the adoption of more complex members in the family of GSM priors, see e.g., \cite{ghosh2019model,cheng2022towards}.

\subsection{SAL via Bayesian Methods for Non-Parametric Models}
\label{subsec:SAL-bayes-nonparametric}
%
{\color{black}In this subsection, we turn our focus on non-parametric models, where the number of the involved parameters in an adopted model is not considered to be known and fixed by the user but, in contrast,  it has to be learnt  from the data during training. A common path to this direction is to assume that the involved number of parameters is infinite (practically a very large number) and then leave it to the algorithm to recover a finite set of parameters out of the, initially, infinite ones. To this end, one has to involve priors that deal with infinite many parameters. The ``true" number of parameters is recovered via the associated posteriors.

\subsubsection{Indian Buffet Process (IBP) Prior} 
\label{subsec:ibp_prior}
\begin{figure}[!t]
\centering
\includegraphics[width= 6 in]{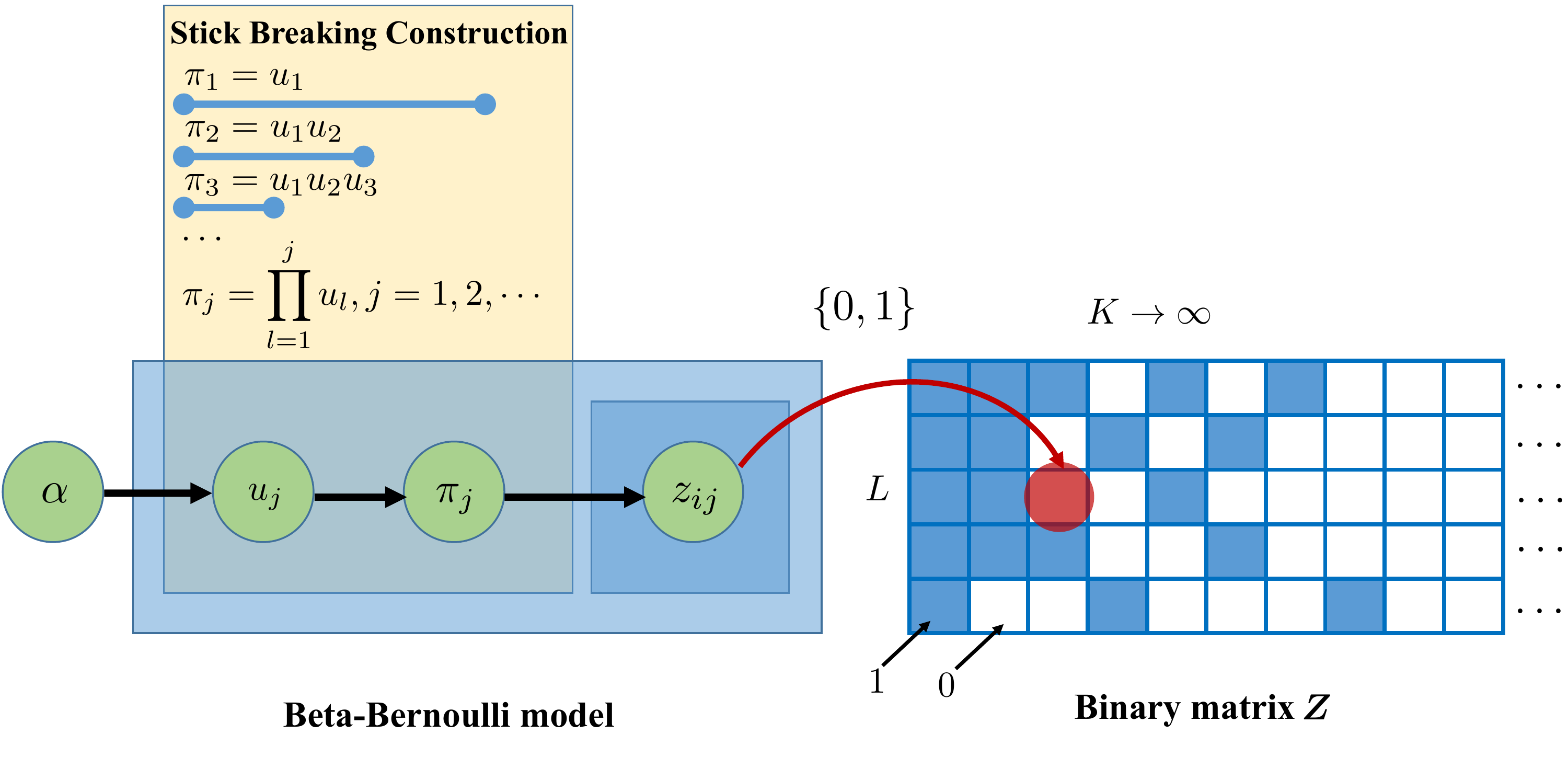}
\caption{Illustrative implementation of IBP via stick breaking construction.}
\label{fig5}
\end{figure}

We will introduce the IBP in a general formulation, and then we will see how to adapt it to fit our needs in the context of designing DNNs. Let us first assume that an Indian restaurant offers a very large number, $K$, of dishes and let $K\rightarrow \infty$. There are $L$ customers. The first one selects a number of dishes, with some probability. The second customer, selects some of the previously selected dishes with some probability and some new ones with another probability, and so on till all, $L$, customers have been considered. In the context of designing NNs, customers are replaced by the dimensions of the input (to each one of the layers) vector, and the infinite many dishes by the number of nodes in a layer. Since we have assumed that the architecture is unknown, that is, the number of nodes (neurons) in a layer, we consider infinite many of those. Then, following the rationale of IBP, the first dimension, say, $x_1$ is linked to some of the infinite nodes, with certain probabilities, respectively. Then, the second dimension, say, $x_2$ is linked to some of the previously linked nodes and to some new ones, according to certain probabilities. This reasoning carries on, till the last dimension of the input vector, $x_L$, has been considered.  As we will see soon, the IBP is a sparsity promoting prior, because out of the infinite many nodes, only a small number of those is {\it probabilistically} selected. 

In a more formal way, we adopt a binary random variable, $z_{ij}\in \{0,1\}$, $i=1,2,\ldots,L$ and $j=1,2,\ldots$. If $z_{ij}=1$, the $i$-th customer ($i$-th dimension) selects the $j$-th dish (is linked to the $j$-th node). On the contrary, if $z_{ij}=0$, the dish is not selected, (the $i$-th dimension is not linked to the $j$-th node). The binary matrix  $\boldsymbol Z$ that is defined from the elements $z_{ij}$ is an infinite dimensional one and the IBP is a prior that promotes zeros in such binary matrices.

}

One way to implement the IBP 
is via the so-called {\it stick breaking construction}. The goal is to populate an infinite binary matrix, $\boldsymbol Z$, with each element being zero or one. To this end, we first generate {\it hierarchically} a sequence of, theoretically, infinite probability values, $\pi_j,~j=1,2,\cdots$. To achieve this, the Beta distribution is mobilized. The Beta distribution, e.g., \cite{theodoridis2020machine}, is defined in terms of two parameters. For the IBP, we fix one of them to be equal to $1$ and the other one, $\alpha$, is left as a (hyper-)parameter, which can either be pre-selected or learnt  during training. Then, the following steps are in order:
\begin{align}
u_j\sim \text{Beta}(u_j|\alpha,1), ~~\pi_j=\prod_{l=1}^j u_l,~j=1,2,\cdots,
\end{align}
where, the notation $\sim$ indicates the sample drawn from a distribution. Then, the generated probabilities, $\pi_j$, are used to populate the matrix $\bm{Z}$, by drawing samples from a Bernoulli distribution, see e.g., \cite{theodoridis2020machine}, that generates an one, with probability $\pi_j$ and a zero with probability $1-\pi_j$, as 
\begin{align}
z_{ij}\sim \text{Bernoulli}(z_{ij}|\pi_j),
\end{align}
for each $i=1,2,\cdots,L$, as illustrated in Fig.~\ref{fig5}. The Beta distribution generates numbers between $[0,1]$, and from the above construction it is obvious that the sequence of probabilities $\{ \pi_j\}$ goes rapidly to zero, due to the product of quantities $\{ u_l \}$ being less than one in magnitude. How fast this takes place is controlled by $\alpha$, which is known as the innovation or strength parameter of the IBP, see e.g., \cite{theodoridis2020machine}.

\section{The Art of Prior: Sparsity-Aware Modeling for Three Case Studies}
\label{sec:prior-with-recent-tools}
In the previous section, we have introduced the indispensable ingredients for obtaining sparsity-aware modeling under the Bayesian learning framework, namely the priors. In this section, we will demonstrate how these priors can be incorporated into some popular data modeling and analysis tools to achieve sparsity-promoting properties. Concretely, we will introduce sparsity-aware modeling for \emph{Bayesian deep neural networks} in Section~\ref{subsec:SA-BNN}, for Gaussian processes in Section~\ref{subsec:SA-GP}, and for tensor decompositions in Section~\ref{subsec:SA-tensor}. 

\subsection{Sparsity-Aware Modeling for Bayesian Deep Neural Networks} 
\label{subsec:SA-BNN}
%

Our focus in this section is to deal with sparsity-promoting techniques in order to prune DNNs. That is, starting from a network with a large number of nodes, to optimally remove nodes and/or links. We are going to follow both paths, namely the parametric one via the GSM priors and the non-parametric one via the IBP prior.

\subsubsection{Fundamentals of DNNs} Neural networks are learning machines that comprise a large number of neurons, which are connected in a layer-wise fashion. After 2010 or so, neural networks with many (more than three) hidden layers, known as \emph{deep neural networks}, have dominated the field of machine learning due to their remarkable representation power and the outstanding prediction performance for various learning tasks. Since their introduction, one of the major tasks associated with their design has been the so-called \emph{pruning}. That is, to remove redundant nodes and links so that their size and, hence, the number of the involved parameters is reduced. Of course, this is another name of what we have called  ``sparsification'' of the network. Over the years, a number of rather ad-hoc techniques have been proposed, see e.g., \cite{theodoridis2020machine, Bishop2006} for a review. In the most recent years, Bayesian techniques have been employed in a more formal and theoretically pleasing way. These techniques comprise our interest in this article.  We will focus on the vanilla deep fully connected networks, yet, such techniques have been extended and can be used for the case of, e.g., convolutional networks \cite{panousis2019nonparametric, panousis21a}.
\begin{figure}[!t]
\centering
\includegraphics[width= 6 in]{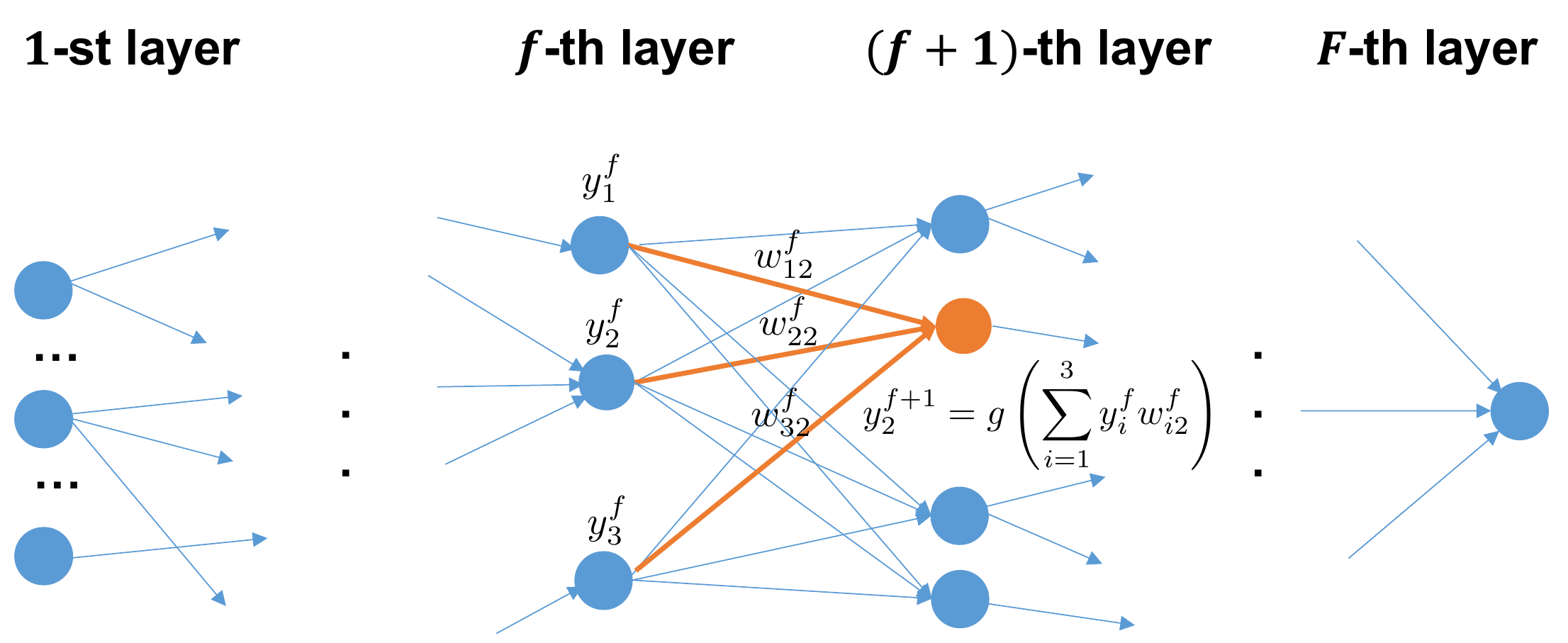}
\caption{Illustration of a deep fully connected network.}
\label{fig6}
\end{figure}

The name ``deep fully connected networks'' stresses out that each node in any of the layers is directly connected to every node of the previous layer. To state it in a more formal way,  without loss of generality, we consider a deep  fully connected network consisting  of $F$ layers.  The number of nodes in the $f$-th ($1 \leq f \leq F-1 $) layer is $a^{f}$. \footnote{\color{black}Note that $f$ in $a^f$ stands for the $f$-the layer and acts as a superscript. It does not denote $a$ to the power of $f$.} For the $i$-th node in the $f$-th layer and the $j$-th node in the $(f+1)$-th layer, the link between them has a weight $w_{ij}^{f}$, as illustrated in Fig.~\ref{fig6}. The input vector to the $(f+1)$-th layer consists of the outputs from the previous layer, denoted as $\boldsymbol y^f = [y^f_1, y^f_2, \cdots, y^f_{a^{f}} ]^T$, where $y^f_i$ is the output at the  $i$-th node. Denote $\boldsymbol w^f_j = [ w_{1j}^{f}, w_{2j}^{f}, \cdots, w_{a^{f} j}^{f} ]^T$ the vector that collects all the link weights associated with the  $j$-th node. Then the output of the $j$-th node is:
\begin{align}
y^{f+1}_j =  g \left(\sum_{i=1}^{a^{f}}  w_{ij}^{f} y^f_i \right) = g\left(\left[\boldsymbol w^f_j \right]^T \boldsymbol y^f \right) , 
\label{eq12}
\end{align}
where $g(\cdot)$ is a non-linear transformation function (also called activation function), and the most widely used ones include the rectified linear unit (ReLU) function, the sigmoid function, and the hyperbolic tanh function, see e.g., [1].


\subsubsection{Sparsity-Aware Modeling Using GSM Priors} The basic idea of this approach can be traced back to the pioneering work \cite{mackay1995probable} of D. J. MacKay in 1995. He pointed out that for a neural network with a single hidden layer, the weights, each associated with a link between two nodes, can be treated as {\it random variables}. The connection weights are associated with zero-mean Gaussian priors, typically with a shared variance hyper-parameter. Then, appropriate (e.g., Gaussian) posteriors are learnt over the connection weights, which can be used for inference at test time. The variance hyper-parameters of the imposed Gaussian priors can be selected to be low enough, so that the corresponding connection weights exhibit \textit{a priori} the tendency of being concentrated around the postulated zero mean. This induces a sparsity ``bias'' to the network. The major differences between the recent works \cite{louizos2017bayesian, ghosh2019model} and the early work \cite{mackay1995probable} lies in their adopted priors. 
\begin{figure}[!t]
\centering
\includegraphics[width= 6.5 in]{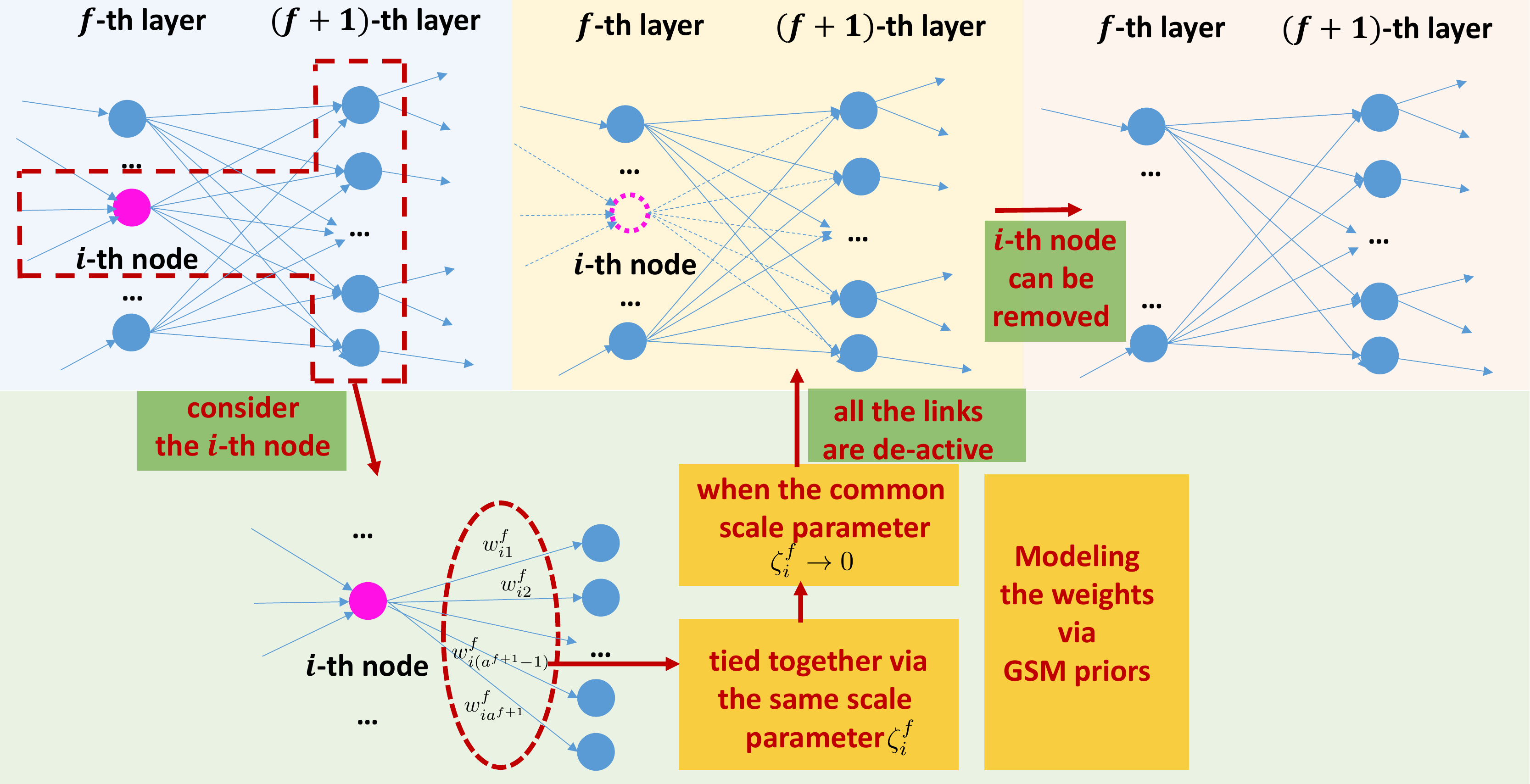}
\caption{Illustration of node-wise sparsity-aware modeling for DNNs using GSM priors.}
\label{fig7}
\end{figure}

Let us consider a network with multiple hidden layers \cite{louizos2017bayesian, ghosh2019model}, as illustrated in Fig.~\ref{fig7}. For the $i$-th node in the $f$-th layer and the $j$-th node in the $(f+1)$-th layer, their link has a weight $w_{ij}^{f}$. For each of the random weights, we can adopt a sparsity-promoting GSM prior so that
\begin{align}
p(w_{ij}^f)  = \int \mathcal{N}(w_{ij}^f ; 0,  \zeta_{ij}^f) p(\zeta_{ij}^f;  \boldsymbol \eta) d\zeta_{ij}^f,
\label{eq13}
\end{align}
in which each functional form of $p(\zeta_{ij}^f; \boldsymbol{\eta})$ in Table~\ref{tab2} corresponds to a GSM prior. Particularly, the Normal-Jeffreys prior and the horseshoe prior were used in \cite{louizos2017bayesian, ghosh2019model}.  Next, we show how to conduct node-wise sparsity-aware modeling (for all the weights connected to that node). Inspired by the idea reported in \cite{mackay1995probable}, we group the weights $\{w_{ij}^{f}\}_{j=1}^{a^{f+1}}$ connected to the $i$-th node, and assign a common scale parameter $\zeta_i^f$ to their GSM priors, i.e., $\zeta_{ij}^f = \zeta_i^f, \forall j$. Then we have the prior modeling for the $i$-th node related weights $\{w_{ij}^{f}\}_{j=1}^{a^{f+1}}$:  
\begin{align}
p(\{w_{ij}^{f}\}_{j=1}^{a^{f+1}}) &= \int p(\{w_{ij}^{f}\}_{j=1}^{a^{f+1}}, \zeta_{i}^f  ) d \zeta_{i}^f \nonumber \\
& =  \int p(\{w_{ij}^{f}\}_{j=1}^{a^{f+1}} | \zeta_{ij}^f ) p(\zeta_i^f;  \boldsymbol \eta) d\zeta_{i}^f  \nonumber \\
& = \int   \prod_{j=1}^{a^{f+1} } \mathcal{N}(w_{ij}^f ; 0, \zeta_i^f) p(\zeta_i^f;  \boldsymbol \eta) d\zeta_i^f.
\label{eq14}
\end{align}
Furthermore, assuming that the nodes in the $f$-th layer are mutually  independent, we obtain the prior modeling for all the weights $ \{\{w_{ij}^{f}\}_{i=1}^{a^{f}}\}_{j=1}^{a^{f+1}}$ forwarded from the $f$-th layer:
\begin{align}
p(\{\{w_{ij}^{f}\}_{i=1}^{a^{f}}\}_{j=1}^{a^{f+1}}) = \prod_{i=1}^{a^{f}} p(\{w_{ij}^{f}\}_{j=1}^{a^{f+1}}) =  \prod_{i=1}^{a^{f}}  \int  \prod_{j=1}^{a^{f+1}}  \mathcal{N}(w_{ij}^f;  0, \zeta_i^f) p(\zeta_i^f;  \boldsymbol \eta) d\zeta_i^f.
\label{eq15}
\end{align}
By this modeling strategy, the weights $\{w_{ij}^{f}\}_{j=1}^{a^{f+1}}$ related to the $i$-th node are tied together in the sense that when $\zeta_i^f$ (a single scalar value) goes to zero, the associated weights $\{w_{ij}^{f}\}_{j=1}^{a^{f+1}}$  all become negligible. This makes the $i$-th node in the $f$-th layer disconnected to the $(f+1)$-th layer, and thus blocks the information flow. Together with the sparsity-promoting nature of GSM priors, the prior derived in \eqref{eq15} inclines a lot of nodes to be removed from the DNN without degrading the data fitting performance. This leads to the node-wise sparsity-aware modeling for deep fully connected neural networks. Of course, as it is always the case with Bayesian  learning, the goal is to learn the corresponding posterior distributions, and node removal is based on the learnt values for the respective means and variances.
\begin{figure}[!t]
\centering
\includegraphics[width= 6.5 in]{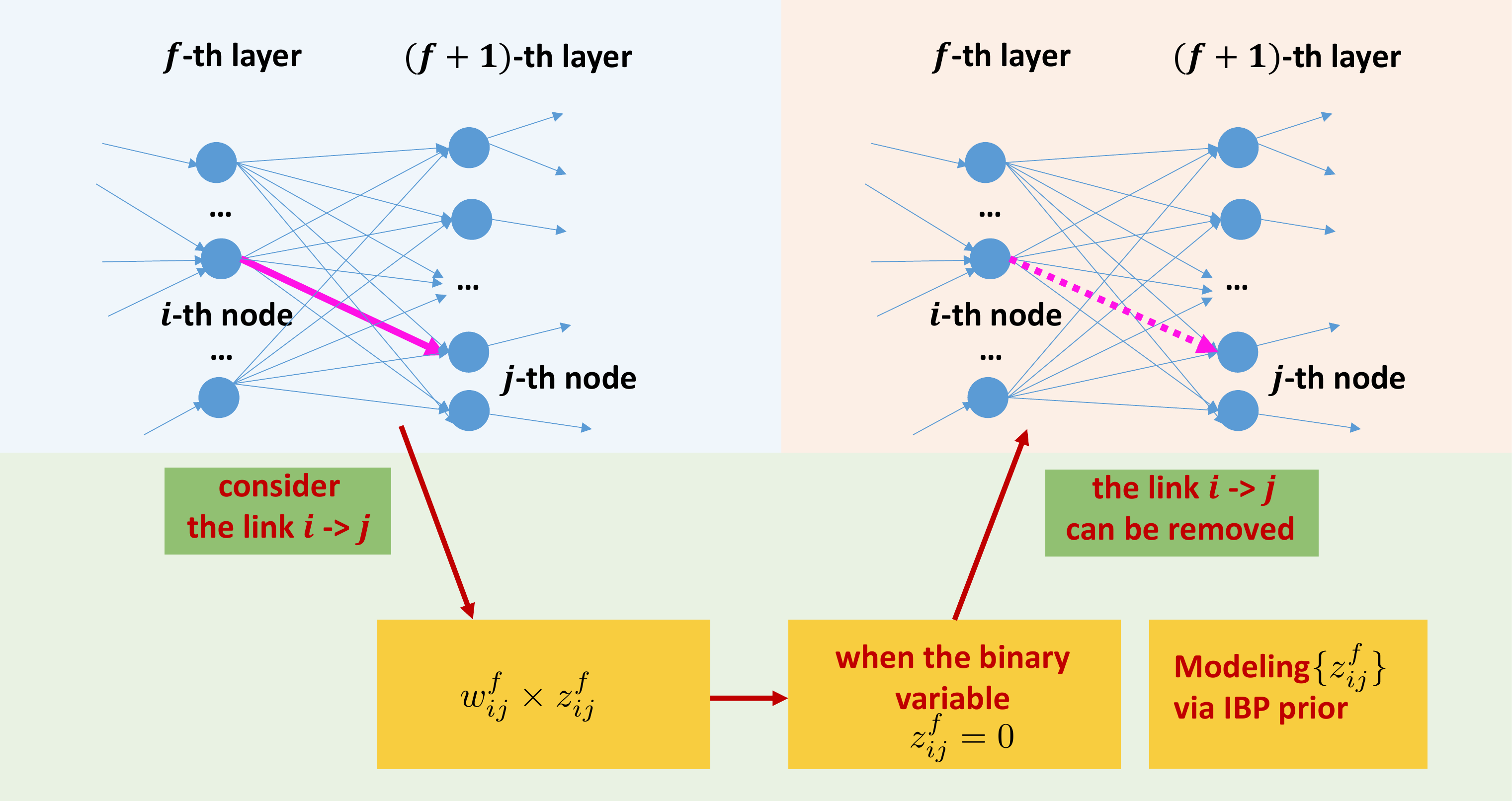}
\caption{Illustration of link-wise sparsity-aware modeling for deep neural networks using IBP prior.}
\label{fig8}
\end{figure}

\subsubsection{Sparsity-Aware Modeling Using IBP Prior}
\label{subsubsec:link_wise_ibp} 
The previous approach on imposing sparsity inherits a  major drawback  that is shared by all  techniques for designing DNNs. That is, the number  of nodes per layer has to be specified and pre-selected. Of course, one may say that we can choose a very large number of nodes and then harness ``sparsity''  to prune the network. However, if one overdoes it, he/she soon runs into problems due to over-parameterization. In contrast, we are now going to turn our attention to non-parametric techniques. We are going to assume that the nodes per layer are theoretically infinite (in practice a large enough number) and then use the IBP prior to enforce sparsity.  

\textcolor{black}{In line with  what has been said while introducing the IBP (Section\ref{subsec:ibp_prior}), we multiply each weight, i.e., $w_{ij}^{f}$, with a corresponding auxiliary (hidden) binary random variable,  $z_{ij}^f$.} 
%
%
The required priors for these variables,  $\{\{z_{ij}^f\}_{i=1}^{a^{f}}\}_{j=1}^{a^{f+1}}$ are generated via the IBP prior. 
In particular, we define a binary matrix $\boldsymbol Z^f \in \mathbb R^{a^f \times a^{f+1}}$, with its $(ij)$-th element being $z_{ij}^f$ for the $f$-th layer. Due to the sparsity-promoting nature of IBP prior, most elements in $\boldsymbol Z^f $ tend to be zero, nulling the corresponding weights in $\{\{w_{ij}^{f}\}_{i=1}^{a^{f}}\}_{j=1}^{a^{f+1}}$, due to the involved  multiplication. 
This leads to an alternative sparsity-promoting modeling for DNNs \cite{panousis2019nonparametric, panousis21a}. 

The \textit{stick breaking construction} for the IBP prior was utilized since it turns out to be readily amenable to variational inference. This is a desirable property that facilitates both training and inference through recent advances in black box variational inference, namely stochastic gradient variational Bayes, as explained in Section \ref{subsec:BNN-inference}. For each $i$, the considered hierarchical construction reads as follows:  
\begin{align}
	u_j^f \sim \text{Beta}(u^f_j|\alpha,1), \quad \pi_j^f = \prod_{l=1}^j u^f_l, \quad z_{ij}^f \sim \text{Bernoulli}(z_{ij}^f|\pi_j^f).
\end{align}
During training, {\it posterior} estimates of the respective probabilities are obtained, which then allow for a naturally-arising \textit{component omission (link  pruning) mechanism} by introducing a \textit{cut-off threshold} $\tau$; any link/weight with inferred posterior below this threshold value is deemed unnecessary and can be safely omitted from computations. This inherent capability renders the considered approach a {\it  fully  automatic, data-driven, principled paradigm} for sparsity-aware learning based on explicit inference of component utility based on \textit{dedicated latent variables}.

By utilizing the aforementioned construction, we can easily incorporate the IBP mechanism in conventional ReLU-based networks and perform inference. However, the flexibility of the link-wise formulation allows us to go one step further. 

In recent works, the stick-breaking IBP prior has been employed in conjunction with a radically different, biologically-inspired and competition-based activation, namely the stochastic local winner-takes-all (LWTA) \cite{panousis2019nonparametric, panousis21a}. In the general LWTA context, neurons in a conventional hidden layer are replaced by LWTA blocks comprising \textit{competing linear units}. In other words, each node comprises a set of linear (inner product) units.  When presented with an input, each unit in each block computes its activation; the unit with the \textit{strongest} activation is deemed to be the \textit{winner} and passes its output to the next layer, while the rest are inhibited to silence, i.e., the zero value. This is how non-linearity is achieved. 

\begin{figure}
	\begin{center}
		\includegraphics[width= 5.5 in]{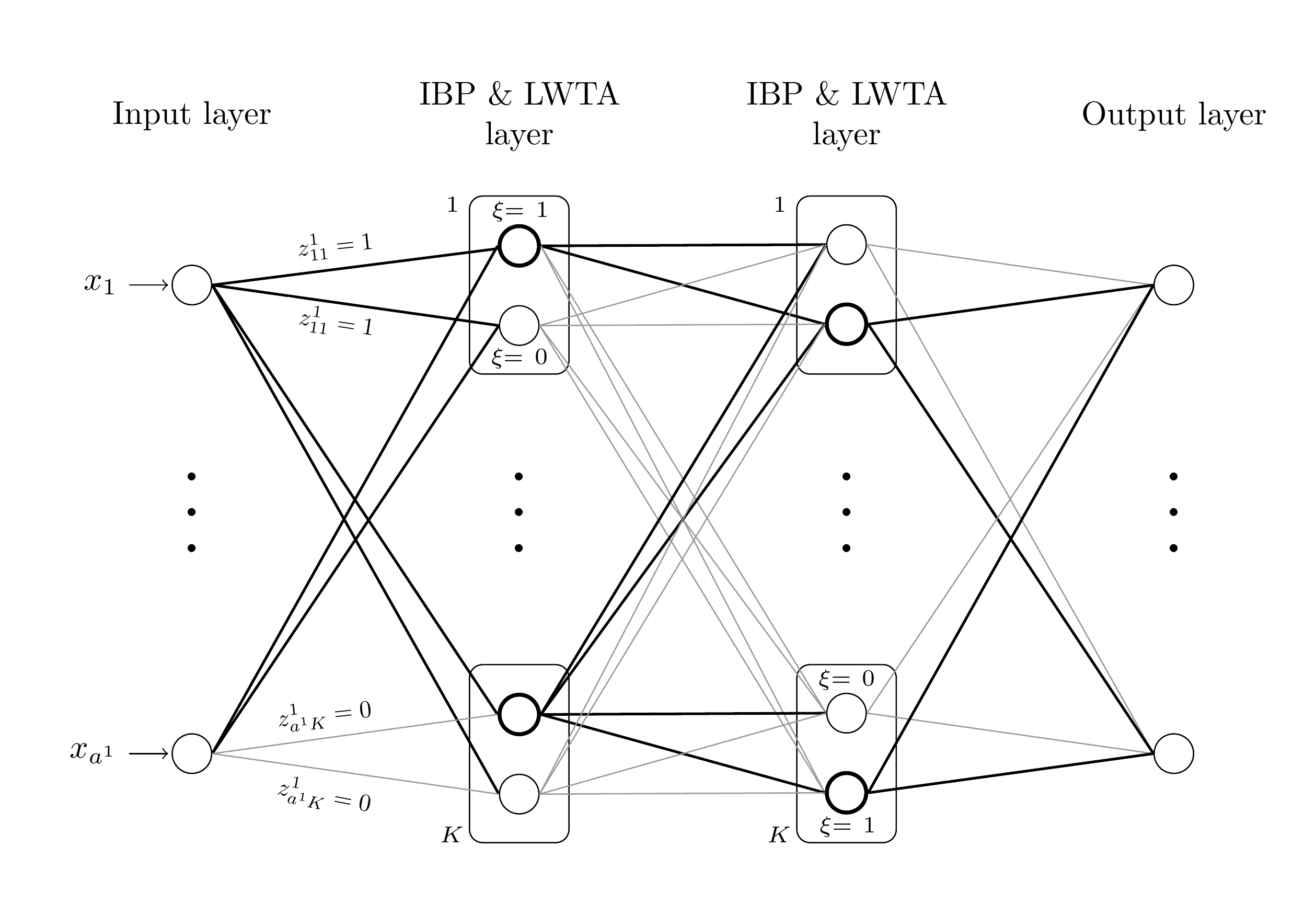}
	\end{center}
	\caption{A graphical illustration of the LWTA and IBP-based architecture. Bold edges denote active (effective) connections (with $z^f_{ik}=1$); nodes with bold contours denote winner units, i.e.,  that correspond to $\xi=1$ (we do not use ${\xi}^f_{kj}$ to unclutter notation); rectangles denote LWTA blocks. For simplicity in the figure, each LWTA block comprises two ($J=2$) competing linear units, $k=1,2,\dots,K$.}
	\label{fig:wta1}
\end{figure}

This deterministic winner selection, known as {\it hard} LTWA, is the standard form of an LTWA. However in \cite{panousis2019nonparametric}, a new variant was proposed to replace the hard LTWA  by a novel \textit{stochastic} adaptation of the competition mechanism implemented via a competitive random sampling procedure founded on Bayesian arguments. {\color{black} To be more specific, let a layer in the NN that comprises $a^f=L$ inputs, i.e., $x_i,~i=1,2,\ldots L$, where we use $\bm{x}$ to denote the input to any layer in order to simplify the discussion.  Also, assume that the number of LWTA blocks in the layer is $a^{f+1}=K$.  We also relax the notation on the number of layer $f$, and our analysis refers to any node of any layer. Each LTWA block comprises $J$ linear units, each one associated with a corresponding weight, $w_{ikj},~i=1,2,\ldots L,~k=1,2,\ldots,K, ~j=1,2\ldots, J$. Consider the $k$-th LTWA block. We introduce an auxiliary {\it latent}  variable, $\xi_{kj}$, and the output of the corresponding $j$-th linear unit in the $k$-th block is given by,
\begin{equation}
y_{kj}=\xi_{kj} \bm{w}^T_{kj}\bm{x}=\xi_{kj}\sum_{i=1}^Lw_{ikj}x_i, ~\xi_{kj}\in\{0,1\},~\sum_{j=1}^J \xi_{kj}=1.\label{sergios1}
\end{equation}
In other words, the outputs of the linear units are either the respective inner product between the input vector and the associated weight vector or zero, depending on the value of $\xi_{kj}$, which can be either zero or one. Furthermore, only one of the $\xi$'s in a block can be one, and the rest are zero. Thus, we can associate with each LWTA block, a vector $\bm{\xi}_k\in \mathbb R^J$, with only one of its elements being one and the rest being zero,  see Fig.~\ref{fig:wta1}. In the ML jargon, this is known as one-hot vector and can be denoted as $\bm{\xi}_k\in \mathrm{one\_hot}(J)$.  If we stack together  all the $\bm{\xi}_k,~k=1,2,\ldots,K$, for the specific layer  together, we can write, $\bm{\xi}\in \mathrm{one\_hot}(J)^K$ (strictly speaking $\bm{\xi}^{a^{f+1}}\in \mathrm{one\_hot}(J)^K$).

In the stochastic LTWA, all  $\xi$'s are treated as binary random variables. The respective probabilities, which control the firing (corresponding $\xi=1$) of each linear unit within a single LTWA,  are computed via a {\it softmax} type of operation, see e.g., \cite{bridle1989training},\cite{theodoridis2020machine},  that is, 
\begin{equation}
P_{nkj}=\frac{\exp(h_{nkj})}{\sum_{j=1}^J\exp(h_{nkj})},~~h_{nkj}=\sum_{i=1}^L(z_{ik}w_{ikj})x_{ni}.\label{sergios2}
\end{equation}

Note that in the above equation, the {\it firing probability} of a linear unit depends on both  the input, $\bm{x}_n$, and  on whether the link to the corresponding LWTA block is active or not (determined by the value of the corresponding utility variable $z_{ik}$). Basically, the stochastic  LWTA introduces a {\it lateral} competition among units in the same layer. 
How the $w$'s as well as the corresponding utility binary variables are learnt is provided in Section \ref{subsec:BNN-inference}.
A graphical illustration of the considered approach is depicted in Fig. \ref{fig:wta1}. Note that as the input changes,  a  {\it different subnetwork}, via different connected links, may be followed,  to pass the input information to the output, with high probability. This is how nonlinearity is achieved in the context of the stochastic LWTA blocks.}

\subsection{Sparsity-Aware Modeling for GPs} 
\label{subsec:SA-GP}
We already discussed in Section~\ref{subsec:gp} that the kernel function determines, to a large extent, the expressive power of the GP model. More specifically, the kernel function profoundly controls the characteristics (e.g., smoothness and periodicity) of a GP. In order to provide a kernel function with more expressive power and adaptive to any given dataset,
%
%
one way is to expand the kernel function as a linear combination of $Q$ \emph{subkernels/basis kernels}, i.e., 
\begin{equation}
k(\boldsymbol{x},\boldsymbol{x}' ) = \sum_{i=1}^Q \alpha_{i} k_{i}(\boldsymbol{x}, \boldsymbol{x}'),
\label{eq:GP-LMK}
\end{equation}
where the weights, $\alpha_i$, $i=1,2,\cdots,Q$, can either be set manually or be optimized. Each one of these subkernels can be any one of the known kernels or any function that admits the properties that define a kernel, see e.g., \cite{Bishop2006}. One may consider to construct such a kernel either in the original input domain or in the frequency domain. The most straightforward way is to linearly combine a set of elementary kernels, such as the SE kernel, rational quadratic kernel, periodic kernel, etc., with varying kernel hyper-parameters in the original input domain, see e.g., \cite{RW06, DLG13, Xu19}. For high-dimensional inputs, one can first detect pairwise interactions between the inputs and for each interaction pair adopt an elementary kernel or an advanced deep kernel \cite{dai2020interpretable}. Such resulting kernel belongs essentially to the Analysis-of-Variance (ANOVA) family as surveyed in \cite{theodoridis2020machine}, which has a hierarchical structure and \emph{good interpretability}. Alternatively, one may perform optimal kernel design in the frequency domain by using the idea of sparse spectrum kernel representation. Due to its solid theoretical foundation, in this paper, we will focus on the sparse spectrum kernel representation and review some representative works, such as \cite{Gredilla10, WA13, yin2020linear}, at the end of this subsection. 

\subsubsection{Rationale behind Sparsity-Awareness} The corresponding GP model with the kernel form in (\ref{eq:GP-LMK}) can be regarded as a linearly-weighted sum of $Q$ independent GPs. In other words, we can assume that the underlying function takes the form $f(\boldsymbol{x}) = \sum_{i=1}^{Q} f_{i}(\boldsymbol{x})$, where $f_{i}(\boldsymbol{x}) \sim \mathcal{GP}(0, \alpha_i k_{i}(\boldsymbol{x}, \boldsymbol{x}'))$, for $i=1,2,\cdots,Q$. In practice, $Q$ is selected to be a large value compared to the ``true'' number of  the underlying effective components that generated the data, whose exact value is not known in practice.  In the sequel, one can mobilize the ARD philosophy (see Section~\ref{subsec:SAL-bayes-parametric}), during the evidence function optimization, to \emph{drive all unnecessary sub-kernels to zero}, namely promoting sparsity on $\boldsymbol{\alpha}$. To this end, let us first establish a bridge between the non-parametric GP model and the Bayesian linear regression model that was considered in Section \ref{subsec:blr-exmaple}. 


From the theory of kernels, see e.g., \cite{theodoridis2020machine, Bishop2006, SC04}, each one of the subkernel functions can be written as the inner product of the corresponding feature mapping function, namely, $k_{i}(\boldsymbol{x}, \boldsymbol{x}') = \phi_{i}^{T}(\boldsymbol{x}) \phi_{i}(\boldsymbol{x}')$, where $\phi_{i}(\boldsymbol{x}): \mathbb{R}^{L} \mapsto \mathbb{R}^{L'}$ and it is often assumed $L' \gg L$. As a matter of fact, the feature mapping function results by fixing one of the arguments of the kernel function and making it a function of a single argument, i.e., $\phi_{i}(\boldsymbol{x}) = k_{i}(\boldsymbol{x}, \cdot)$, where ``$\cdot$'' denotes the free variable(s) of the function and is filled by $\boldsymbol{x}'$ before. In general, $\phi_{i}(\boldsymbol{x})$ is a function. However, in practice, if needed, this can be approximated by a very high dimensional vector constructed via the famous random Fourier feature approximation \cite{Rahimi07},\cite{theodoridis2020machine}.  Then, each independent GP process, $f_{i}(\boldsymbol{x}) \sim \mathcal{GP}(0, \alpha_i k_{i}(\boldsymbol{x}, \boldsymbol{x}'))$, by mobilizing the definition of the covariance matrix, can be equivalently interpreted as $f_{i}(\boldsymbol{x}) \triangleq \boldsymbol{\theta}_{i}^T \phi_{i}(\boldsymbol{x})$, where the weights, $\boldsymbol{\theta}_{i}$, of size $L' \times 1$, are assumed to follow a zero-mean Gaussian distribution, i.e., $\boldsymbol{\theta}_{i} \sim \mathcal{N}( \mathbf{0}, \alpha_i \boldsymbol{I})$. Therefore, one can alternatively  write $f(\boldsymbol{x}) = \sum_{i=1}^{Q} \boldsymbol{\theta}_{i}^T \phi_{i}(\boldsymbol{x})$, where $\boldsymbol{\theta}_{i}$ and $\boldsymbol{\theta}_{j}$ are assumed to be mutually independent for $i \neq j$. \textcolor{black}{Essentially, a Gaussian process with such kernel configuration is a special case of the more general sparse linear model family, which can also incorporate, apart from a Gaussian prior, heavy-tailed priors to promote sparsity such as those surveyed in Section~\ref{subsec:SAL-bayes-parametric} . A more detailed presentation of sparse linear models can be found in some early references, such as \cite{wipf2004sparse, Seeger08}.}

As we mentioned before, the GP model hyper-parameters can be optimized through maximizing the logarithm of the evidence function, $L(\boldsymbol{\eta}) \triangleq \log p(\boldsymbol{y}; \boldsymbol{\eta})$, and using (\ref{eq:GP-evidence}), we obtain:
\begin{align}
\hat{\boldsymbol{\eta}} &= \arg \max_{\boldsymbol{\eta}} \log \mathcal{N}(\boldsymbol{y}; \mathbf{0}, \beta^{-1} \boldsymbol{I} + \sum_{i=1}^{Q} \alpha_{i} \boldsymbol{K}_i(\boldsymbol{X}, \boldsymbol{X}))   \nonumber \\
&\equiv \arg \min_{\boldsymbol{\eta}} \left\lbrace \log \det \left( \beta^{-1} \boldsymbol{I} + \sum_{i=1}^{Q} \alpha_i \boldsymbol{K}_i(\boldsymbol{X}, \boldsymbol{X}) \right) + \boldsymbol{y}^T \left( \beta^{-1} \boldsymbol{I} + \sum_{i=1}^{Q} \alpha_i \boldsymbol{K}_i(\boldsymbol{X}, \boldsymbol{X}) \right)^{-1} \boldsymbol{y} \right\rbrace  \label{eq:evidence-for-gp} \\
&\equiv \arg \min_{\boldsymbol{\eta}} \left\lbrace \log \det \left( \beta^{-1} \boldsymbol{I} + \sum_{i=1}^{Q} \alpha_i \boldsymbol{\Phi}_i(\boldsymbol{X}) \boldsymbol{\Phi}_{i}^{T}(\boldsymbol{X}) \right) + \boldsymbol{y}^T \left( \beta^{-1} \boldsymbol{I} + \sum_{i=1}^{Q} \alpha_i \boldsymbol{\Phi}_{i}(\boldsymbol{X}) \boldsymbol{\Phi}_{i}^{T}(\boldsymbol{X})  \right)^{-1} \boldsymbol{y} \right\rbrace, \label{eq:evidence-for-blr} 
\end{align} 
where $\boldsymbol{\eta} = [\boldsymbol{\alpha}^T, \beta]^T$, and (\ref{eq:evidence-for-gp}) in the second line corresponds to the original GP model, and \eqref{eq:evidence-for-blr} in the third line corresponds to the equivalent Bayesian linear model mentioned above. Note that $\boldsymbol{K}_i(\boldsymbol{X}, \boldsymbol{X})$ represents the $N \times N$ kernel matrix of $k_{i}(\boldsymbol{x}, \boldsymbol{x}')$ evaluated for all the training input pairs, while $\boldsymbol{\Phi}_{i}(\boldsymbol{X}) \triangleq [\phi_{i}(\boldsymbol{x}_1),  \phi_{i}(\boldsymbol{x}_2),\cdots, \phi_{i}(\boldsymbol{x}_N)]^T$, of size $N \times L'$, contains the explicit mapping vectors evaluated at the training data. In the sequel, they are denoted as $\boldsymbol{K}_i$ and $\boldsymbol{\Phi}_{i}$ for brevity. 

Mathematical proof of the sparsity property follows that of the \emph{relevance vector machine} (RVM) \cite{Tipping03fastmarginal} for the classic sparse linear model. Let us focus on the last expression in (\ref{eq:evidence-for-blr}), which involves both the log-determinant and the inverse of the overall covariance matrix, $\boldsymbol{C}$, and mathematically, 
\begin{equation}
\boldsymbol{C} = \beta^{-1} \boldsymbol{I} + \sum_{m=1, m \neq i}^{Q} \alpha_{m} \boldsymbol{\Phi}_{m} \boldsymbol{\Phi}_{m}^{T} + \alpha_{i} \boldsymbol{\Phi}_{i} \boldsymbol{\Phi}_{i}^{T} = \boldsymbol{C}_{-i} + \alpha_{i} \boldsymbol{\Phi}_{i} \boldsymbol{\Phi}_{i}^{T},
\end{equation}
where we have separated the $i$-th subkernel from the rest. For clarity, let us focus on the kernel hyper-parameters, $\boldsymbol{\alpha} = [\alpha_1, \alpha_2,\cdots, \alpha_Q]^T$, namely we regard $\beta$ as known and remove it from $\boldsymbol{\eta}$. Applying the classic matrix identities \cite{RW06}, and inserting the results back to (\ref{eq:evidence-for-blr}), we get $L( \boldsymbol{\alpha} ) = L( \boldsymbol{\alpha}_{-i} ) + \gamma(\alpha_i)$, where $L(\boldsymbol{\alpha}_{-i})$ is simply the evidence function with the $i$-th subkernel removed, and the newly introduced quantity $\gamma(\alpha_i)$ is 
\begin{equation}
 \gamma \left(\alpha_{i}\right) \triangleq -\frac{1}{2} \ln (\alpha_i) -\frac{1}{2} \log \left| \boldsymbol{I} + \alpha_{i} \boldsymbol{\Phi}_{i}^{\top} \boldsymbol{C}_{-i}^{-1} \boldsymbol{\Phi}_{i} \right|+\frac{1}{2} \boldsymbol{y}^{\top} \boldsymbol{C}_{-i}^{-1} \boldsymbol{\Phi}_{i}\left( \alpha_{i}^{-1} \boldsymbol{I} + \boldsymbol{\Phi}_{i}^{\top} \boldsymbol{C}_{-i}^{-1} \boldsymbol{\Phi}_{i}\right)^{-1} \boldsymbol{\Phi}_{i}^{\top} \boldsymbol{C}_{-i}^{-1} \boldsymbol{y}.
 \label{eq:GP-RVM-gamma}
\end{equation}
It is not difficult to verify that the evidence maximization problem in (\ref{eq:evidence-for-blr}) boils down to maximizing the $\gamma \left(\alpha_{i}\right)$ when fixing the rest of the parameters to their previous estimates. This means that we can solve for the hyper-parameters in a sequential manner. Taking the gradient of $\gamma \left(\alpha_{i}\right)$ with respect to the scalar parameter $\alpha_i$ and setting it to zero gives the global maximum as can be verified by its second order derivative. Interestingly, the solution to $\alpha_i$ is either zero or a positive value, mainly depending on the relevance between the $i$-th subkernel function and the observed data \cite{Bishop2006}. Only if their relevance is high enough, $\alpha_i$ will take a non-zero, positive value. This explains the sparsity-promoting rationale behind the method. In Section~\ref{sec:inference}, we will provide an advanced numerical method for solving the hyper-parameters that maximize the evidence function. 

\subsubsection{Sparse Spectrum Kernel Representation} At the beginning of this subsection, we expressed kernel expansion in terms of a number of subkernels and introduced two major paths (either in the original input domain or in the frequency domain). In the sequel, we turn our focus to the frequency domain representation of a kernel function and on techniques that promote sparsity in the frequency domain, leading to the sparse spectrum kernel representation. 

To start with, it is assumed that the underlying function only has a few effective frequency bands/points in the real physical world. Second, the kernel function takes a linearly-weighted sum of basis functions, similar to the ARD method for linear parametric models, thus only a small number of which are supposed to be relevant to the given data from the algorithmic point of view. Sparse solutions can be obtained from maximizing the associated evidence function as will be introduced in Section~\ref{sec:inference}. 

For ease of narration, we constrain ourselves to one-dimensional input space, namely, $x \in \mathbb{R}^1$, but the idea can be easily extended to the multi-dimensional input space. Often, we have $x =t$ for one-dimensional time series modeling.

The earliest sparse spectrum kernel representation was proposed in \cite{Gredilla10} and developed upon a Bayesian linear regression model with trigonometric basis functions, namely, 
\begin{equation}
f(x) = \sum_{i=1}^{Q} a_i \cos(2 \pi \omega_i x) + b_i \sin(2 \pi \omega_i x),
\label{eq:trigonometric-basis-functions}
\end{equation}
where $\{ \cos(2 \pi \omega_i x), \sin(2 \pi \omega_i x) \}$ constitute one pair of basis functions parameterized in terms of the center frequencies $\omega_i$, $i \in \{ 1,2,\cdots,Q\}$, and the random weights, $a_i$ and $b_i$ are independent and follow the same Gaussian distribution, $\mathcal{N}(0, \sigma^2_{0}/Q)$.\footnote{It is noteworthy that $\omega \in [0, 1/2)$ represents a normalized frequency, namely the physical frequency over the sampling frequency.} Under such assumptions, $f(x)$ can be regarded as a GP according to Section~\ref{subsec:gp}, and the corresponding covariance/kernel function can be easily derived as 
\begin{equation}
k(x,x') = \frac{\sigma_{0}^2}{Q} \boldsymbol{\phi}^{T}(x) \boldsymbol{\phi}(x') = \frac{\sigma_{0}^2}{Q} \sum_{i=1}^{Q} \cos(2\pi \omega_{i}(x-x')),
\label{eq:sparse-spectrum-kernel}
\end{equation}
where the feature mapping vector $\boldsymbol{\phi}(x)$ contains all $Q$ pairs of trigonometric basis functions. Usually, we favor a large value of $Q$, well exceeding the expected number of effective components. If the frequency points are \textit{randomly} sampled from the underlying spectral density, denoted by $\tilde{S}(\omega)$, then (\ref{eq:sparse-spectrum-kernel}) is equivalent to the random Fourier feature approximation of a stationary kernel function \cite{Rahimi07}. However in \cite{Gredilla10}, the center frequencies are optimized through maximizing the evidence function. As it is claimed in the paper, such additional flexibility of the kernel obtained through optimization can significantly improve the fitting performance as it enables automatic learning of the best kernel function for any specific problem. The resulting spectral density of (\ref{eq:sparse-spectrum-kernel}) is a set of \textit{sparse Dirac deltas} for approximating the underlying spectral density. 

\begin{figure} [!t]
\setcounter{subfigure}{0}
\centering
\subfigure[] {
\includegraphics[width=3.0 in]{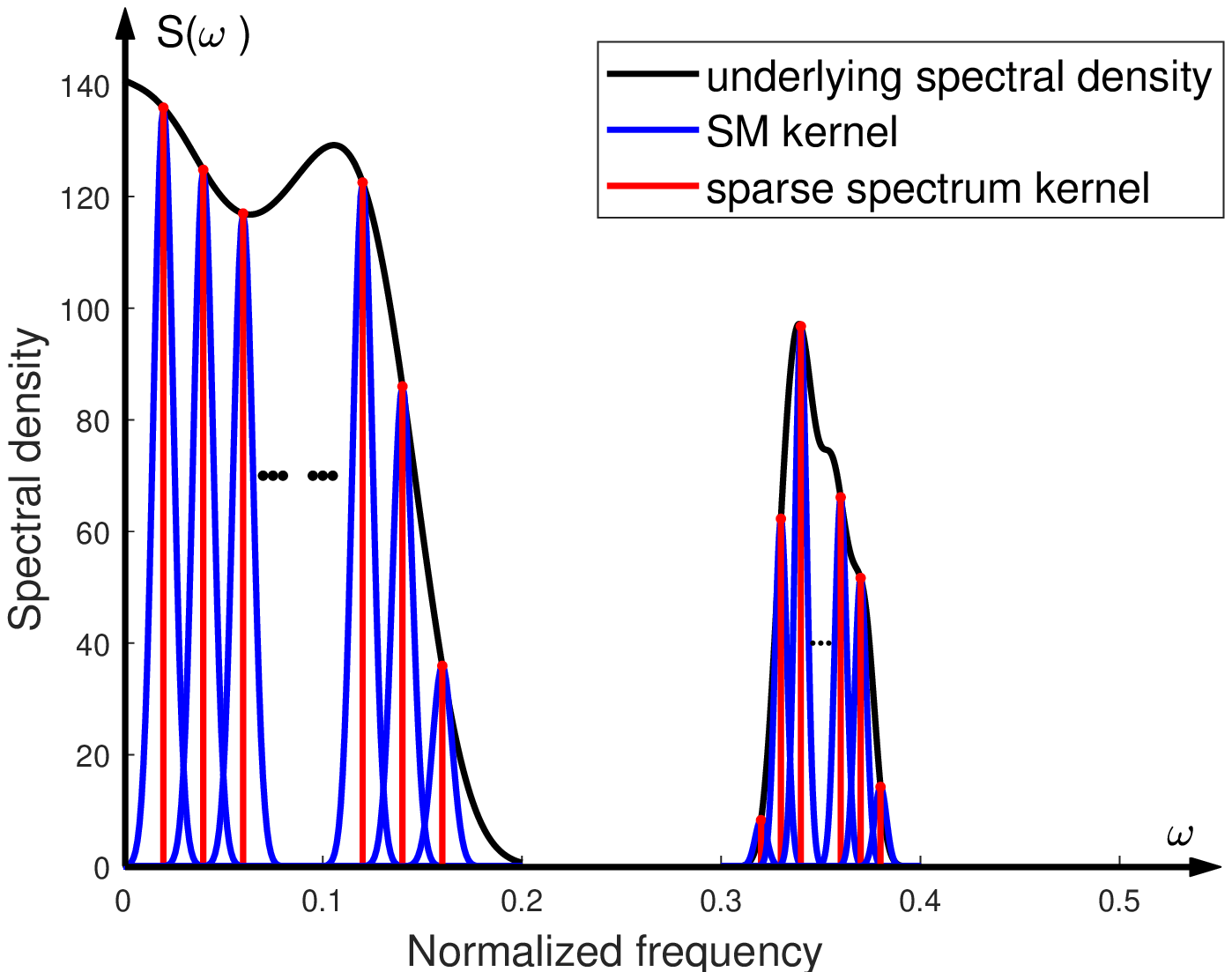}
}
\subfigure[] {
\includegraphics[width=3.0 in]{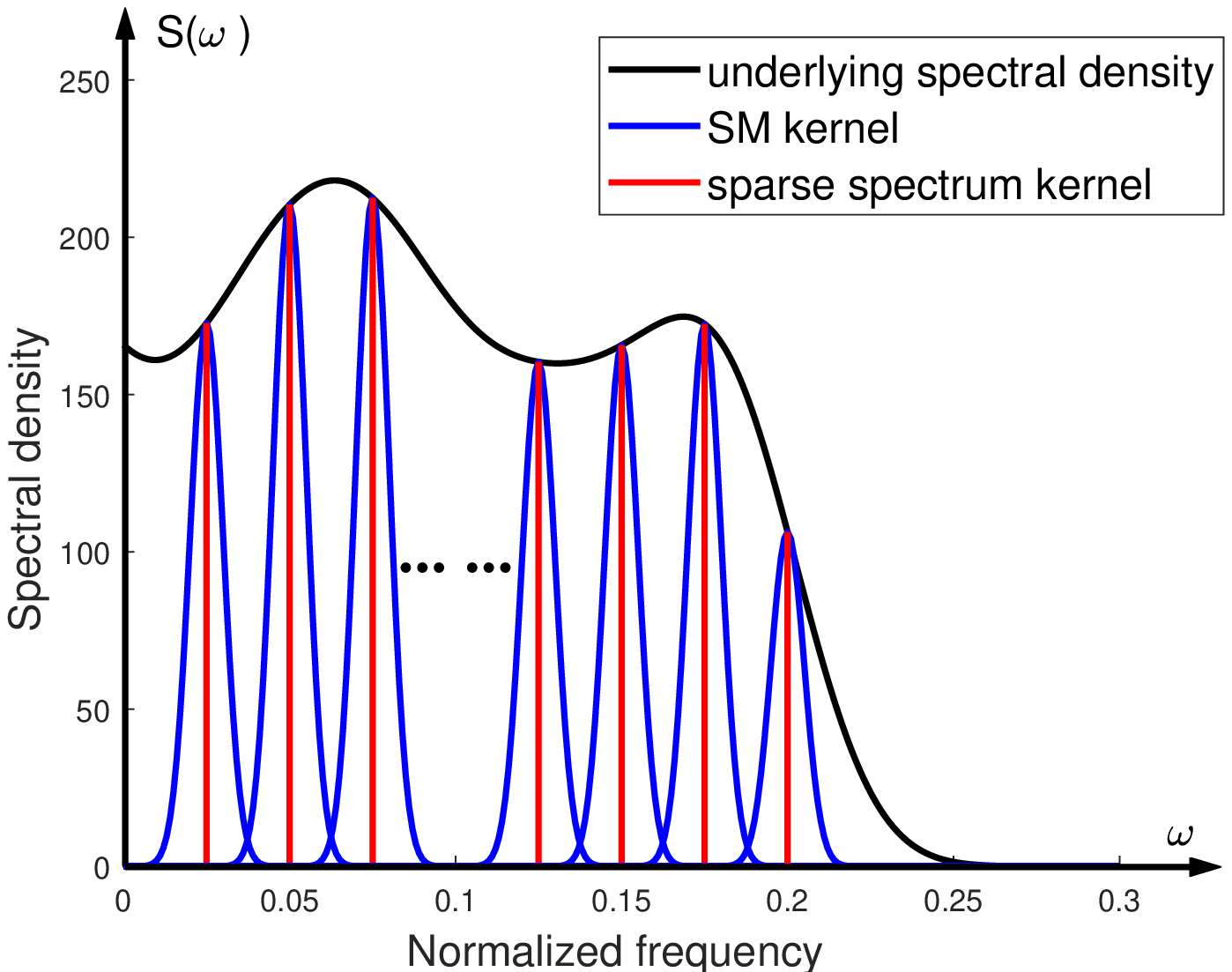}
}
\caption{Comparison between the SM kernel and the original sparse spectrum kernel in (\ref{eq:sparse-spectrum-kernel}) for approximating the underlying spectral density. Herein, the SM kernel employs a mixture of Gaussian basis functions (see the blue curves), while the sparse spectrum kernel employs a mixture of Dirac deltas (see the red vertical lines).}
\label{fig:sparse-GP-kernels}
\end{figure}

In \cite{WA13}, the Dirac deltas are replaced with a \emph{mixture of Gaussian} basis functions in the frequency domain, leading to the so-called \emph{spectral mixture} (SM) kernel. The SM kernel can approximate any stationary kernel arbitrarily well in the $\ell_1$ norm sense due to the Wiener's theorem of approximation \cite{Achieser92}. Concretely, \textcolor{black}{the underlying spectral density} is approximated by a Gaussian mixture as
\begin{equation}
S(\omega) = \frac{1}{2} \sum_{i=1}^{Q} \frac{\alpha_i}{\sqrt{2 \pi \sigma_{i}^2}} \left\lbrace \exp \left[ \frac{-( \omega - \mu_i)^2}{2 \sigma_{i}^{2}} \right] + \exp \left[ \frac{-( \omega + \mu_i)^2}{2 \sigma_{i}^{2}} \right] \right\rbrace ,
\label{eq:GMapproximatedSPD}
\end{equation}
where $Q$ is a fixed number of mixture components, and $\alpha_i$, $\mu_i$, $\sigma_{i}^{2}$ are the weight, mean and variance parameters of the $i$-th mixture component, respectively. It is noteworthy that the sum of the two exponential functions on the right-hand-side of (\ref{eq:GMapproximatedSPD}) ensure the symmetry of the spectral density. For illustration purpose, we draw the comparison between the original sparse spectrum kernel \cite{Gredilla10} and the SM kernel \cite{WA13} in Fig.~\ref{fig:sparse-GP-kernels}. 

Taking the inverse Fourier transform of $S(\omega)$ yields a stationary kernel in the time-domain as follows:
\begin{equation}
k(t, t'; \boldsymbol{\eta}_{p}) = k(\tau; \boldsymbol{\eta}_{p}) = \sum_{i=1}^{Q} \alpha_i \exp \left[ -2 \pi^{2} \tau^2 \sigma_{i}^2 \right] \cos(2 \pi \tau \mu_i),
\label{eq:SM-Kernel}
\end{equation}
where $\boldsymbol{\eta}_{p} =  [\alpha_1, \alpha_2,\cdots,\alpha_Q, \mu_1,\mu_2,\cdots,\mu_Q, \sigma_{1}^2, \sigma_{2}^2,\cdots, \sigma_{Q}^2]^T$ denotes the hyper-parameters of the SM kernel to be optimized and $\tau \triangleq x - x'$ owing to the stationary assumption. For accurate approximation, however, we need to choose a large $Q$, which potentially leads to an over-parameterized model with many redundant localized Gaussian components. Besides, optimizing the frequency and variance parameters is numerically difficult as a non-convex problem,  and often incurs bad local minimal. 

To remedy the aforementioned numerical issue, in \cite{yin2020linear}, it was proposed to fix the frequency and variance parameters, $\mu_1,\mu_2,\cdots,\mu_Q, \sigma_{1}^2, \sigma_{2}^2,\cdots, \sigma_{Q}^2$, in the original SM kernel to some known grids and focus solely on the weight parameters, $\alpha_1, \alpha_2,\cdots,\alpha_Q$. The resulting kernel is called \emph{grid spectral mixture} (GridSM) kernel. By fixing the frequency and variance parameters, the above GridSM kernel can be regarded as a linear multiple kernel with $Q$ basis subkernels, $k_{i}(\tau) \triangleq \exp \left[ -2 \pi^{2} \tau^2 \sigma_{i}^2 \right] \cos(2 \pi \tau \mu_i), i=1,2,\cdots,Q$. In \cite{yin2020linear}, it was shown that for sufficiently small variance, each subkernel matrix has a low-rank smaller than $N/2$, namely half of the data size. Therefore, it falls under the formulation in (\ref{eq:evidence-for-gp}). The corresponding weight parameters of such over-parameterized kernel can be obtained effectively via optimizing the evidence function in (\ref{eq:evidence-for-gp}), and the solution turns out to be sparse, as it is demonstrated in Section~\ref{sec:inference}.

\subsection{Sparsity-Aware Modeling for Tensor Decompositions} 
\label{subsec:SA-tensor}
In the previous subsections, we elucidate the sparsity-aware modeling for the two recent supervised data analysis tools, namely the DNNs and GPs. The underlying idea of employing an over-parameterized model and embedding sparsity via an appropriate prior has inspired  recent sparsity-aware modeling for the unsupervised learning tools in the context of tensor decomposition, see e.g., \cite{cheng2022towards, cheng2020learning, zhou2019bayesian, cheng2016probabilistic, zhao2015bayesian, zhang2018variational}. 

For a pedagogical purpose, we first introduce the basics of tensors and tensor \emph{canonical polyadic decomposition} (CPD), the most fundamental tensor decomposition model in \emph{unsupervised learning}. 

\subsubsection{Tensors and CPD} Tensors are regarded as \emph{multi-dimensional generalization of matrices}, thus providing a natural representation for any multi-dimensional dataset. Specifically, a $P$-dimensional ($P$-D) dataset can be represented by a $P$-D tensor $\bc D \in \mathbb R^{J_1 \times J_2 \times \cdots J_P}$ \cite{sidiropoulos2017tensor}. Given a tensor-represented dataset $\bc D$, the unsupervised learning considered in this article aims to identify the underlying source signals that generate the observed data. In different fields, this task gets different names, such as  ``clustering'' in social network analysis \cite{Papa13}, ``blind source separation'' in  electroencephalogram (EEG) and functional magnetic resonance imaging (fMRI)  data analysis \cite{becker2015brain, chatzichristos2019blind}, and ``blind signal estimation'' in radar/sonar signal processing  \cite{TensorRadar}. In these applications, tensor CPD  has been proven to be a powerful  tool with good interpretability. 

Formally, the tensor CPD is defined as follows:
\begin{tcolorbox}
\textit{Definition of Tensor CPD}  \cite{sidiropoulos2017tensor}: Given a $P$-D tensor  $\bc D \in \mathbb R^{J_1 \times J_2 \times \cdots J_P}$, CPD seeks to find the vectors  $\{ \boldsymbol a^{(1)}_{r}, \boldsymbol a^{(2)}_{r}, \cdots, \boldsymbol a^{(P)}_{r}\}_{r=1}^R$ such that 
\begin{align}
\bc D & =  \sum_{r=1}^R  \underbrace{\boldsymbol a^{(1)}_{r}  \circ \boldsymbol a^{(2)}_{r} \circ \cdots \circ   \boldsymbol a^{(P)}_{r}}_{\text{rank-1 tensor}}, \nonumber \\
& \triangleq \llbracket  \boldsymbol A^{(1)},  \boldsymbol A^{(2)},\cdots, \boldsymbol A^{(P)}  \rrbracket,
\label{cpd_def}
\end{align} 
where $\circ$ denotes vector outer product; $\boldsymbol A^{(p)} \triangleq [\boldsymbol a^{(p)}_{1}, \boldsymbol a^{(p)}_{2}, \cdots, \boldsymbol a^{(p)}_{R}] \in \mathbb R^{J_p \times R},\forall p$, is called the factor matrix. The minimal number $R$ that yields the above expression is termed as the tensor rank.
\end{tcolorbox}

From this definition, it is readily seen that the tensor CPD is a multi-dimensional generalization of a matrix decomposition in terms of  rank-1 representation.  Particularly, when $P = 2$, Eq. \eqref{cpd_def} reduces to decomposing a matrix $\boldsymbol D \in \mathbb R^{J_1 \times J_2}$ into the summation of $R$ rank-1 matrices, i.e., $\boldsymbol D =   \sum_{r=1}^R  \boldsymbol a^{(1)}_{r} \circ  \boldsymbol a^{(2)}_{r} $. 
By defining the term $\boldsymbol a^{(1)}_{r}  \circ \boldsymbol a^{(2)}_{r} \circ \cdots \circ   \boldsymbol a^{(P)}_{r}$ as a $P$-D rank-1 tensor, CPD essentially seeks for $R$ rank-1 tensors/components from the observed dataset, each corresponding to  one specific underlying source signal. Thus, the tensor rank $R$ has a clear physical meaning, namely it corresponds to the number of the underlying source signals. Different from the matrix decomposition, where the rank-1 components are in general not unique, CPD for a $P$-D tensor ($P > 2$) gives unique rank-1 components under mild conditions  \cite{sidiropoulos2017tensor}. The uniqueness endows superior interpretability of the CPD model used in various unsupervised data analysis tasks. 

\subsubsection{Low-Rank CPD and Sparsity-Aware Modeling} In real-world data analysis,  the number of underlying source signals is usually small. For instance,  in brain-source imaging \cite{becker2015brain, chatzichristos2019blind}, both the EEG and fMRI data analysis outcomes have shown that only a small fraction of source signals contribute to the observed brain activities. This suggests that the assumed  CPD model should have a small tensor rank $R$ to avoid data overfitting.
\begin{figure}[!t]
\centering
\includegraphics[width= 6 in]{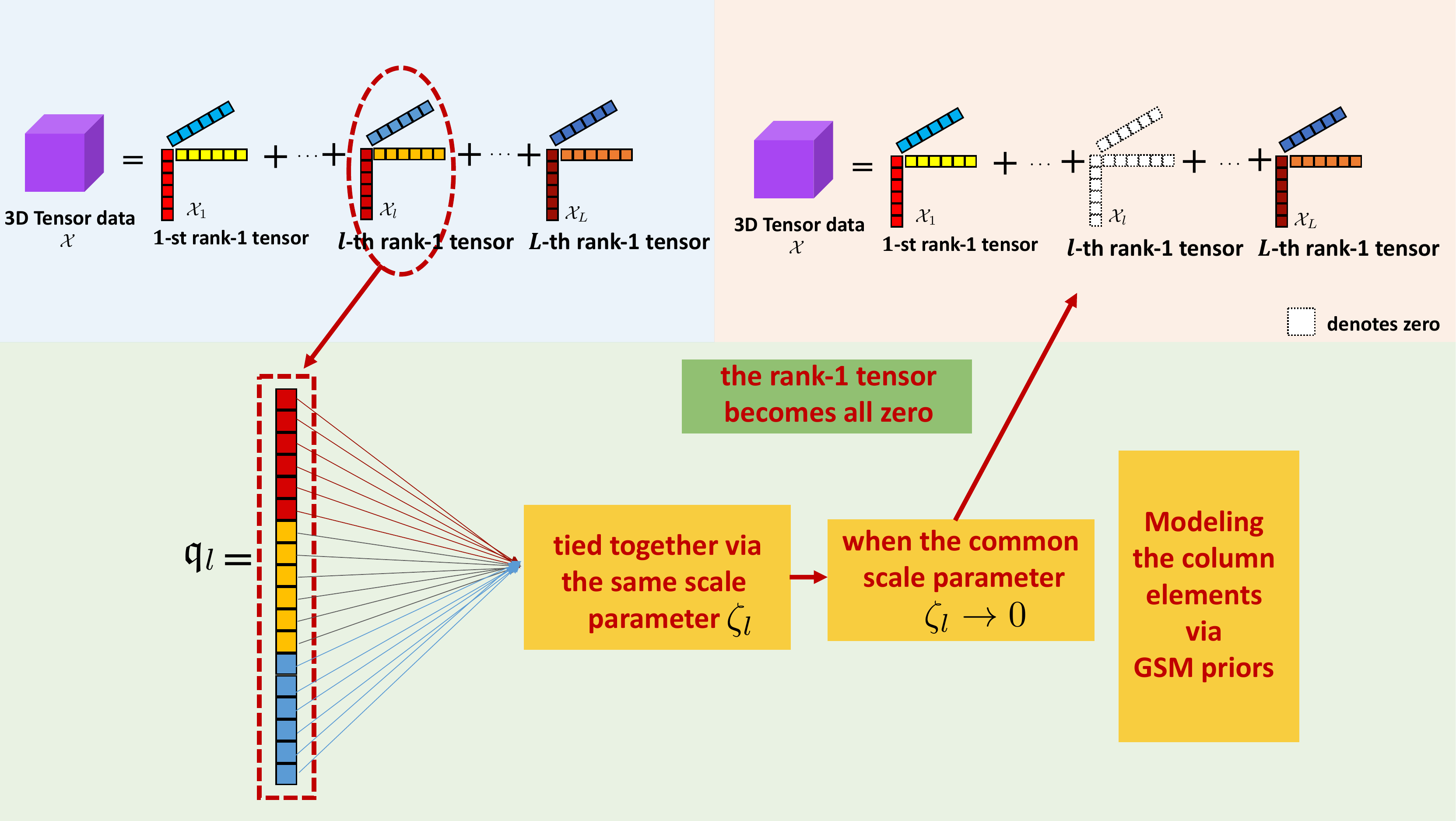}
\caption{Illustration of sparsity-aware modeling for rank-1 tensors using GSM priors.}
\label{fig11}
\end{figure}

In the sequel, we show how the \emph{low-rankness} is embedded into the CPD model through practicing the ideas reported in the previous two sections. First, we employ an over-parameterized model for CPD by assuming an upper-bound value $L$ of tensor rank $R$, i.e., $L \gg R$. The low-rankness implies that $L-R$ rank-1 tensors should be zero, each specified by vectors $\{\bm a_l^{(p)}\}_{p=1}^P, \forall l$. In other words, let vector $\mathfrak q_l \triangleq [\boldsymbol a_{l}^{(1)}; \boldsymbol a_{l}^{(2)}; \cdots; \boldsymbol a_{l}^{(P)} ] \in \mathbb R^{\sum_{p=1}^P J_p}, \forall l$. The low-rankness indicates that a number of vectors in the set $\{\mathfrak q_l  \}_{l=1}^L$ are zero vectors. To model such sparsity, we adopt the following multivariate extension of GSM prior introduced in Section \ref{subsec:SAL-bayes-parametric}, that is:
\begin{align}
p(\mathfrak q_l) & = \int \prod_{i=1}^{\sum_{p=1}^P J_p} \mathcal N([\mathfrak q_l]_i ; 0, \zeta_l) p(\zeta_l; \boldsymbol{\eta}_{p}) d \zeta_l, \nonumber \\
& = \int  \mathcal N( \mathfrak q_l ; 0, \zeta_l \boldsymbol I) p(\zeta_l; \boldsymbol{\eta}_{p}) d \zeta_l, \nonumber \\
&= \int \prod_{p=1}^P \mathcal N( \boldsymbol a_{l}^{(p)} ; 0, \zeta_l \boldsymbol I ) p(\zeta_l; \boldsymbol{\eta}_{p}) d \zeta_l,
\label{gsm_prior_cpd}
\end{align}
where $[\mathfrak q_l]_i$ denotes the $i$-th element of vector $\mathfrak q_l$. Since the elements in $\mathfrak q_l$ are assumed to be statistically independent, then according to the definition of a multivariate Gaussian distribution, we have the second and third lines of \eqref{gsm_prior_cpd} showing the equivalent prior modeling on the concatenated vector  $\mathfrak q_l $ and the associated set of vectors $\{\bm a_l^{(p)}\}_{p=1}^P$,  respectively. The mixing distribution $p(\zeta_l; \boldsymbol{\eta}_{p})$ can be any one listed  in Table \ref{tab2}. Note that in \eqref{gsm_prior_cpd}, the elements in vector $\mathfrak q_l$ are tied together via a common hyper-parameter $\zeta_l$. Once the learning phase is over, if $\zeta_l$ approaches zero, the elements in  $\mathfrak q_l$ will shrink to zero simultaneously, thus nulling a rank-1 tensor, as illustrated in Fig. \ref{fig11}.  Since the prior distribution given in \eqref{gsm_prior_cpd} favors zero-valued rank-1 tensors, it promotes the low-rankness of the CPD model. 

{\it Remark 2:} If the factor matrices are further constrained to be non-negative for enhanced interpretability in certain applications, simple modification, that is, multiplying a unit-step function  $\text{U}(\boldsymbol a_{l}^{(p)} \geq 0)$ (which returns one when $\boldsymbol a_{l}^{(p)} \geq 0$ or zero otherwise) to the prior derived in \eqref{gsm_prior_cpd}, can be made to embed both the non-negativeness and the low-rankness, see in-depth discussions in \cite{cheng2020learning}.

\subsubsection{Extensions to Other Tensor Decomposition Models} Similar ideas have been applied to other tensor decomposition models including Tucker decomposition (TuckerD)  \cite{zhao2015bayesiantucker} and tensor train decomposition (TTD)  \cite{hawkins2020compact, clssp21, xu2020learning}. In these works, one first assumes an over-parametrized model by setting the model configuration parameters (e.g., multi-linear ranks in TuckerD and TT ranks in TTD) to be large numbers, and then impose GSM prior on the associated model parameters to control the model complexity, see detailed discussions  in \cite{zhao2015bayesiantucker, hawkins2020compact, clssp21, xu2020learning}.

{\it Remark 3:} Some further suggestions are given on choosing appropriate tensor decomposition models for different data analysis tasks, see e.g., \cite{proceeding_tensor}. If the interpretability is crucial, one might try CPD (and its structured variants) first, due to its appealing uniqueness property. On the other hand, if the task is related to subspace learning, Tucker decomposition should be considered since its model parameters can be interpreted as the basis functions and the associated coefficients. For missing data imputation, TTD is a good choice as it disentangles different tensor dimensions. More concrete examples can be found in the recent overview paper \cite{proceeding_tensor}. 

%

\section{The Art of Inference: Evidence Maximization and Variational Approximation} 
\label{sec:inference}

Having introduced sparsity-promoting priors, we are now at the stage of deriving the associated Bayesian inference algorithms that aim to learn both the posterior distributions of the unknown parameters/functions and the optimal configurations of the model hyper-parameters. In Section~\ref{subsec:evidence-max-framework}, we will first show that the inference algorithms developed for our considered data analysis tools can be unified into a common evidence maximization framework. Then, for each data analysis tool, we will further show how to leverage recent advances in \emph{variational approximation} and \emph{non-convex optimization} to deal with specific problem structures for enhanced learning performance. Concretely, we introduce inference algorithms for GP in Section~\ref{subsec:GP-inference}, for tensor decompositions in Section~\ref{subsec:infer_tensor}, and for Bayesian deep neural networks in Section~\ref{subsec:BNN-inference}. 

\subsection{Evidence Maximization Framework}
\label{subsec:evidence-max-framework}
Given a data analysis task and having selected the learning model, $\mathcal M$, that is the associated likelihood function $p_{\mathcal M}(\mathcal D | \boldsymbol \theta)$ and a sparsity-promoting prior $p_{\mathcal M}(\boldsymbol \theta; \boldsymbol{\eta}_{p} )$, the goal of Bayesian SAL is to infer the posterior distribution $p_{\mathcal M}( \boldsymbol \theta |  \mathcal D; \boldsymbol \eta)$, using the Bayes' theorem given in \eqref{eq1}, and to compute the model hyper-parameters $\boldsymbol \eta$ by maximizing the evidence $p_{\mathcal M}( \mathcal D; \boldsymbol \eta)$.

We differentiate the following two cases of the evidence function. First, if the evidence $p_{\mathcal M}( \mathcal D; \boldsymbol \eta)$ defined in \eqref{evidence_def} can be derived analytically, such as \eqref{evidence_lr} in the Bayesian linear regression example, the model hyper-parameters $\boldsymbol \eta$ can be learnt via solving the evidence maximization problem, for which advanced non-convex optimization tools, e.g., \cite{kingma2014adam, razaviyayn2013unified, hong2016convergence, sun2016majorization, chang2020distributed}, can be utilized to find high-quality solutions. In this case, since the prior, likelihood and evidence all have analytical expressions, applying the Bayes' theorem \eqref{eq1} yields a closed-form posterior distribution of the unknown parameters.

Unfortunately in most cases, the multiple integration required in computing the evidence \eqref{evidence_def} turns out to be analytically intractable. Inspired by the ideas of the \emph{Minorize-Maximization} (also called Majorization-minimization (MM)) optimization framework \cite{sun2016majorization}, we can seek for a tractable lower bound (or a valid surrogate function in general) that minorizes the evidence function, and maximize the lower bound iteratively until convergence. It has been shown, see e.g., \cite{parisi1988statistical}, \cite{theodoridis2020machine, Bishop2006}, that such an optimization process can push the evidence function to a stationary point.  More concretely, the logarithm of the evidence function is lower bounded as follows:
\begin{align}
\log p_{\mathcal M}( \mathcal D; \boldsymbol \eta) \geq \mathcal{L} (q(\boldsymbol \theta); \boldsymbol \eta ),
\label{ELBO-inequality}
\end{align}
where the lower bound 
\begin{align}
\mathcal{L} (q(\boldsymbol \theta); \boldsymbol \eta ) \triangleq \int q(\boldsymbol \theta) \log \frac{p(\mathcal D, 
\boldsymbol \theta; \boldsymbol \eta)}{q(\boldsymbol \theta)} d\boldsymbol \theta,
\label{ELBO}
\end{align}
is called \textit{evidence lower bound} (ELBO), and $q(\boldsymbol \theta)$ is known as the variational distribution. The tightness of the ELBO is determined by the closeness between the variational distribution $q(\boldsymbol \theta)$ and the posterior  $p_{\mathcal M}( \boldsymbol \theta |  \mathcal D; \boldsymbol \eta)$, measured by  the Kullback-Leibler (KL) divergence, $ \text{KL} \left(q(\boldsymbol \theta) || p_{\mathcal M}( \boldsymbol \theta |  \mathcal D; \boldsymbol \eta) \right)$. In other words, the ELBO becomes tight, i.e., the lower bound becomes equal to the evidence when $ \text{KL} \left(q(\boldsymbol \theta) || p_{\mathcal M}( \boldsymbol \theta |  \mathcal D; \boldsymbol \eta) \right) = 0$, which holds true if and only if $q(\boldsymbol \theta) = p_{\mathcal M}( \boldsymbol \theta |  \mathcal D; \boldsymbol \eta)$. This is easy to see if we expand (\ref{ELBO}) and reformulate it as
\begin{equation}
\log p_{\cal M}({\cal D};\bm{\eta})={\cal L}(q(\boldsymbol{\theta});\bm{\eta}) + \text{KL}( q(\boldsymbol{\theta}) ||p_{\mathcal{M}}(\bm{\theta} | \mathcal{D}; \bm{\eta})).
\end{equation}
Since the KL divergence is nonnegative, the equality in (\ref{ELBO-inequality}) holds if and only if it is equal to zero. 

Since the ELBO in \eqref{ELBO} involves two arguments, namely, $q(\boldsymbol \theta)$ and $\boldsymbol \eta$, solving the maximization problem
\begin{align}
\max_{q(\boldsymbol \theta), \boldsymbol \eta}  \mathcal{L} (q(\boldsymbol \theta); \boldsymbol \eta ),
\label{evidence_max-ELBO}
\end{align}
can provide both an estimate of the model hyper-parameters and the posterior distributions. These two terms can be optimized in an alternating fashion. Different strategies for optimizing $q(\boldsymbol \theta)$ and $\boldsymbol \eta$ result in different inference algorithms. For example, the variational distribution $q(\boldsymbol \theta)$ can be optimized either via functional optimization \cite{zhang2018advances}, or via Monte Carlo method \cite{zhang2019cyclical}, while the hyper-parameters $\boldsymbol \eta$ can be optimized via various non-convex optimization methods  \cite{kingma2014adam, bottou2010large, razaviyayn2013unified, hong2016convergence, lan2020first}. 

In the following subsections, we will introduce some inference algorithms designed specifically for the three popular data analysis tools introduced in Section~\ref{sec:prior-with-recent-tools} that have been equipped with certain sparsity-promoting priors. 

\subsection{Inference Algorithms for GP Regression}
\label{subsec:GP-inference}
Let us start with the GP model for regression, because in this case the evidence function $p_{\mathcal{M}}(\mathcal{D}; \boldsymbol{\eta})$ can be derived analytically owing to the Gaussian prior and likelihood assumed throughout the modeling process. In this subsection, we introduce an effective inference algorithm for GP regression based on the linear multiple kernel in (\ref{eq:GP-LMK}). In Section~\ref{sec:prior-with-recent-tools}, we have already derived  the logarithm of the evidence function in analytical form, as shown in (\ref{eq:evidence-for-gp}). Therefore, we can optimize it directly to obtain an estimate of the model hyper-parameters $\boldsymbol{\eta}$. 

Traditionally, one could estimate the weights of the subkernels, $\alpha_{i}, i=1,2,\cdots,Q$, as well as the precision parameter, $\beta$, using an iterative algorithm similar to the one derived in \cite{Tipping03fastmarginal}. In particular, one sequentially solves for $\alpha_i$, $i=1,2,\cdots,Q$, from the equation $\gamma(\alpha_i) = 0$ derived in (\ref{eq:GP-RVM-gamma}) by fixing the rest of the weights to their latest estimate and then check its relevance with the data in each iteration. This iterative method works quite well for various different datasets. In the sequel, however, we introduce a potentially more effective numerical method in terms of the sensitivity to an initial guess and the data fitting performance than the original one \cite{wipf2004sparse}. 

Next, we take the GridSM kernel in (\ref{eq:SM-Kernel}) as an example of the linear multiple kernel, and rewrite the  evidence maximization problem as
\begin{align}
\boldsymbol{\eta}^{*} = \arg \min_{\boldsymbol{\eta}}  \,  l(\boldsymbol{\eta}) \triangleq  g(\boldsymbol{\eta}) - h(\boldsymbol{\eta}), 
\label{eq:MLnew}
\end{align}
where $\boldsymbol{\eta}=[\boldsymbol{\alpha}^T; \beta]^T$ with $\boldsymbol{\alpha} \geq \boldsymbol{0}$ and $\beta > 0$, $g(\boldsymbol{\eta}) \triangleq \boldsymbol{y}^T \boldsymbol{C}^{-1}(\boldsymbol{\eta}) \boldsymbol{y}$, and $h(\boldsymbol{\eta}) \triangleq -\log \det (\boldsymbol{C}(\boldsymbol{\eta}))$. Let us introduce a short notation $\boldsymbol{C}(\boldsymbol{\eta}) \triangleq \sum_{i=1}^{Q} \alpha_{i} \boldsymbol{K}_{i} +  \beta^{-1} \boldsymbol{I}$, where $\boldsymbol{K}_{i}$ represents the $i$-th sub-kernel matrix evaluated with the training inputs. It can be shown that $g(\boldsymbol{\eta})$ and $h(\boldsymbol{\eta}): \Theta \to \mathbb{R}$ are both convex and differentiable functions with $\Theta$ being a convex set. Therefore, the cost function in (\ref{eq:MLnew}) is a difference of two convex functions with respect to $\boldsymbol{\eta}$, and the optimization problem belongs to the well known \emph{difference-of-convex} program (DCP) \cite{BV04}. Instead of adopting the classic iterative procedure proposed for the RVM \cite{Tipping03fastmarginal}, we take advantages of the DCP optimization structure. Such a favorable structure may help the speed-up of the convergence process, and avoiding to be trapped in a bad local minimum of the optimization problem, and, thus, to further improve the level of sparsity \cite{yin2020linear}. 

\subsubsection{Sequential Majorization-Minimization (MM) Algorithm}
The main idea is to solve $\min_{\boldsymbol{\eta} \in \Theta}\, l(\boldsymbol{\eta})$ with $\Theta \subseteq \mathbb{R}^{Q+1}$ through an iterative scheme, where in each iteration a so-called majorization function $\bar{l}(\boldsymbol{\eta},\boldsymbol{\eta}^{k})$ of $l(\boldsymbol{\eta})$ at $\boldsymbol{\eta}^{k} \in \Theta$ is minimized, i.e.,
\begin{equation}
\boldsymbol{\eta}^{k+1} = \arg \min_{\boldsymbol{\eta} \in \Theta} \bar{l}(\boldsymbol{\eta},\boldsymbol{\eta}^{k}),
\label{eq:MM}
\end{equation}
where the majorization function $\bar{l}(\cdot, \cdot): \Theta \times \Theta \to \mathbb{R}$ satisfies $\bar{l}(\boldsymbol{\eta},\boldsymbol{\eta}) = l(\boldsymbol{\eta})$ for $\boldsymbol{\eta} \in \Theta$ and $l(\boldsymbol{\eta}) \le \bar{l}(\boldsymbol{\eta},\boldsymbol{\eta}')$ for $\boldsymbol{\eta}, \boldsymbol{\eta}' \in \Theta$. We adopt the so-called linear majorization. Concretely, we make the convex function $h(\boldsymbol{\eta})$ affine by performing the first-order Taylor expansion and obtain:
\begin{equation}
\bar{l}(\boldsymbol{\eta},\boldsymbol{\eta}^{k}) \triangleq g(\boldsymbol{\eta}) - h(\boldsymbol{\eta}^{k}) - \nabla_{\boldsymbol{\eta}}^{T} h(\boldsymbol{\eta}^{k}) (\boldsymbol{\eta} - \boldsymbol{\eta}^{k}).
\label{eq:linearMM}
\end{equation}
In this way, minimizing the cost function in (\ref{eq:MM}) becomes a convex optimization problem in each iteration. By fulfilling the regularity conditions, the MM method is guaranteed to converge to a stationary point \cite{sun2016majorization}. Next, we show how (\ref{eq:MM}) with the linear majorization in (\ref{eq:linearMM}) can be solved. 
%

Since $g(\boldsymbol{\eta})$ is a matrix fractional function, in each iteration \eqref{eq:MM} actually solves a convex matrix fractional minimization problem \cite{BV04}, which is equivalent to a semi-definite programming (SDP) problem via the Schur complement.
%
%
This problem can be further cast into a second-order cone program (SOCP) problem and can efficiently and reliably be solved using the off-the-shelf convex solvers, e.g., MOSEK \cite{yin2020linear}. 

Although the previous MM algorithm can often lead to rather good solution, it cannot ensure local minimal in all cases. Occasionally, we found that it provides less satisfactory results, and they can be significantly improved by using a novel non-linearly constrained \emph{alternating direction method of multipliers} (ADMM) algorithm as proposed in \cite{yin2020linear}. In general, the ADMM algorithm takes the form of a decomposition coordination procedure, where the original large problem is decomposed into a number of small local subproblems that can be solved in a coordinated way \cite{Boyd11}.

For our problem, the idea is to reformulate the original problem by introducing an $N \times N$ matrix $\boldsymbol{S}$ and solve instead 
\begin{align}
 \arg & \min_{\boldsymbol{S}, \boldsymbol{\alpha}} \boldsymbol{y}^{T} \boldsymbol{S} \boldsymbol{y} - \log \det(\boldsymbol{S}), \nonumber \\
&\textrm{s.t.} \,\, \boldsymbol{S} \left( \sum_{i}^{Q} \alpha_{i} \boldsymbol{K}_{i} + \beta^{-1} \boldsymbol{I} \right) = \boldsymbol{I}, \quad \boldsymbol{\alpha} \geq \boldsymbol{0}.
\end{align}
Then, an augmented Lagrangian function can be formulated and solved by the ADMM algorithm through iteratively updating the auxiliary matrix variable $\boldsymbol{S}$, the kernel hyper-parameters, $\boldsymbol{\alpha}$, and some associated dual variables. By introducing an auxiliary matrix variable $\boldsymbol{S}$, all ADMM subproblems become convex; in particular, the weight parameters, $\boldsymbol{\alpha}$, are derived in closed form. From the experimental evaluation results given in \cite{yin2020linear}, this ADMM algorithm can potentially find a better local minimum with improved prediction accuracy compared with the MM algorithm, however, at the cost of increased computational time in practice.

In addition to the above mentioned MM- and ADMM algorithms, one could also resort to some other advanced optimization algorithms to solve the problem; for instance, the successive convex approximation (SCA) algorithms reviewed in \cite{sun2016majorization} are of great potential.

\textcolor{black}{The computational complexity for one iteration of the MM algorithm is $\mathcal{O}(n^2 \cdot \max(n, \sum_{i=1}^{Q} r_i))$, where $r_i$ stands for the rank of $\boldsymbol{K}_i$. The MM algorithm benefits from the low-rank property of the GSM subkernels. Let the average rank of the GSM subkernels be, i.e., $\bar{r} = \frac{1}{Q} \sum_{i=1}^{Q} r_i \ll n$; if $Q \bar{r} > n$, then the overall complexity of the MM algorithm scales as $\mathcal{O}(Q \cdot \bar{r} \cdot n^2)$; otherwise, it scales as $\mathcal{O}(n^3)$. Similar conclusions hold for the ADMM algorithm too. These results show that the complexity also relies on the preselected number of subkernels, $Q$, which is often set to a larger value than the one that  it is actually required; however, how to set this parameter adaptively and economically for different datasets remains an open challenge.}

%
%
%

\subsection{Inference Algorithms for Bayesian Tensor Decompositions}
\label{subsec:infer_tensor}
In this subsection, we introduce the inference algorithm design for Bayesian tensor decompositions. Our focus will be on presenting the key ideas for deriving inference algorithms for the Bayesian tensor CPD model \cite{cheng2022towards, cheng2020learning, zhou2019bayesian, cheng2016probabilistic, zhao2015bayesian, zhang2018variational} via the Gaussian likelihood and the GSM prior (introduced in Section~\ref{subsec:SA-tensor}). For other tensor decomposition formats, e.g., the Bayesian tensor TuckerD \cite{zhao2015bayesiantucker} and TTD \cite{xu2020learning}, since they share the same prior design principle as that of CPD, the associated inference algorithm follows a similar rationale.

In the Bayesian tensor CPD,  the goal of inference is to estimate the posterior distributions of factor matrices $\{\boldsymbol A^{(p)} \in \mathbb R^{J_p \times L}\}_{p=1}^P$ from possibly incomplete $P$-D tensor data observations $\bc Y_{\boldsymbol \Omega} \in \mathbb R^{J_1 \times \cdots \times J_P}$, where $\bc Y_{j_1, \cdots, j_P}$ is observed if the $P$-tuple indices $(j_1, \cdots, j_P)$ belongs to the set $\boldsymbol \Omega$. The forward problem is commonly modeled as a Gaussian likelihood:
\begin{align}
p(\bc Y_{\boldsymbol \Omega} | \{\boldsymbol A^{(p)}\}_{p=1}^P; \beta) = \prod_{(j_1,\cdots,j_P) \in \boldsymbol \Omega } \mathcal N( \bc Y_{j_1,\cdots,j_P} ~ ;~  \llbracket  \boldsymbol A^{(1)}, \cdots, \boldsymbol A^{(P)}  \rrbracket_{j_1,\cdots,j_P}, \beta^{-1}),
\label{cpd_likelihood}
\end{align} 
where $\beta$ is the precision (the inverse of variance) of the Gaussian noise. Since it is unknown, a non-informative prior (e.g., Jeffery prior, $p(\beta) \propto 1/\beta$) can be employed. To promote the low-rankness, for the $l$-th columns of all the factor matrices, a GSM sparsity-promoting prior with latent variance variable $\zeta_l$ has been adopted, see detailed discussions in Section \ref{subsec:SA-tensor}. Usually, the hyper-parameters $\boldsymbol \eta$ in the adopted GSM priors are pre-selected to make the prior non-informative, and thus need no further optimization. The unknown parameters $\boldsymbol \theta$ include the factor matrices $\{\boldsymbol A^{(p)}\}_{p=1}^P$, the latent variance variables $\{ \zeta_l\}_{l=1}^L$ of the GSM priors, and the noise precision $ \beta$. Under the evidence maximization framework, the inference problem can be formulated as \eqref{evidence_max-ELBO} with unknown parameters\footnote{The expression of the objective function in \eqref{evidence_max-ELBO} is quite lengthy,  see e.g., \cite{zhao2015bayesian}, and thus is not included here.} 
\begin{align}
& \boldsymbol \theta \triangleq \{\{\boldsymbol A^{(p)}\}_{p=1}^P, \{ \zeta_l\}_{l=1}^L, \beta\},
\end{align}
and the joint pdf 
\begin{align}
& p(\mathcal D, \boldsymbol \theta)  \triangleq p( \bc Y_{\boldsymbol \Omega}, \{\{\boldsymbol A^{(p)}\}_{p=1}^P, \{ \zeta_l\}_{l=1}^L, \beta\} ),
\end{align}
which can be computed by the product of the likelihood and the priors. 

Without imposing any constraint on the pdf $q(\boldsymbol \theta)$, the optimal solution is just the posterior, i.e.,  $q^*(\boldsymbol \theta) = p_{\mathcal M}( \boldsymbol \theta |  \bc Y_{\boldsymbol \Omega})$, whose computation using the Bayes' theorem will, however,  encounter the intractable multiple integration challenge. To get over this difficulty, modern approximate inference techniques propose to solve problem \eqref{evidence_max-ELBO} by further constraining $q(\boldsymbol \theta)$ into a functional family $\mathcal F$, i.e., $q(\boldsymbol \theta) \in  \mathcal F$.
It is hoped that  the family $\mathcal F$ is as flexible as possible to allow accurate posterior estimates and at the same time simple enough to enable tractable optimization algorithm designs.

Among all the functional families, the \textit{mean-field} family is undoubtedly the most favorable one in recent Bayesian tensor research \cite{cheng2022towards, cheng2020learning, zhou2019bayesian, cheng2016probabilistic, zhao2015bayesian, zhang2018variational}. It assumes that the variational pdf $q(\boldsymbol \theta) = \prod_{k=1}^K q(\boldsymbol \theta_k)$, where $\boldsymbol \theta$ is partitioned into mutually disjoint non-empty subsets $\boldsymbol \theta_k$ (i.e., $\cup_{k=1}^K \boldsymbol \theta_k = \boldsymbol \theta$ and $\cap_{k=1}^K \boldsymbol \theta_k = \O $). In the context of the Bayesian tensor CPD, the mean-field assumption states that
\begin{align}
q(\boldsymbol \theta ) = \prod_{p=1}^P q(\boldsymbol A^{(p)}) q(\{ \zeta_l\}_{l=1}^L) q(\beta).
\label{mf_cpd}
\end{align}
The factorized structure in \eqref{mf_cpd} inspires the idea of block minimization in the optimization theory. In particular, for the ELBO maximization problem \eqref{evidence_max-ELBO}, specified after fixing the variational pdfs  $\{q(\boldsymbol \theta_j)\}_{j \neq k}$, the resulting subproblem that optimizes $q(\boldsymbol \theta_k)$ has been shown to have the following optimal solution, see e.g., \cite{theodoridis2020machine}:
\begin{align}
& q^* \left(  \boldsymbol \theta_k\right) = \frac{\exp\left ( \mathbb E_{ \prod_{j \neq k}q \left(   \boldsymbol \theta_j\right) } \left [ \ln  {p\left(  \mathcal D,  \boldsymbol  \theta  \right)} \right]  \right)}{\int \exp\left ( \mathbb E_{ \prod_{j \neq k}q \left(   \boldsymbol \theta_j\right) } \left [ \ln  {p\left( \mathcal D, \boldsymbol  \theta    \right)} \right]  \right)  d\boldsymbol \theta_k },
\label{opt_mf_cpd}
\end{align}
where $\mathbb E_{q(\cdot) } \left[ \cdot \right]$ denotes the expectation with respect to the variational pdf $ q(\cdot)$. The inference framework under the mean-field assumption is termed as \emph{mean-field variational inference} (MF-VI).

Whether the integration in the denominator \eqref{opt_mf_cpd} has a closed-form is determined by the functional forms of the likelihood and the priors. In particular, if they are conjugate pairs within the exponential family of probability distributions, see, discussions in, e.g., \cite{theodoridis2020machine, Bishop2006}, the optimal variational pdf in \eqref{opt_mf_cpd} will accept a closed-form expression. Fortunately, for the Bayesian tensor CPD adopting the Gaussian likelihood and the GSM prior for the columns of the factor matrices, this condition is usually satisfied, which enables the derivation of closed-form updates in recent advances \cite{cheng2022towards, cheng2020learning, zhou2019bayesian, cheng2016probabilistic, zhao2015bayesian, zhang2018variational}.

\emph{Remark 4:}  To facilitate  the algorithm derivation, MF-VI imposes a factorization structure on $q(\boldsymbol \theta)$, which implies the statistical independence of the variables $\boldsymbol \theta_k$ given the observed dataset $\mathcal D$. If this is not the case, the mean-field approximation will lead to mismatch when approaching the ground-truth posteriors. In general, the MF-VI tends to provide posterior approximations that are more compact compared to the true ones, which means that the posterior estimates are usually ``over-confident''\cite{Bishop2006}. To achieve more accurate posterior estimation, there is a research trend to employ more advanced variational approximation techniques than the mean-field approximation. For example, recent tensor-aided Bayesian deep neural network research  \cite{hawkins2020compact} utilizes the kernelized Stein discrepancy to derive the inference algorithm that can approximate the posterior better. The interested reader may refer to \cite{zhang2018advances} for some recent advances in variational approximation methods. 

{\color{black} Some computational and theoretical difficulties that are commonly encountered in Bayesian tensor decompositions are summarized as follows. First, due to the block coordinate descent nature of the MF-VI  \cite{zhang2018advances}, it is crucial to choose informative initial values to avoid poor local minima. On the other hand, the associated computational complexity is cubic with respect to the tensor rank, see, e.g., \cite{cheng2022towards, cheng2020learning, xu2020learning, zhao2015bayesian}, which is high if the initial tensor rank is set to a large value. Finally, it was found that the algorithm performance significantly degrades in some challenging regimes, e.g., low signal-to-noise-ratio (SNR) and/or high rank, see, e.g.,  \cite{cheng2022towards, cheng2020learning, xu2020learning, zhao2015bayesian}. To overcome these difficulties, suggestions on the real algorithm implementations are provided in Section~\ref{subsec:TD-applications}. }

\subsection{Inference Algorithms for Bayesian Deep Neural Networks}
\label{subsec:BNN-inference}
{\color{black}The step of inference (training) for Bayesian deep neural networks follows the same backpropagation-type of philosophy as that of training their deterministic counterparts. There are, however, two notable differences. First, the unknown (synaptic) parameters/weights are now described via parameterized distributions. Thus, the cost function to be optimized has to be expressed in terms of the hyper-parameters that define the respective distributions, instead  of the weights/synapses.   This involves the so-called {\it reparameterization step} and we will describe it in Section \ref{subsec:BNN-inference}. Second, the evidence function to be maximized is not of a tractable form and it has to be approximated by its ELBO (see the definition in \eqref{ELBO}).

In this subsection, we will outline the basic steps that are followed for  
variational inference in the case of a Bayesian deep network that comprises layers with: (a) stochastic LWTA blocks; (b) stochastic synaptic weights of Gaussian form; and (c) a sparsity-inducing mechanism imposed over the network synapses that is driven via an IBP prior. We have already discussed this type of network in Section IV-A.3; see, also,  Fig.~\ref{fig:wta1}. 
To facilitate understanding, we provide a graphical illustration of the considered stochastic LWTA block in  Fig.~\ref{cnn_sb_lwta}. 

\begin{figure}
			\centering
			\includegraphics[width= 6.5 in]{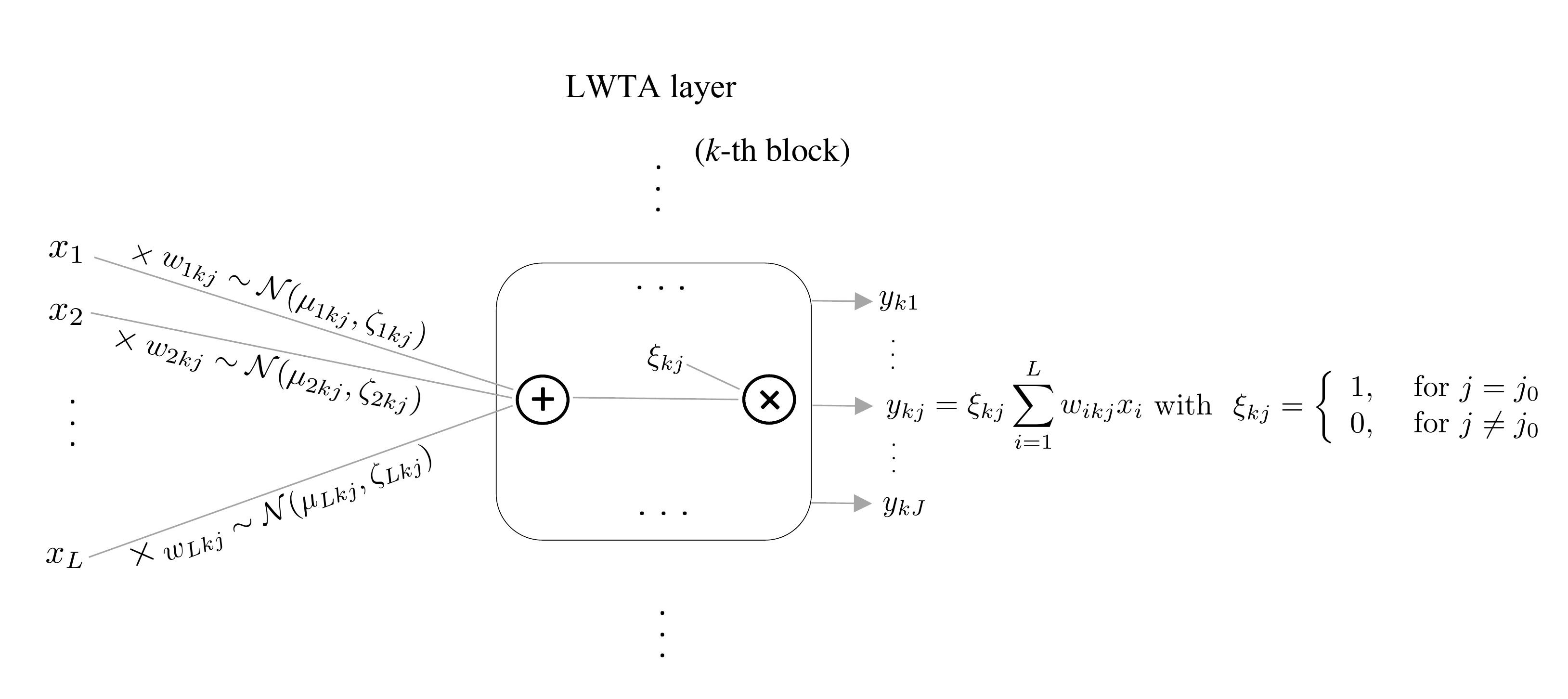}
		\caption{ A zoomed-in graphical illustration of the $k$-th block of a stochastic LWTA layer. Input $\boldsymbol{x} = [x_1, x_2, \ldots, x_{L}]$ is presented to each unit, $j=1,2,\ldots,J$, in the block. Assume that the index of the winner unit is $j=j_0$. Then, the output of the block is a vector with a single non-zero value at index $j_0$. }
	\label{cnn_sb_lwta}
	\end{figure}

Without harming generality, let us focus on a specific layer, say, the  $f+1$ one. In order to slightly unclutter notation, assume that for this layer, the input dimension  $a^f=L$ and the number of nodes (LWTA units) $a^{f+1}=K$.  Thus the corresponding input matrix to the layer becomes $ \boldsymbol{X} \in \mathbb{R}^{N\times L}$ with $N$ samples, each comprising $L$ features. Under the stochastic LWTA-based modeling rationale, nodes (neurons) are replaced by LWTA blocks, each containing a set of  $J$ \emph{competing linear} units. Thus, the layer input is now presented to each different block and each unit therein, via different weights. 
Thus, the weights for this layer are now represented via a three-dimensional matrix $\boldsymbol W \in \mathbb{R}^{L \times K \times J} $ (we again refrain our notation on the dependence on $f$, i.e. the layer index). Recall from Section \ref{subsubsec:link_wise_ibp} that each layer is associated with a latent discrete random vector,
$\boldsymbol \xi_n \in \mathrm{one\_hot}(J)^K$, that encodes the outcome of the local competition among the units in all $K$ LWTA blocks of a network layer, when the $n$-th input sample is presented. 

Furthermore, recall that each link connecting an input dimension, of the $n$-th sample, e.g., $x_{ni}$, to an LWTA block, e.g., the $k$-th one, 
is weighted by a utility binary random variable, $z_{ik}$. This is set equal to one, if the $i$-th dimension of the input is presented to the $k$-{th} LWTA block, otherwise $z_{ik} = 0$. We impose the sparsity-inducing IBP prior over these utility hidden variables. 

We are now ready to write the output of a specific  layer of the considered model, i.e., $\boldsymbol{y}_n \in \mathbb{R}^{K\cdot J}$, as follows:
\begin{align}
y_{nkj} = \xi_{nkj} \sum_{i=1}^L (w_{ikj} z_{ik})   x_{ni} \in \mathbb{R},
\label{eqn:layer_output_ff}
\end{align}
where $\bm{x}_n$ is the $L$-th dimensional input that coincides with the output of  the previous layer.  
The involved random variables, whose {\it posterior} distributions are to be learnt  during training, are: a) the synaptic weights, $w_{ikj},~i=1,2\ldots,L,~k=1,2,\ldots,K,~j=1,2,\ldots,J$, for {\it all layers}, b) the utility variables, $z_{ik}$, for all layers and  c) the indicator vectors, $\bm{\xi}_{nk}$, for the $n$-th sample and the $k$-th LWTA,  for all layers. The functional form of the respective distributions are:

\noindent $\blacksquare$ \underline{Synaptic weights:}
\begin{align}
\mbox{Prior}:  p(w_{ikj})\sim \mathcal{N}(w_{ikj}|0,1),~~\mbox{Posterior}:~q(w_{ikj})\sim \mathcal{N}(w_{ikj}|\mu_{ikj},\zeta_{ikj}), \nonumber
\end{align}
where the mean and variance, $\mu_{ikj}$,  $\zeta_{ikj}$, respectively,  are learnt during training.

\noindent $\blacksquare$ \underline{Utility binary random variables:}
\begin{align}
\mbox{Prior}: \mbox{Bernoulli}(z_{ik}|\pi_{ik}),~~\mbox{Posterior}:~q(z_{ik})=\mbox{Bernoulli}(z_{ik}|\tilde{\pi}_{ik}),
\nonumber
\end{align}
where $\pi_{ik}$ come form the IBP prior (Section \ref{subsec:ibp_prior}) and $\tilde{\pi}_{ik}$ are learnt  during training. The use of the Bernoulli distribution is imposed by the binary nature of the variable.

\noindent $\blacksquare$ \underline{Indicator random vectors, $\bm{\xi}_{nk}$:}
\begin{align}
&\mbox{Prior}:~p(\bm{\xi}_k)=\mbox{Categorical}(\bm{\xi}_k|\frac{1}{J},\ldots,\frac{1}{J}),~\mbox{i.e., ~all linear units equiprobable}. \nonumber \\
&\mbox{Posterior}:~q(\boldsymbol{\xi}_{nk} ) = \mbox{Categorical} \left ( \boldsymbol{\xi}_{nk}\big | P_{nk1},\ldots, P_{nkJ}\right), \nonumber
\end{align} 
where $P_{nkj}$ is defined via the softmax operation, e.g., Eq. (\ref{sergios2}). The Categorical distribution is imposed because only one out of the $J$ elements of $\bm{\xi}_{nk}$ is equal to 1 and the rest are zeros. The $j$-th element becomes 1 with probability $P_{nkj}$. 

Note that besides the previous random variables, which are directly related to the DNN architecture, there is another set of hidden random variables, i.e., the $u_j$'s, which are used for generating the IBP prior. These have also to be considered as part of the palette of the involved random variables. As already said in Section \ref{subsec:ibp_prior}, these follow the Beta distribution, with prior Beta($u_j|\alpha,1$) and posteriors Beta($u_j|a_j,b_j$), where $a_j,~b_j$ are learnt  during training.

To train the proposed model, we resort to the maximization of the ELBO. The trainable model parameters, in our case, are the set of all the weights'   posterior means and variances, i.e., $\mu_{ikj}$ and $\zeta_{ikj}$, the synaptic utility indicator posterior probabilities, $\tilde{\pi}_{ik}$, and the stick-variable posterior parameters $a_j$ and $b_j$, across all network blocks and layers. Let us refer to this set as $\mathbf{\Theta}$.  We shall assume that our task comprises $C$ classes and the softmax nonlinearity is used in the output layer. 

In the following, we denote $\mathcal{D}$ as the input-output training dataset. In addition, let $\mathbf{Z}$ be the set of the synaptic utility indicators across the network layers; $\mathbf{\Xi}$ be the set of winner unit indicators across all blocks of all layers; $\mathbf{W}$ be the set of synapse weights across all layers; and $\mathbf{U}$ be the set of the stick-variables of the sparsity-inducing priors imposed across the network layers. Employing the mean-field approximation on the joint posterior pdf, i.e., factorizing $q(\mathbf{W},\mathbf{Z},\mathbf{\Xi},\mathbf{U})$, it is readily shown, e.g., \cite{panousis2019nonparametric}, that

\begin{equation}
\begin{aligned}
\mbox{ELBO}(\mathbf{\Theta})= -\mathbb{E}_q\Big [ \sum_{n=1}^N\sum_{c=1}^C y_{nc} \ln \tilde{y}_{nc}(\boldsymbol{x}_n; \mathbf{W}, \mathbf{Z}, \mathbf{\Xi},\mathbf{U}) \Big]
&+\underbrace{\mathbb{E}_q\ln \frac{p(\mathbf{Z}|\mathbf{U})}{q(\mathbf{Z})}+
                    \mathbb{E}_{q}\ln \frac{p(\mathbf{U})}{q(\mathbf{U})}}_{ \mbox{regularizing terms}}\\
                    &+\underbrace{\mathbb{E}_{q}\ln\frac{p(\mathbf{\Xi})}{q(\mathbf{\Xi}|\mathbf{Z},\mathbf{W})}+\mathbb{E}_{q}\ln \frac{p(\mathbf{W})}{q(\mathbf{W})}}_{\mbox{regularizing terms}},\label{sergios5}
\end{aligned}
\end{equation}
where $y_{nc}$ are the outputs in the training set and $\tilde{y}_{nc}$, $c=1,2,\ldots,C,$ are the class probability outputs as estimated via the softmax nonlinearities that implement  the output layer. Observe that the first term on the right hand side is the expectation of the 
cross-entropy of the network. The only difference with the deterministic DNNs that use this function to optimize the network is that now the expectation over the posterior is involved. In practice, it turns out that drawing one sample from the involved distributions  suffices to lead to good approximation, provided that this sample is written as a differentiable function of $\mathbf{\Theta}$ and some low-variance random variable, e.g.,  \cite{shayer2018learning, Kingma}. The rest of the  terms in Eq. (\ref{sergios5}) are Kullback-Leibler (KL) divergences that bias the posteriors to be as close as possible to the corresponding priors. In other words, they act as regularizers  that bias the solution towards certain regions in the parameter space as dictated by the adopted priors. For example, in the last term, the posterior of the synaptic weights is biased towards the normal Gaussian and bears close similarities with the  $\ell_2$ regularization, when it is  enforced on the  synaptic weights. 



Drawing samples to approximate the expectations in Eq. (\ref{sergios5}) is closely related to what we called before as reparameterization. Recall 
 that in the framework of the backpropagation algorithm, during the forward pass,  one needs specific values/samples of the involved random parameters in order to compute the outputs, given the input to the network. This is performed via sampling the respective distributions, based on the current estimates of the involved posteriors. However,  in order to be able to optimize with respect to their defining hyper-parameters, the corresponding current estimates should be explicitly considered. Let us take the Gaussian synaptic weights as an example. Let $\mathcal{N}(w_{ikj}|\mu_{ikj},\zeta_{ikj})$ be the current estimate of some weight in some layer. Instead of sampling from this Gaussian, it is easy to see that it is equivalent to obtaining a corresponding sample of the weight, as
\begin{equation}
\tilde{w}_{ikj} = \mu_{ikj} + \zeta_{ikj}^{1/2} \cdot \epsilon, \qquad \epsilon \sim \mathcal{N}(0,1).
\label{70a}
\end{equation}
In this way, every link in the network is determined explicitly by the pair ($\mu_{ijk}, \zeta_{ijk}$), and the backpropagation optimizes with respect to the means and variances. Reparameterization of the rest of the involved random variables follows a similar rationale, yet the involved formulae are slightly more complex; however, they are still given in terms of analytic expressions. For example, for the utility variables, $z_{ik}$,  reparameterization is achieved via the posteriors $\tilde{\pi}_{ik}$ and the so-called Gumbel-Softmax relaxation, e.g., \cite{Maddison}. For the stick breaking variables, reparameterization is achieved via the so-called Kumaraswamy approximation \cite{kumaraswamy1980generalized}. Details and the exact formulea can be found in \cite{panousis2019nonparametric}.

Once samples have been drawn for all the involved variables,   the ELBO in (\ref{sergios5}) is expressed directly in terms of the drawn sample, i.e, $\tilde{\mathbf{W}}, \tilde{\mathbf{Z}},\tilde{\mathbf{\Xi}}, \tilde{\mathbf{U}}$, without any expectation being involved.  The trainable parameters' set, $\mb \Psi$, can be obtained by means of any off-the-shelf gradient-based optimizer, such as the Adam \cite{kingma2014adam}. Note that, by adopting the reported reparameterizations, one  yields low-variance stochastic gradients that ensure convergence. 

Training a fully-Bayesian model slightly increases the required complexity, since more parameters are involved, e.g., instead of a single weight one has to train with respect to the respective mean value and variance, as well as the hidden utility variables. However,  the  training timing remains of the same order as that required by the deterministic versions.

Once training has been completed, during testing, given an input:  a) One can use the mean values of the obtained posterior Gaussians, $\mu_{ijk}$, in place of the synaptic weights, $w_{ijk}$. Sampling from the distribution could also be another possibility. Usually, the mean values are used. b) One can employ a threshold value, e.g.,  $\tau$, and remove all links where the corresponding posterior $\tilde{\pi}$ is below this threshold. Sampling is also another alternative.  c) One samples from the respective categorical distributions to determine which linear unit ``fires" in each block. Selecting the one with the largest probability is another alternative.
\newline
\emph{Remark 5:} At this point, we must stress that the learnt posterior variances over the network weights can be also used for reducing the floating-point bit precision required for storing them; this effectively results in memory footprint reduction. The main rationale behind this process consists of the fact that, the  higher the posterior weight variance,  the more their fluctuation under sampling at inference time. This implies that, eventually, some bits fluctuate too much under sampling, and therefore, storing and retaining their values does not contribute to inference accuracy. Thus, Bayesian methods offer this added luxury, to optimally control the required bit precision for individual nodes. For example, in \cite{panousis2019nonparametric}, it is reported that for the case of LENET-300-100 trained on MNIST, the bit precision can be reduced from 23 bit mantissa to just 2 bits. For more details, see, e.g.,  \cite{panousis2019nonparametric}.\newline
\emph{Remark 6:} In  \cite{panousis2019nonparametric}, it was shown that the resulting architectures are able to yield a significant reduction in their computational footprint, retaining, nevertheless, state-of-the-art performance; positively, the flexibility of the link-wise IBP allowed for a more potent sparsity-activation-aware blend based on {\it stochastic} LWTA, offering significant benefits compared to conventional non-linearities introduced by the activation functions such as sigmoid and ReLU. For example, in the case of CIFAR-10 VGG-like network, it turns out that only around $5\%$ of the original nodes are only retained, without affecting the performance of the network, even though quantized arithmetic was also employed to reduce the required number of bits, as explained in the previous remark, see, e.g.,  \cite{panousis2019nonparametric}.

}

 %


\section{Applications in Signal Processing and Machine Learning} 
\label{sec:application}
In this section, we showcase typical applications of the sparsity-promoting data analysis tools introduced in Section~\ref{sec:prior-with-recent-tools}. More specifically, advanced time series prediction using Gaussian process models is considered in Section~\ref{subsec:GP-application}; adversarial learning using Bayesian deep neural networks is presented in Section~\ref{subsec:BNN-application}; and lastly social group clustering and image completion using unsupervised tensor decompositions are demonstrated in Section~\ref{subsec:TD-applications}.

\subsection{Time Series Prediction via GPs}
\label{subsec:GP-application}
In the following, we present an important signal processing  and machine learning  application, namely the \emph{time series prediction}, using non-parametric GPs. We will focus on the GP regression models with the family of sparse spectrum kernels introduced in Section~\ref{subsec:SA-GP}. To demonstrate the advantages of the sparsity-promoting GP models over other competing counterparts, we selected a number of classic time series datasets such as \textit{$\textrm{CO}_2$, Electricity, Unemployment}\footnote{These datasets are available from the UCL repository} as well as a ``fresh'' real-world  \textit{5G wireless traffic} dataset in our tests. Data descriptions are given in Table~\ref{tab:table-GP-datasets}.
\begin{table}[h]
\begin{center}
   \caption{Descriptions of the selected datasets. The training data, $\mathcal{D}$, is used for optimizing the hyper-parameters of the learning model, while the test data, $\mathcal{D}_{*}$, is used for evaluating the prediction accuracy. The numbers given in the last two columns are the training sample size and test sample size, respectively.}
   \label{tab:table-GP-datasets}
    \begin{tabular}{|c|p{0.6\columnwidth}|c|c|}
    \hline
    Name & Data Description & Training $\mathcal{D}$ & Test $\mathcal{D}_{*}$ \\ \hline
    ECG & Electrocardiography of an ordinary person measured over a period of time & 680 & 20 \\ \hline
    $\textrm{CO}_2$ &  Carbon dioxide concentration observed from 1958 to 2003 & 481 & 20 \\ \hline
    Electricity & Monthly average residential electricity usage in Iowa City from 1971 to 1979 & 86 & 20 \\ \hline
    Employment & Wisconsin employment status observed from January 1961 to October 1975 & 158 & 20 \\ \hline
    Hotel & Monthly hotel occupied rooms collected from 1963 to 1976  & 148 & 20 \\ \hline
    Passenger & Passenger miles flown domestic U.K. form July 1962 to May 1972 & 98 & 20 \\ \hline 
    Clay & Monthly production of clay bricks from January 1956 to August 1995 & 450 & 20 \\ \hline 
    Unemployment & Monthly U.S. female (16-19 years) unemployment figures from 1948 to 1981 & 380 & 20 \\ \hline 
    5G wireless traffic & Downlink data usage in a small cell observed in four weeks of 2021 & 607 & 67 \\ \hline
    \end{tabular}
\end{center}
\end{table}

\subsubsection{Classic Datasets} In the sequel, we compare the performance of the sparsity-promoting GP models using the original SM kernels \cite{Gredilla10, WA13} and the modified GridSM kernel \cite{yin2020linear} with that of a classic deep learning based time series prediction model, namely the long-short-term-memory (LSTM) \cite{Hochreiter97}, as well as a canonical statistical model, namely the autoregressive integrated moving average (ARIMA) model\footnote{\textcolor{black}{ARMA model can be regarded as a special case of a GP model adopting a specific sparse kernel matrix.}} \cite{Brockwell86}, from various different aspects. Furthermore, we compare GP models with recently proposed Transformer-based time series prediction model, called \textit{Informer}, which successfully addressed the computation issues and some inherent limitations of the encoder-decoder architecture in the original Transformer model. For brevity, we name the first sparse spectrum GP model (equivalent to using trigonometric basis functions) proposed in \cite{Gredilla10} as \textit{SSGP}, the GP model using the original SM kernel \cite{WA13} as \textit{SMGP}, and the most recent one with the rectified GridSM kernel \cite{yin2020linear} as \textit{GSMGP}. Their configurations can be found in detail in \cite{yin2020linear}. 

Table~\ref{tab:table-Prediction-MSE} shows the obtained prediction accuracy of the various methods quantified in terms of the prediction \emph{mean-squared-error} (MSE). It is readily observed that the sparsity-promoting GP models, and in particular SMGP and GSMGP, outperform all other competitors by far. The following facts need to be mentioned. For both the classic LSTM and Informer models to achieve good performance, the time series should, in general, be long so that the underlying pattern can be learnt during the training phase. The ARIMA model strongly relies on the optimal configuration of the parameters $(p, d, q)$, and it is incapable for long term prediction. In contrast, the sparsity-promoting GP models can automatically fit the underlying data pattern through solving the hyper-parameters from maximizing the evidence function. We have also shown in the supplement of \cite{yin2020linear} that the GSMGP is also superior to the GP models with elementary kernel (such as the SE kernel, rational quadratic kernel, etc.) as well as a hybrid of those. 
\begin{table*}[t]
	\begin{center}
		\caption{Comparisons of various time series prediction models in terms of the prediction MSE. Herein, we let both the GSMGP and the SMGP employ $Q=500$ Gaussian mixture modes in their kernels, and the SSGP employs the same amount of trigonometric basis functions. The GSMGP samples $Q$ normalized frequency parameters, $\mu_i$, $i=1,2,\cdots,Q$, uniformly from $[0,1/2)$, while set the variance parameter to $\sigma = 0.001$. The LSTM model follows a standard setup with one hidden layer and the dimension of the hidden state is set to 30. The Informer model follows the default setup given in the original paper \cite{Zhou21}. The $\textrm{ARIMA}(p,d,q)$ model is a standard one with $(p=5, d=1, q=2)$.}
		\label{tab:table-Prediction-MSE}
		\begin{tabular}{|c|c|c|c|c|c|c|c|c|c|c|}
			\hline
			Name &GSMGP   & SSGP    & SMGP   & LSTM   & Informer  & ARIMA\\ 
			          & MSE      & MSE      & MSE     & MSE     & MSE         & MSE \\ \hline
			ECG   & \textbf{1.3E-02} & 1.6E-01 & 1.9E-02  & 2.1E-02 & 5.4E-02 & 1.8E-01\\ \hline
			CO2  & 1.5E+00  & 2.0E+02 & \textbf{1.1E+00} & 2.1E+00 & 8.4E+01 & 4.9E+00\\ \hline
			Electricity & \textbf{4.7E+03} & 8.2E+03 & 7.5E+03 & \textbf{4.7E+03} & 8.3E+03 & 1.2E+04\\ \hline
			Employment & 1.1E+02 & 7.7E+01 & \textbf{0.7E+02} & 4.3E+02 & 2.0E+03 & 3.9E+02\\ \hline
			Hotel & \textbf{8.9E+02} & 1.9E+04 & 2.8E+03 & 7.8E+03  & 2.3E+04 & 1.7E+04\\ \hline
			Passenger & 1.9E+02 & 6.9E+02 & 1.6E+02 & 1.6E+02 & \textbf{1.2E+02} & 4.5E+03\\ \hline
			Clay  & 1.9E+02 & 5.3E+02 & 3.3E+02 & 2.7E+02 & \textbf{1.4E+02} & 3.3E+02 \\ \hline
			Unemployment & 3.6E+03 & 2.1E+04 & 1.4E+04  & \textbf{3.5E+03}  & 3.8E+03 & 1.5E+04\\ \hline
		\end{tabular}
	\end{center}
\end{table*}

Besides the improved prediction performance, the training time of the GSMGP outperforms the original SMGP by far.\footnote{Here, training time refers to the computational time required for training the learning models introduced in Section~\ref{sec:inference}.} For the selected datasets, SMGP requires training time in the magnitude of $10^3$ seconds, while the GSMGP only requires $10^2$ seconds. By reducing the number of Gaussian modes, $Q$, the SMGP is able to reduce its training time albeit at the cost of sacrificing the fitting performance. On the other hand, the GSMGP improves its training time by fixing the frequency and variance parameters of the original SM kernel to known grids, so that the evidence maximization task enjoys the favorable difference-of-convex structure that can be efficiently handled by the MM algorithm introduced in Section~\ref{subsec:GP-inference}. In addition to the reduced training time, the overall optimization performance (including the convergence speed, chance of being trapped in a bad local minimum, etc.) and the sparsity level of the solution have been significantly improved. Detailed comparisons and pictorial illustrations can be found in \cite{yin2020linear}. In comparison, the LSTM and ARIMA models require the least training time in the magnitude of $10^1$ seconds on average. However, due to the huge architecture adopted in the Informer model, the computational time is in the magnitude of $10^2$ seconds on average. \textcolor{black}{As it is readily observed from the results, the sparsity-promoting property helps reducing the computational time significantly; more importantly, the sparse solution identifies the most effective frequency components of the data and, thus, leads to good model interpretability.}

\subsubsection{Real 5G Dataset} Next, we focus on another favorable advantage of the GP models over their deep learning counterparts, namely the natural uncertainty region of a point prediction. We specifically select the real 5G wireless traffic dataset for visualization purposes due to the high demand of such a wireless application on a reasonable prediction uncertainty \cite{Xu19, Xu20}. The dataset was collected in a small cell of a southern city in China, and it contains the downlink data volume consumed by the mobile users located in the cell within each hour during a period of four weeks. Accurate prediction of the future downlink data consumption is vital to the operators for tuning the transmit power of the base station and switching on/off it automatically. In this application, uncertainty information is even more crucial because wrongly reducing the transmit power may largely influence the mobile users' surfing experience. 

\begin{figure*}[t]
\centering
\includegraphics[width=.48\textwidth]{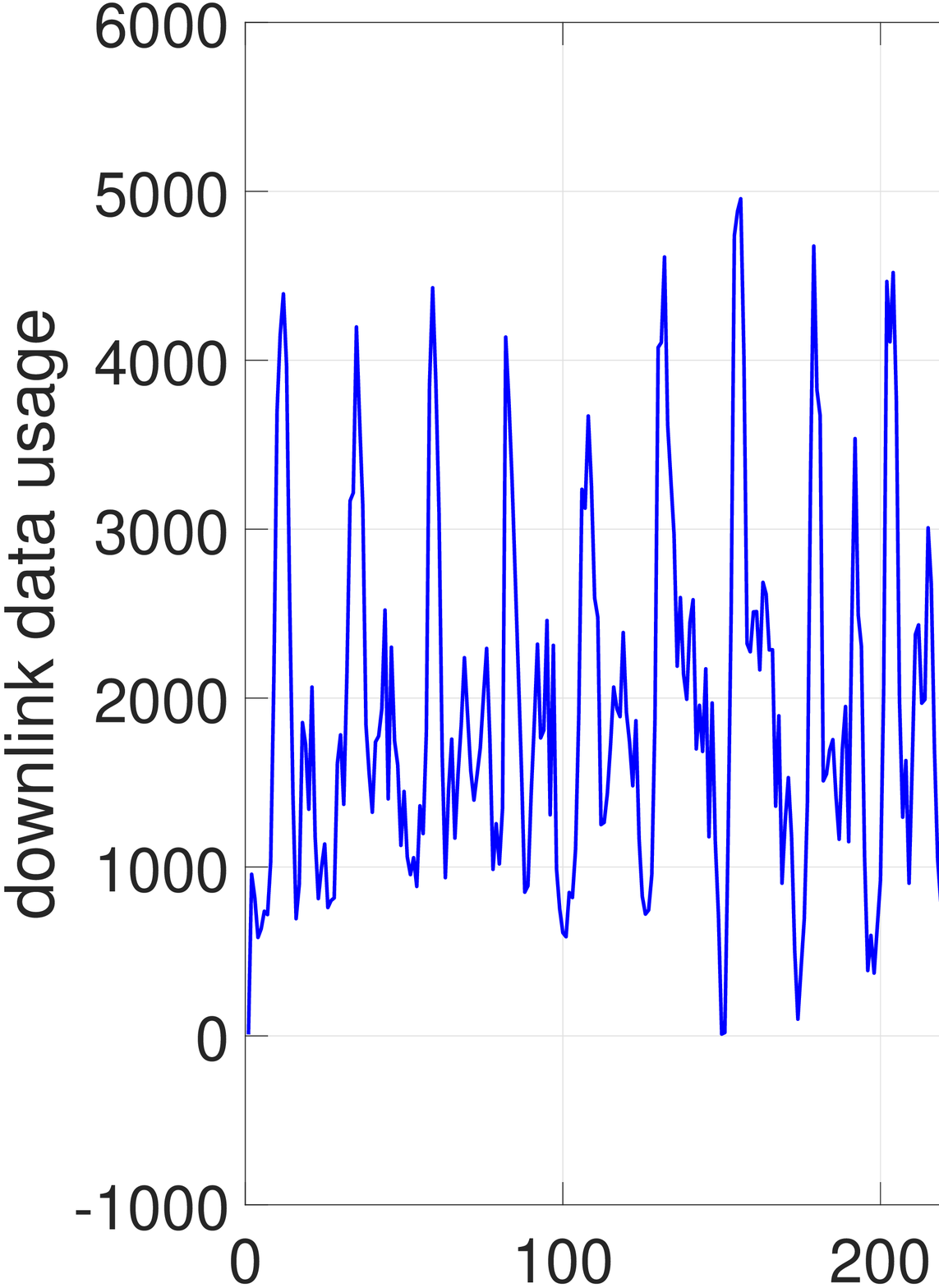}\quad 
\includegraphics[width=.48\textwidth]{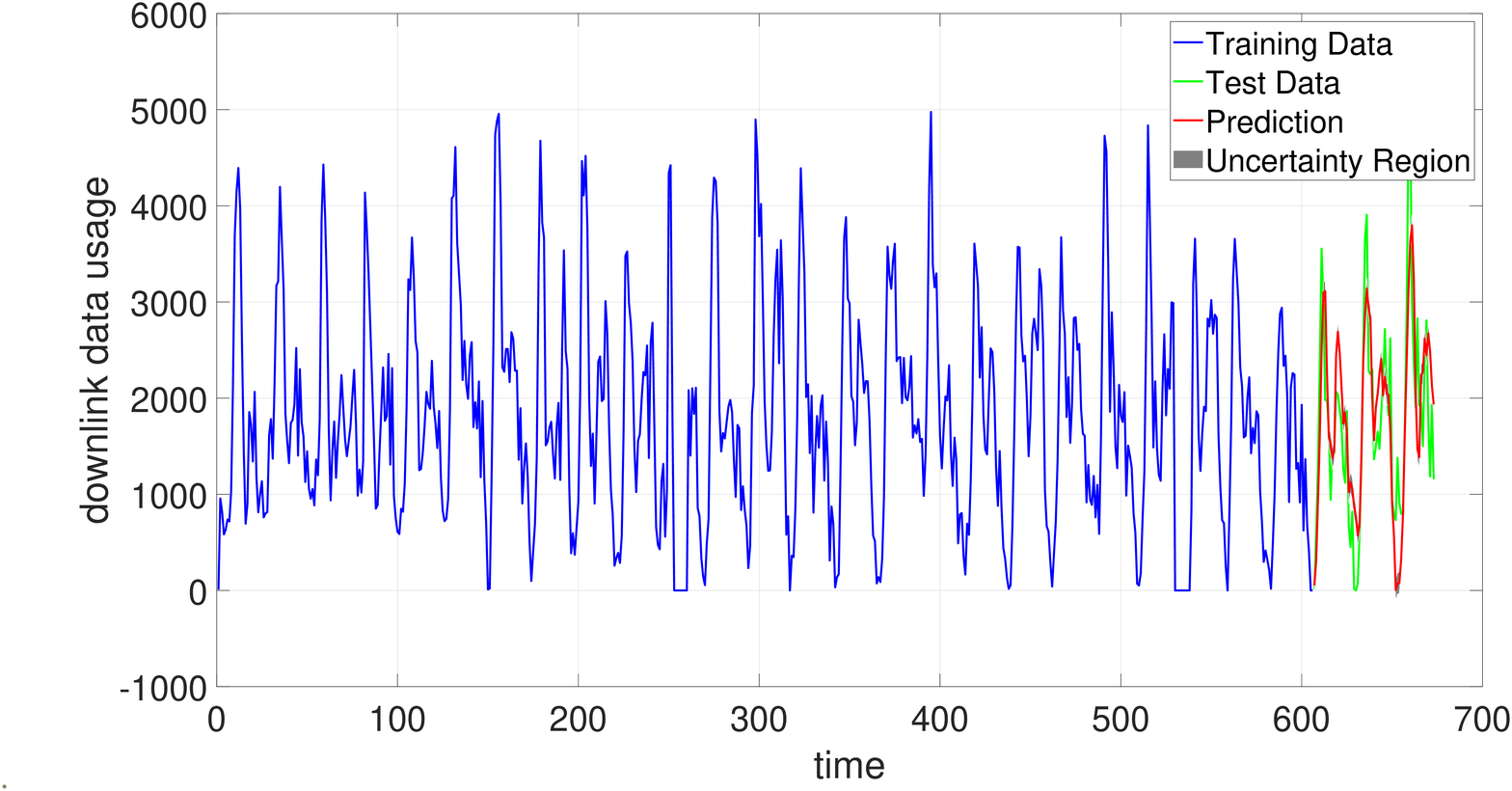}
\vfil
\includegraphics[width=.48\textwidth]{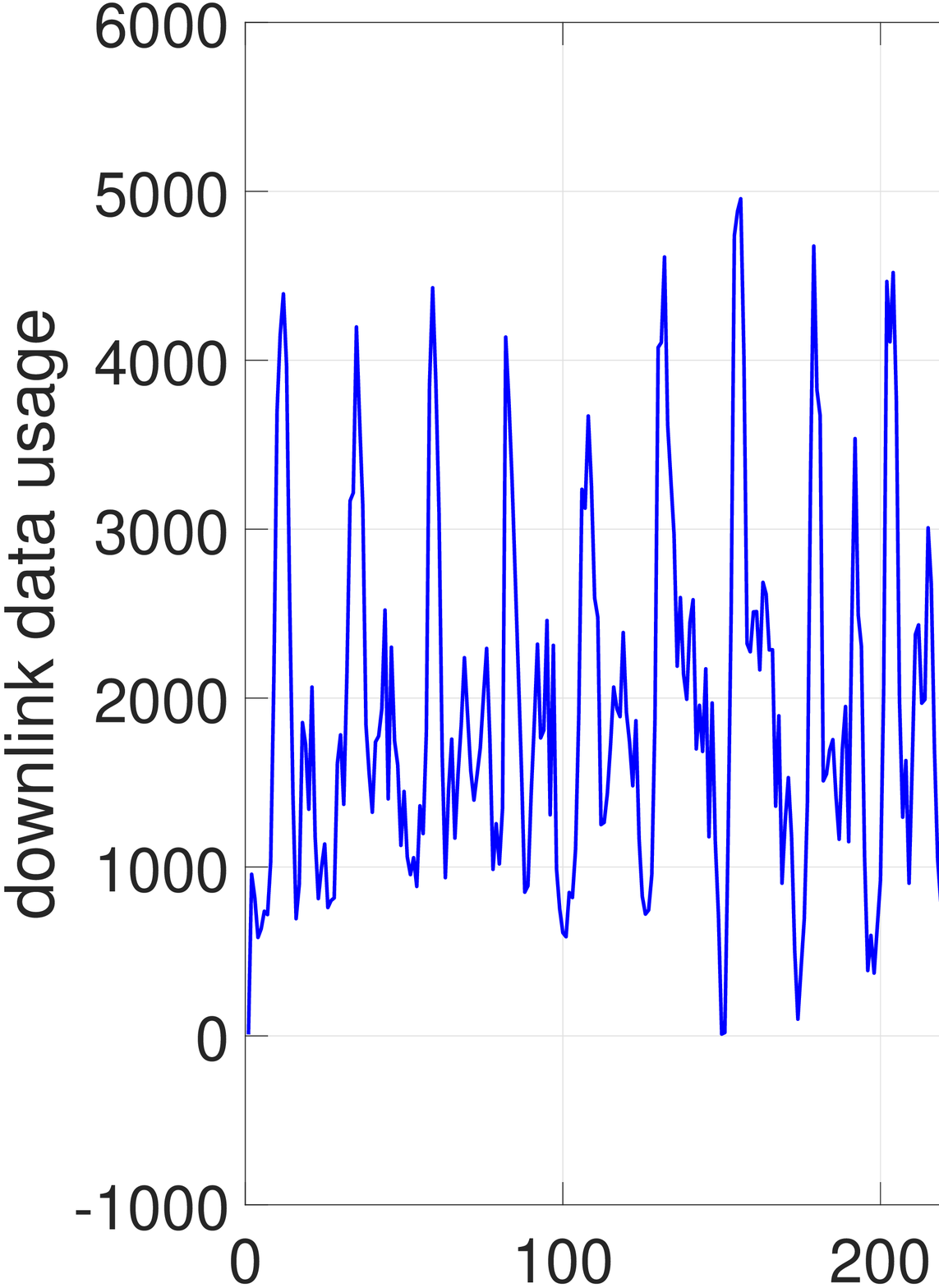}\quad 
\includegraphics[width=.48\textwidth]{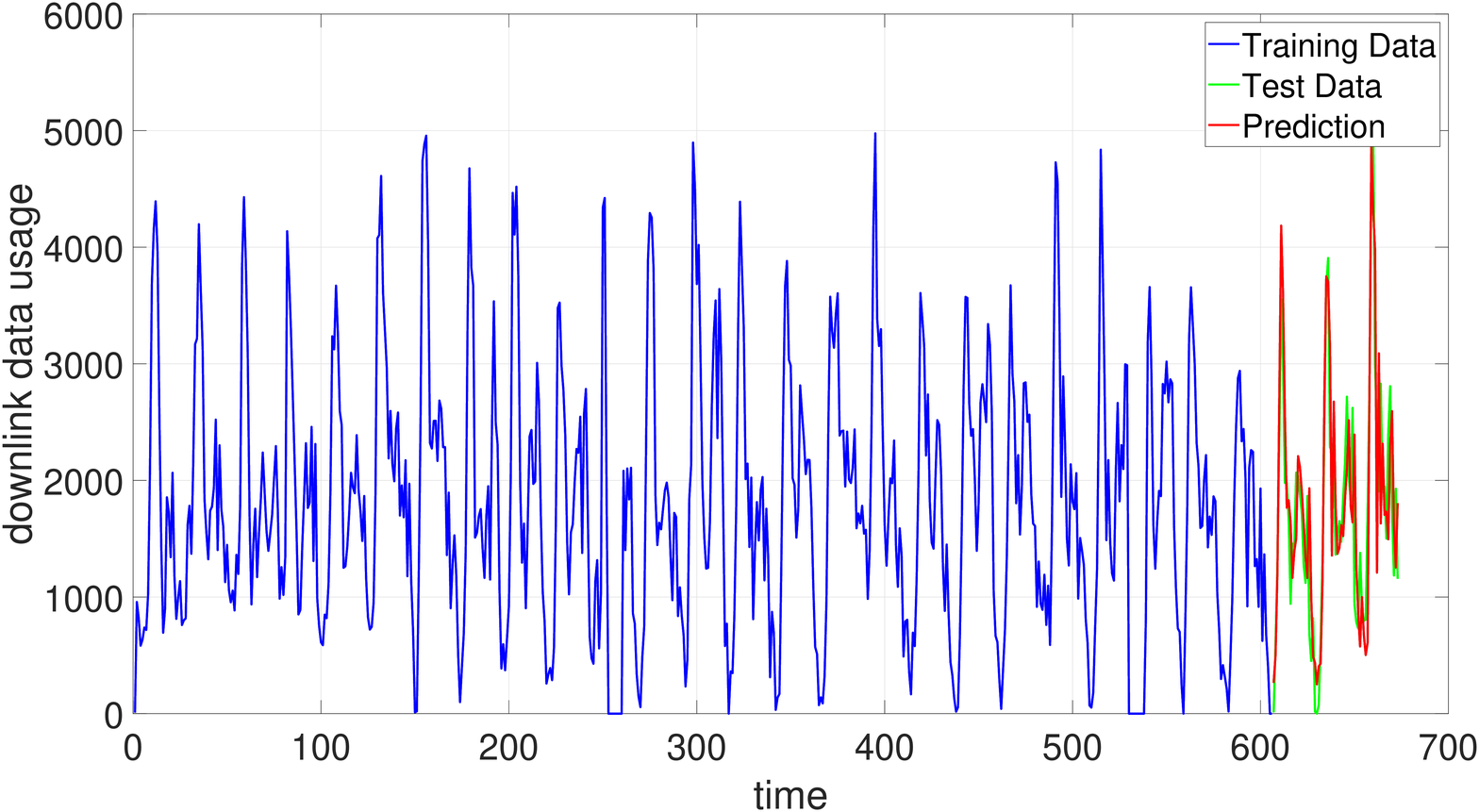}
\caption{Comparison of 5G wireless traffic prediction performance obtained from different models. Top left: GSMGP model with $Q=500$ fixed grids whose $e_{\mathrm{MAPE}}=0.28$; Top right: SMGP model with $Q=500$ modes whose $e_{\mathrm{MAPE}}=0.42$; Bottom left: a standard GP with a hybrid of 3 elementary kernels whose $e_{\mathrm{MAPE}}=0.30$; Bottom right: LSTM model whose $e_{\mathrm{MAPE}}=1.12$. The gray shaded areas represent the uncertainty region (computed as the posterior variances) of the GP models.}
\label{fig:GP-comp-uncertainty}
\end{figure*}

For this 5G wireless traffic prediction example, we constrained ourselves to a GSMGP with $Q=500$, an SMGP with $Q=500$, a standard GP with a hybrid of 3 elementary kernels (two periodic kernels plus an SE kernel) as was used in \cite{Xu19}, and a classic LSTM deep learning model as described above. We used the data collected in the first 607 samples for training the models and used the last 67 samples to test their prediction performance. For comparing their prediction accuracy, we chose the \emph{mean-absolute-percentage-error} (MAPE) measure, which is commonly used for evaluating the wireless traffic prediction error. The MAPE averaged over multiple test points is given by 
\begin{equation}
e_{\mathrm{MAPE}}=\frac{1}{n_{*}} \sum_{i=1}^{n_{*}}\left|\frac{ y_{i} - \hat{y}_{i} }{ y_{i} }\right| \times 100\%.
\end{equation}
The prediction performance of the above learning models is shown in Fig.~\ref{fig:GP-comp-uncertainty}. It is readily seen that the GSMGP gives the best point prediction in terms of the MAPE. Moreover, as we mentioned before, the focus of this example is primarily on the uncertainty quantification. For GPs, the desired uncertainty region can be obtained naturally by computing the posterior variances associated with the test samples. In contrast, the classic LSTM model can only provide point predictions without any uncertainty quantification. \textcolor{black}{A recent technique using the so-called deep ensembles \cite{Lakshminarayanan17} can be applied to quantify the predictive uncertainty of the LSTM model. The common characteristic of these techniques  lies in that one has to train the models for multiple times, using different configurations (such as different initial guesses, step sizes, etc). However, such an approach increases substantially the computational load compared to the Bayesian approach, especially when complex (deterministic) learning models are involved.} When comparing the uncertainty regions of the GP models, we can observe that the one using a mixture of elementary kernels tend to be conservative and show the largest uncertainty region. In contrast, both SMGP and GSMGP provide rather accurate point predictions as well as smaller uncertainty levels. It is noteworthy that SMGP presents less accurate point prediction (using its posterior mean) compared to that of GSMGP, but its uncertainty level is modestly larger that its counterpart. This suggests that, in this case, SMGP is less favorable because wrong decisions of switching on/off the BS are more likely made.

\textcolor{black}{The above fitting results clearly demonstrate the advantages of the sparse spectrum kernel-based GP models; however, the obtained performance depends on the quality of the initialization. In particular, the method is sensitive to the initial guess of the SM kernel. According to our experience, a reliable initial guess can be obtained by fitting a periodogram (namely a nonparametric approximation of the true spectral density) in the frequency domain. We could also combine this strategy with the bootstrap technique to generate a number of candidate initial guesses for avoiding bad local minima. Codes for implementing the GSM kernel-based GP model are online available from \url{https://github.com/Paalis/MATLAB_GSM}.}

%
%
%
%

\subsection{Adversarial Learning via Bayesian Deep Neural Networks}
\label{subsec:BNN-application}
Despite the widespread success of DNNs, recent investigations have revealed their high susceptibility to \textit{adversarial examples}; that is, cleverly crafted examples whose sole purpose is that of \textit{fooling} a considered model into \textit{misclassification}. Adversarial examples can be constructed using various approaches, e.g. FSGM\cite{fgsm} and CW\cite{carlini017}.  A popular and powerful attack is the projected gradient descent (PGD)\cite{madry2017towards} attack. Under this scheme, the adversary is assumed to have access to the objective function of the target model, $L(\boldsymbol{w}, \boldsymbol{x}, \boldsymbol{y})$, where $\boldsymbol{w}$ are the model trainable parameters, $\boldsymbol{x}$ the input, and $\boldsymbol{y}$ the predicted output variables. On this basis, the adversary performs an iterative computation; at each iteration, $t$, the adversary computes 
an (e.g., $\ell_\infty$-bounded) adversarial perturbation of the training set examples $\boldsymbol{x}$, based on a multi-step PGD procedure that reads:
\begin{align}
	\boldsymbol x^{t+1} = \prod_{\boldsymbol x+ \mathcal{S}} (\boldsymbol x^t + a \ \mathrm{sgn}(\nabla_{\boldsymbol x} L(\boldsymbol w, \boldsymbol x, y)))
\end{align}
where $\mathcal{S}$ is the set of allowed perturbations, that is the manipulative power of the adversary, e.g. $\ell_{\infty}$-ball around $\boldsymbol x$; $\mathrm{sgn}(\cdot)$ denotes the sign function that  extracts the sign of a real number.  In this context, even some minor, and many times imperceptible modifications, can successfully ``\textit{attack}'' the model, resulting in severe performance degradation. This \textit{frailness} of DNNs casts serious doubt about their deployment in \emph{safety-critical applications}, e.g., autonomous driving \cite{McAllister17}. 

Drawing upon this vulnerability, significant research effort has been recently devoted towards more reliable and robust DNNs. On this basis, several adversarial attacks and defenses have been proposed in the literature, e.g. adversarial training\cite{madry2017towards, tramer2017ensemble, shrivastava2017learning}. Among these, lies the \textit{stochastic modeling} rationale; its main operating principle is founded upon the introduction of \textit{stochasticity} in the considered architecture, e.g., by randomizing the input data and/or the learning model itself \cite{xie2017mitigating, prakash2018deflecting, dhillon2018stochastic}. Clearly, the Bayesian reasoning, which treats parameters as random entities instead of deterministic values, seeking to infer an appropriate generative process, seems to offer a natural stochastic defense framework towards more adversarially robust networks. It must be emphasized that the Bayesian techniques differ from the more standard randomized ones, which simply rely on the randomization of deterministic variables, in the context of the standard deterministic neural networks. Such techniques can be fairly easily handled and attacked. In contrast, in the Bayesian framework, the whole modeling and learning is built upon statistical arguments and the training involves learning of distributions. 

Therefore, in the following, we focus on a recent application of the Bayesian rationale towards {\it adversarial robustness}. Specifically, we present the novel Bayesian deep network design paradigm proposed in \cite{panousis21a} that yields state-of-the-art performance against powerful gradient-based adversarial attacks, e.g. PGD \cite{madry2017towards}. The key aspect of this method is its doubly stochastic nature stemming from two separate sampling processes relying on Bayesian arguments: a) the sparsity inducing non-parametric link-wise IBP prior introduced in Section~\ref{subsubsec:link_wise_ibp}, and b) a stochastic adaptation of the biologically inspired and competition-based LWTA activation,  as discussed in Sections~\ref{subsubsec:link_wise_ibp} and V-D.

{\color{black} We investigate the potency of LWTA-based networks against adversarial attacks under an Adversarial Training regime; we employ a PGD adversary \cite{madry2017towards}. To this end, we use the well-known WideResNet-34 \cite{wideresnet} architecture, considering three different widen factors: 1, 5, and 10;  note from the definition of the WideResNet-34, the larger the widen factor the larger the network. We focus on the CIFAR-10 dataset and adopt experimental settings similar to \cite{wu2021wider}. We use a batch size of 128 and an initial learning rate of 0.1; we halve the learning rate at every epoch after the $75$-th epoch. We use a single sample  for prediction. All experiments were performed using a single NVIDIA Quadro P6000.
	
For evaluating the robustness of this structure, we initially consider the conventional PGD attack with 20 steps, step size $0.007$ and $\epsilon = 8/255$, which are the two parameters required by the PGD.  In Table \ref{tab:trades_width_1}, we compare the robustness of LWTA-based WideResNet networks against the baseline results of \cite{wu2021wider}. As we observe, the Stochastic LWTA-based networks yield significant improvements in robustness under a traditional PGD attack; they retain extremely high natural accuracy (up to $\approx 13\%$ better), while exhibiting a staggering, up to $\approx 32.6\%$, difference in robust accuracy compared to the \textit{exact same architectures} employing the conventional ReLU-based nonlinearities and trained in the \textit{same fashion}. Natural accuracy refers to the performance based on non-adversarial examples, while robust accuracy refers to the case where the network is tested against adversarial examples.
	\begin{table}
		\caption{Natural and Robust accuracy under a conventional PGD attack with 20 steps and $0.007$ step-size using WideResNet-34 models with different widen factors. We use the same PGD-based Adversarial Training scheme for all models \cite{madry2017towards}.}
		\label{tab:trades_width_1}
		\renewcommand{\arraystretch}{1.1}
		\centering
		\resizebox{0.7\textwidth}{!}{
			\begin{tabular}{ccc|cc}
				\multicolumn{5}{c}{Adversarial Training-PGD}\\
				\hline
				& \multicolumn{2}{c}{Natural Accuracy ($\%$)} & \multicolumn{2}{c}{Robust Accuracy ($\%$)}\\
				\cline{2-3}\cline{4-5}
				Widen Factor & Baseline & Stochastic LWTA & Baseline & Stochastic LWTA \\\hline
				1  & 74.04 & \textbf{87.0} & 49.24 & \textbf{81.87} \\
				5  & 83.95 & \textbf{ 91.88} & 54.36 & \textbf{83.4} \\
				10 & 85.41 & \textbf{92.26} & 55.78 &  \textbf{84.3} \\\hline
			\end{tabular}
		}
	\end{table}

	\begin{table}[h!]
		\caption{Robust Accuracy $(\%)$ comparison under the AutoAttack framework. $\dagger$ denotes models that are trained with additional unlabeled data. The AutoAttack performance corresponds to the final robust accuracy after employing all the attacks in AA. Results directly from the AA leaderboard. }
		\label{tab:sota}
		\centering
		\resizebox{0.65\textwidth}{!}{%
			\renewcommand{\arraystretch}{1.1}
			\begin{tabular}{c|c}
				\hline
				Method & AutoAttack\\\hline
				HE \cite{pang2020} & 53.74\\
				WAR \cite{wu2021wider} & 54.73\\ \hhline{=|=}
				Pre-training \cite{hendrycks2019}$\dagger$ & 54.92\\
				\cite{gowal2021uncovering}$\dagger$ & 65.88\\
				WAR \cite{wu2021wider}$\dagger$ & 61.84\\\hline
				Ours (Stochastic-LWTA/PGD/WideResNet-34-1) & \textbf{74.71}\\
				Ours (Stochastic-LWTA/PGD/WideResNet-34-5) & \textbf{81.22}\\
				Ours (Stochastic-LWTA/PGD/WideResNet-34-10) & \textbf{82.60}\\ \hline
			\end{tabular}
		}
	\end{table}
	
	Further,   to ensure that this approach does not cause the well-known obfuscated gradient problem \cite{AthalyeC018},  stronger parameter-free attacks were adopted using the newly introduced AutoAttack (AA) framework \cite{croce2020reliable}. AA comprises an ensemble of four powerful white-box and black-box attacks, e.g., the commonly employed APGD attack; this is a step-free variant of the standard PGD attack \cite{madry2017towards}, which avoids the complexity and ambiguity of step-size selection. In addition, for the entailed $L_\infty$ attack,  the common $\epsilon = 8/255$ value was used. Thus, in Table \ref{tab:sota}, we compare the LWTA-based networks to several recent state-of-the-art approaches evaluated on AA\footnote{\url{https://github.com/fra31/auto-attack}}. The reported accuracies correspond to the final reported robust accuracy of the methods after sequentially performing all the considered AA attacks.  Once again, we observe that these stochastic and sparse networks yield state-of-the-art (SOTA) robustness against all SOTA methods, with an improvement of $\approx 16.72\%$, even when compared with methods that employ substantial data augmentation to increase robustness, e.g. \cite{gowal2021uncovering}. More results are reported in \cite{panousis21a}.  All results vouch for the potency of Stochastic LWTA networks in adversarial settings. 
	
	Finally, since the newly proposed networks consist of stochastic components, i.e., the competitive random sampling procedure to determine the winner in each LWTA block, the output of the classifier might change at each iteration; this obstructs the attacker from altering the final decision. To counter such randomness in the involved computations, in \cite{croce2020reliable}  the APGD attack is combined with an averaging procedure of 20 computations of the gradient at the same point. This technique is known as Expectation over Transformation (EoT) \cite{AthalyeC018}. Thus,  AA was used jointly with EoT for further performance evaluation of the LWTA-based networks. The corresponding results are presented in Table \ref{tab:eot}. As we observe, all of the considered networks retain state-of-the-art robustness against the powerful AA \& EoT attacks. This conspicuously supports the usefulness of the stochastic LWTA activations towards adversarial robustness. Further explanations on why this  performance is obtained are provided in, e.g., \cite{panousis2019nonparametric}.

	\begin{table}[h!]
		\caption{Robustness against AA combined with 20 iterations of EoT. APGD-DLR corresponds to the APGD attack, using a different loss, i.e., the Difference of Logits Ratio \cite{croce2020reliable}.}
		\label{tab:eot}
		\renewcommand{\arraystretch}{1.1}
		\centering
		\begin{tabular}{ccccc}
			\hline
			Widen Factor & Natural Accuracy & APGD & APGD-DLR\\\hline
			1  & 87.00 & 79.67 & 76.15  \\
			5  & 91.88 & 81.67 & 77.65 \\
			10  & 92.26 & 82.55 & 79.00 \\\hline
		\end{tabular}
	\end{table}

     }

\subsection{Unsupervised Learning via Bayesian Tensor Decompositions}
\label{subsec:TD-applications}
In this subsection, we present some recent advances of Bayesian tensor decompositions in two unsupervised learning applications: \emph{social group clustering} and \emph{image completion}. The first application adopts Bayesian tensor CPD \cite{zhao2015bayesian, cheng2020learning}, while the second one employs Bayesian tensor TTD  \cite{hawkins2020compact, clssp21, xu2020learning}. Exploiting the GSM-based sparsity-promoting prior as introduced in Section~\ref{subsec:SA-tensor} and the effective MF-VI inference as introduced in Section~\ref{subsec:infer_tensor}, the resulting algorithms offer nice features bypassing the need of hyper-parameters tuning and in dealing with overfitting.  

\subsubsection{Bayesian Tensor CPD for Social Group Clustering} In contrast to matrix decompositions, which are not unique in general (unless certain constraints are imposed), tensor CPD is provably unique under mild conditions \cite{sidiropoulos2017tensor}. This appealing property has made CPD an important tool for extracting the underlying signals/patterns from the observed data. The interpretability of CPD model can be further enhanced by incorporating some side information, e.g., non-negativeness \cite{cheng2020learning}, into model learning. Here, using the {\it ENRON E-mail corpus} dataset (a $3$-D tensor  with the size $184 \times 184 \times 44$), we demonstrate how the Bayesian tensor CPD  (with non-negative factor matrices) \cite{cheng2020learning} can be used to simultaneously determine the number of social groups, cluster people into different groups, and extract interpretable temporal profiles of different social groups. 

The considered dataset records the number of E-mail exchanges between $184$ people within $44$ months. In particular, each entry is the number of E-mails exchanged between two people within a certain month. The physical meaning of three tensor dimensions are: the people who sent E-mails, the people who received E-mails, and the months, respectively. After applying Bayesian tensor CPD, the automatically determined tensor rank can be interpreted as the number of underlying social groups. For the first two factor matrices, the physical meaning for each element is that it quantifies the ``score''  that a particular person belongs to a particular E-mail sending and receiving group, respectively. For the third factor matrix, each column corresponds to the temporal profile of the associated social group, see discussion in \cite{cheng2020learning}.
\begin{figure}[!t]
\centering
\includegraphics[width= 6 in]{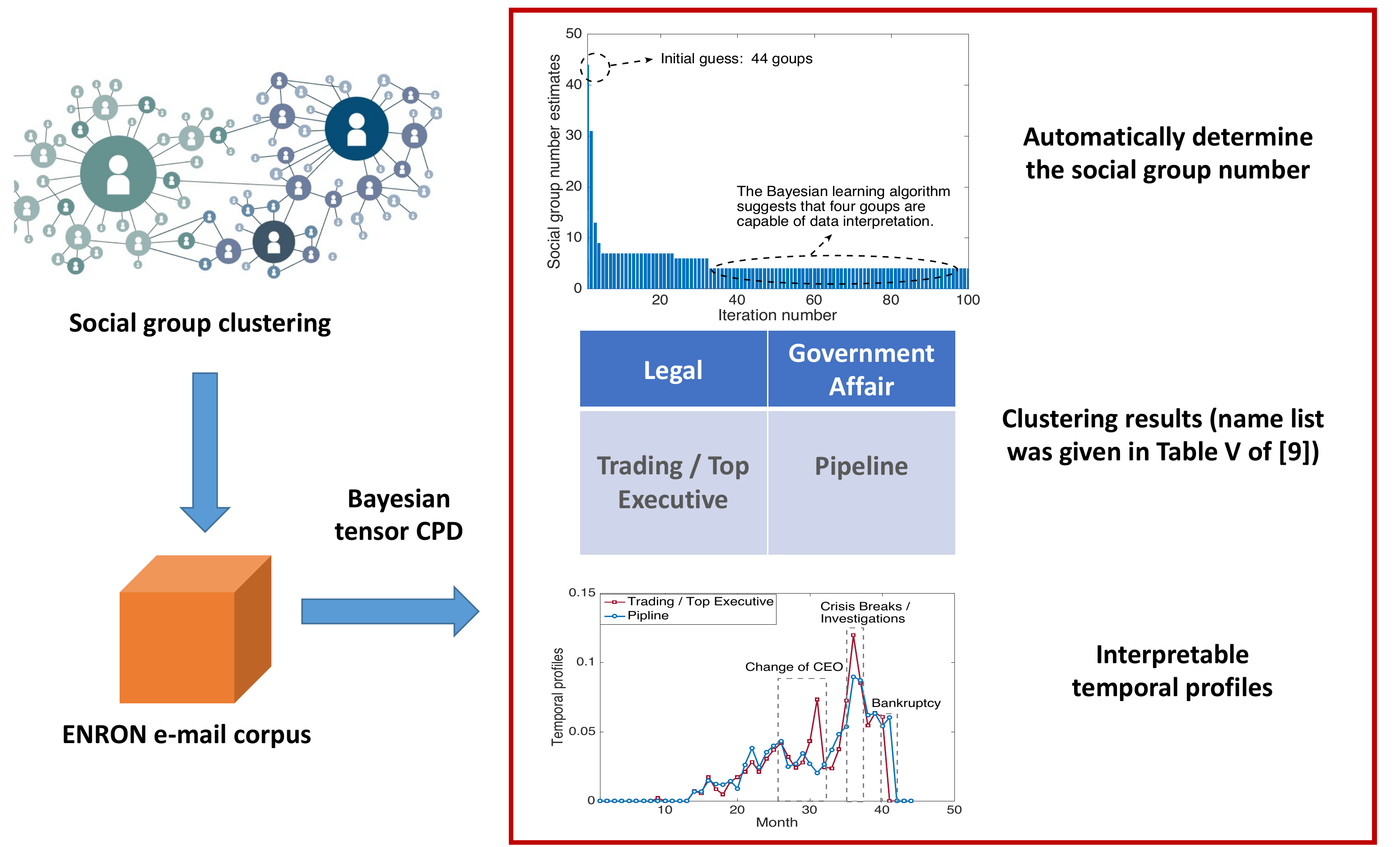}
\caption{Bayesian tensor CPD for social group clustering.}
\label{fig:tensorcpd_clustering}
\end{figure}

Typically, we set the initial number of the social groups (i.e., tensor rank) large, (e.g., the minimal dimension of the tensor data, $44$, in the above example), and then run the Bayesian learning algorithm \cite{cheng2020learning} to automatically determine the number of social groups that best interprets the data. As seen in Fig. \ref{fig:tensorcpd_clustering}, the estimated number of social groups gradually reduces to the value $4$, indicating four underlying social groups. This is consistent with the results published in \cite{Papa13, bader2006temporal}, which are obtained via trial-and-error experiments. The clustering results can be read from the first factor matrix, which is of size $184 \times 4$. Specifically,  for each column, only a few elements have non-zero values, and they can be used to identify the significance of the corresponding people in this social group.  After sorting the scores of each column in the first factor matrix, the people with top $10$ scores in each social group are shown in Table V of  \cite{cheng2020learning}. The clustering results are well interpretable as illustrated in Fig. \ref{fig:tensorcpd_clustering}. For example, the people in the first group work either in legal department or as lawyers, thus are clustered together. Moreover, interesting temporal patterns can be observed from the third factor matrix. It is clear that when the company has important events such as the change of CEO, crisis breaks and bankruptcy, distinct peaks appear.\footnote{This information can be acquired via checking the E-mail content and related research works, e.g.,  \cite{Papa13, bader2006temporal}.}  {\color{black} The E-mail data analysis results have showcased the appealing advantage of Bayesian SAL in the context of tensor CPD, that is, the automatic determination of the social group number. This is important, since it leads to interpretable results on the group member and the temporal profiles can be naturally obtained.}

\subsubsection{Bayesian Tensor TTD for Image Completion} Color images are naturally $3$-D tensor (with two spatial dimensions and one RGB dimension). To fully exploit the inherent structures of images, recent image completion works 
 \cite{ko2020fast, bengua2017efficient, yuan2018high} usually fold an image into a higher dimensional tensor (e.g., $9$-D tensor), and then apply tensor decompositions to recover the missing pixels, among which TTD is one of the most important tools due to its excellent performance. The folding operation is called tensor augmentation, see details in, e.g., \cite{ko2020fast, bengua2017efficient, yuan2018high}.  For a $P$-D tensor, TTD has $P-1$ hyper-parameters (called TT ranks). Manually  tuning different combinations of these hyper-parameters for overfitting avoidance is time-consuming. To facilitate this process, recent advances, see e.g., \cite{hawkins2020compact, clssp21, xu2020learning}, first assume large values for TT-ranks, employ the sparsity-promoting GSM prior, and use variational inference for effective Bayesian SAL. The resulting algorithms can automatically learn the most suitable TT-ranks to match the underlying data pattern \cite{hawkins2020compact, clssp21, xu2020learning}. 
 
\begin{figure*}[!htb]
\centering
\begin{tabular}{@{} m{7 em} m{ 7 em} m{ 7 em} m{7 em} m{ 7 em} @{}}
		

\includegraphics[width=.18\textwidth,keepaspectratio]{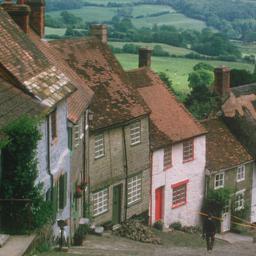}&
  
\includegraphics[width=.18\textwidth,keepaspectratio]{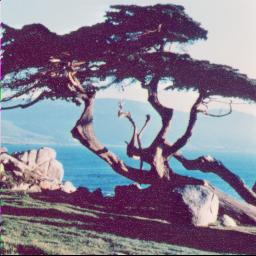}&
    
\includegraphics[width=.18\textwidth,keepaspectratio]{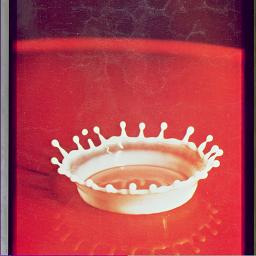}&

\includegraphics[width=.18\textwidth,keepaspectratio]{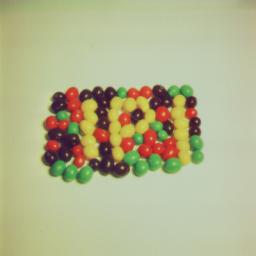}&

\includegraphics[width=.18\textwidth,keepaspectratio]{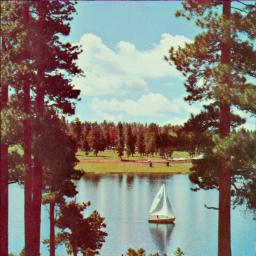} \\
\includegraphics[width=.18\textwidth,keepaspectratio]{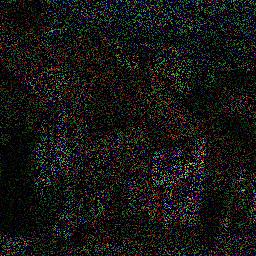}&
  
\includegraphics[width=.18\textwidth,keepaspectratio]{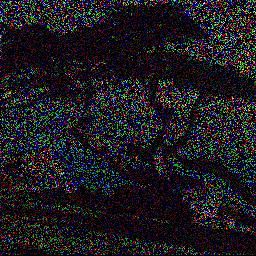}&
    
\includegraphics[width=.18\textwidth,keepaspectratio]{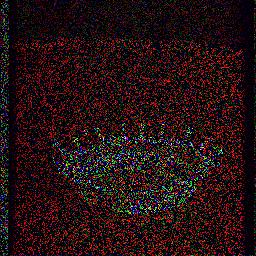}&

\includegraphics[width=.18\textwidth,keepaspectratio]{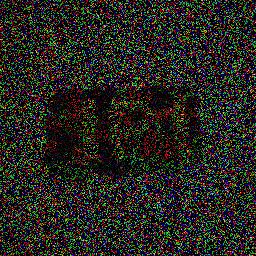}&

\includegraphics[width=.18\textwidth,keepaspectratio]{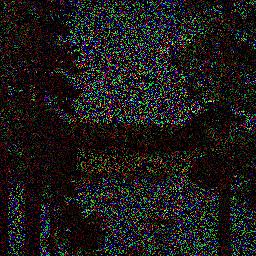} \\
\includegraphics[width=.18\textwidth,keepaspectratio]{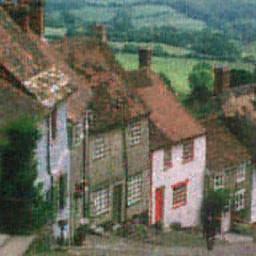}&
  
\includegraphics[width=.18\textwidth,keepaspectratio]{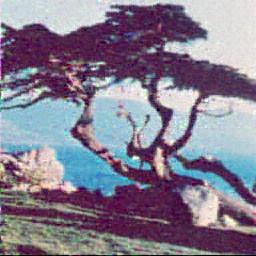}&
    
\includegraphics[width=.18\textwidth,keepaspectratio]{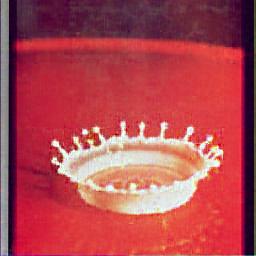}&

\includegraphics[width=.18\textwidth,keepaspectratio]{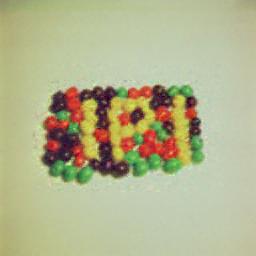}&

\includegraphics[width=.18\textwidth,keepaspectratio]{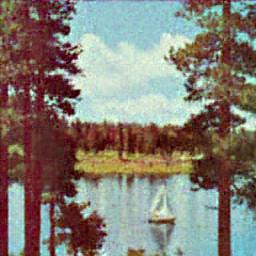} \\

\includegraphics[width=.18\textwidth,keepaspectratio]{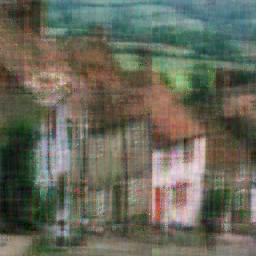}&
  
\includegraphics[width=.18\textwidth,keepaspectratio]{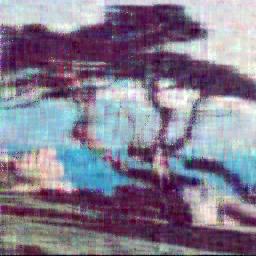}&
    
\includegraphics[width=.18\textwidth,keepaspectratio]{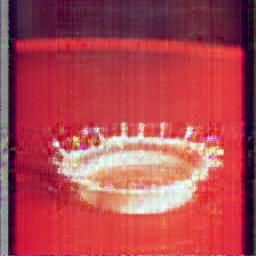}&

\includegraphics[width=.18\textwidth,keepaspectratio]{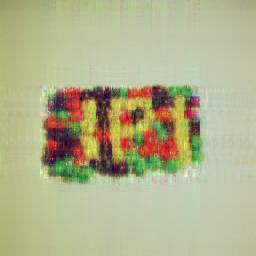}&

\includegraphics[width=.18\textwidth,keepaspectratio]{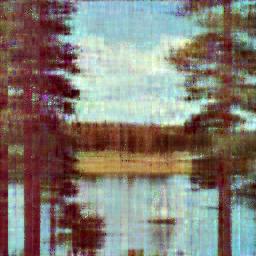} \\


\includegraphics[width=.18\textwidth,keepaspectratio]{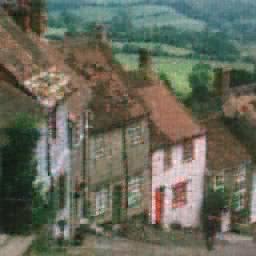}&
  
\includegraphics[width=.18\textwidth,keepaspectratio]{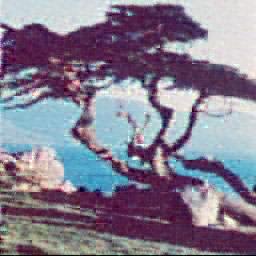}&
    
\includegraphics[width=.18\textwidth,keepaspectratio]{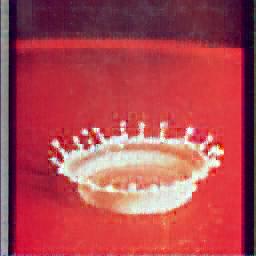}&

\includegraphics[width=.18\textwidth,keepaspectratio]{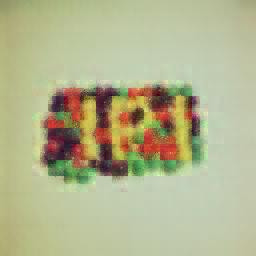}&

\includegraphics[width=.18\textwidth,keepaspectratio]{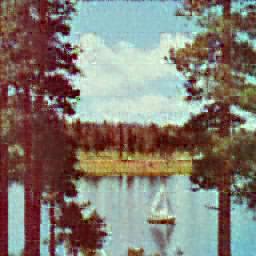} \\  


\includegraphics[width=.18\textwidth,keepaspectratio]{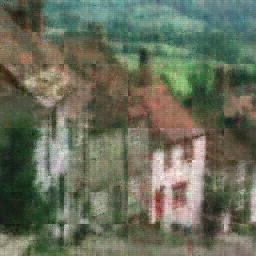}&
  
\includegraphics[width=.18\textwidth,keepaspectratio]{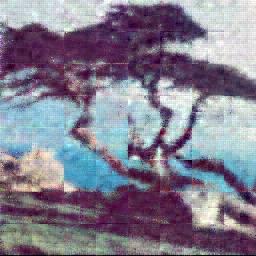}&
    
\includegraphics[width=.18\textwidth,keepaspectratio]{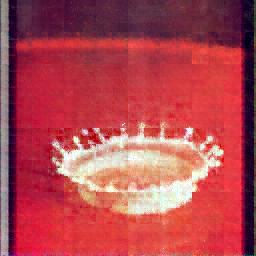}&

\includegraphics[width=.18\textwidth,keepaspectratio]{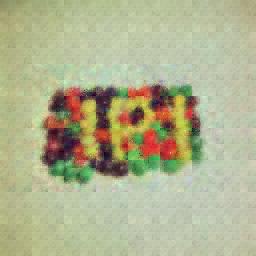}&

\includegraphics[width=.18\textwidth,keepaspectratio]{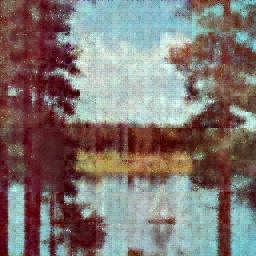} \\  
\end{tabular}
\caption{ Experimental results for visual comparison on image completion with $80\%$ missing data. Ground-truth images are in the top row. The second row includes the images with missing values. The third to the bottom rows include results from Bayesian TTD, TTC-TV,  TMAC-TT, and STTO.}
\label{fig:completionsamples}
\end{figure*}

As an illustration, we consider the image completion of $5$ images. Each image is  with size $256 \times 256 \times 3$, and $80$ percent of its pixels are randomly removed.  After tensor augmentation, they are folded into a $9$-D tensor with size $16\times4\times4\times4\times4\times4\times4\times4\times3$. We assume the initial TT-ranks is set as large as $60$, and apply the Bayesian TTD algorithm \cite{xu2020learning} to complete the missing pixels. In comparison, we present the image completion results from other recent TTD algorithms: TTC-TV \cite{ko2020fast}, TMAC-TT  \cite{bengua2017efficient}, and STTO  \cite{yuan2018high}, with the suggested hyper-parameter settings in their papers. The widely-used metrics, e.g., the peak signal-to-noise ratio (PSNR), were reported in Table II of  \cite{xu2020learning}, from which it can be concluded that the Bayesian TTD algorithm achieves the best overall performance. Particularly, in most cases, the Bayesian TTD algorithm recovers images with $1$-$5$ dB higher PSNR than other algorithms. This is visually evident in the recovered images shown in Fig. \ref{fig:completionsamples}. {\color{black} In this example, the Bayesian SAL-based tensor TTD  \cite{xu2020learning} gets rid of the costly hyper-parameter tuning process for balancing the trade-off between data fitting and the noise overfitting; it directly learns these hyper-parameters from observations and shows excellent image restoration performance.}

{\color{black} Some suggestions are provided on the real implementations of Bayesian tensor decomposition algorithms. a) Initialization: To assist the algorithm to avoid being trapped in a poor local minima, the initial factor matrix is usually set equal to the singular value decomposition approximation of the matrix, which is obtained by unfolding the tensor data along a specific dimension, see e.g., \cite{cheng2022towards, cheng2020learning, xu2020learning, zhao2015bayesian}. b) On-the-fly pruning:  To accelerate the learning process while not affecting the convergence, in each iteration, if some of the columns in the factor matrices are found to be indistinguishable from an all-zeros column vector, they can be safely pruned, see, e.g., the discussion in \cite{cheng2022towards, cheng2020learning}. c) Robustness against strong noise: When the corrupting noise sources are of large power, it was shown in \cite{cheng2022towards} that slowing the noise precision learning can increase the robustness of the algorithm.  Demo codes of Bayesian tensor CPD and TTD algorithms are online available from \url{https://github.com/leicheng-tensor?tab=repositories}.}

\section{Concluding Remarks and Future Research}  
\label{sec:conclusion}
In this article, we have presented an overview of some state-of-the-art sparsity-promoting priors for both Bayesian linear and non-linear modeling, as well as parametric and non-parametric models. In particular, these priors are incorporated into three advanced data analysis tasks, namely, the GP models, Bayesian deep neural networks, and tensor decomposition models, that can be applied to a wide spectrum of signal processing and machine learning applications. Commonly used inference algorithms for estimating the associated hyper-parameters and the (approximate) posterior distributions have also been discussed. 

To demonstrate the effectiveness of the considered advanced sparsity-promoting models, we have carefully selected four important use cases, namely the time series prediction via the GP regression, adversarial learning via Bayesian deep neural networks, social group clustering and image completion using tensor decompositions. The reported results indicate that: a) Sparsity-promoting priors are able to adapt themselves to the given data and enable automatic model structure selection; b) The resulting sparse solution can better reveal the underlying (physical) characteristics of a target system/signal with only a few effective components; c) Sparsity-promoting priors, acting as the counterparts of the regularizers in optimization-based methods, can effectively help to avoid data overfitting, especially when the data size is relatively small; and (4) Sparsity-promoting priors lead to natural and more reasonable uncertainty quantification which is hard to obtain via traditional deep learning models. 

Despite the rapid development of Bayesian learning and the enumerated advantages of sparsity-promoting models, still, such models are confronted with some challenges. Some open research directions are summarized as follows. 
\begin{itemize}
\item \textit{Quality of the posterior/predictive distribution}. As we mentioned before, a unique feature of Bayesian learning models lies in its posterior distribution that can be used to generate a point prediction and meanwhile provide an uncertainty quantification of the point prediction. Various recent works \cite{wilson2020bayesian,Wenzel20,Adlam20} indicate that the quality of the posterior distribution that is derived via the Bayesian deep neural networks and GP models can be significantly improved by using \textit{cold tempering}: 
\begin{equation}
p(\boldsymbol{\theta} | \boldsymbol{X}, \boldsymbol{y}) \propto \left( p(\boldsymbol{y} | \boldsymbol{X}, \boldsymbol{\theta}) \cdot p(\boldsymbol{\theta}) \right)^{1/T}, \quad T < 1.
\end{equation}
Two conjectures lie in the misspecification of the learning model and careless adoption of an inadequate, unintentionally informative prior\cite{Wenzel20}. Deeper analysis of such behavior is highly demanded. It is of great value to verify either analytically or experimentally if adopting the sparsity-promoting priors can help to avoid the use of cold tempering. It is also interesting to investigate the generalization property of the sparsity-promoting Bayesian learning models. 

Another path to improve the quality of posterior/predictive distribution is through designing more effective inference methods. There is a recent trend to integrate the strengths of the variational inference \cite{zhang2018advances} and Monte Carlo sampling \cite{angelino2016patterns} in a principled fashion, see, e.g., \cite{han2020stein}, in order to achieve the best trade-off between the inference accuracy and the computational efficiency. Given multiple inference results, Bayesian deep ensembles, e.g., \cite{wilson2020bayesian}, were also proposed for improved posterior/predictive distribution. It will be interesting to investigate how these recent general-purpose advances can be tailored to the sparsity-aware Bayesian modeling introduced in this article.
\item \textit{More emerging applications in complex systems.} We have witnessed various applications of Bayesian learning models, and they will surely continue to play important role in \emph{large and complex systems}, such as 6G wireless communication systems \cite{Xu19} and autonomous systems \cite{McAllister17}, that are constantly facing rapid changing environments and critical decision making. Sparsity-promoting models are flexible enough to adapt themselves (for instance by nulling irrelevant basis kernels in the GP models) to changing data profile and provide rather reliable uncertainty quantification with small computational expense. 
\item \textit{More and tighter interactions of the three data analysis tools.}  Each of the three data analysis tool (introduced in this article) has already tapped into the design of other  tools, see, e.g., DNN and GP \cite{dai2020interpretable}, GP and tensor \cite{izmailov2018scalable}, DNN and tensor \cite{tjandra2017compressing}, to achieve performance enhancement by borrowing the strengths of other tools. However, many of these works are  not under the framework of Bayesian SAL, and thus do not possess the associated  comparative advantages. It is promising to investigate how to combine the strengths of the three popular models, especially under the Bayesian SAL umbrella, to tackle challenging tasks such as non-linear regression for multi-dimensional and even heterogeneous data with deep kernels. 
\item \textit{Sparsity-awareness in emerging learning paradigms.} Recently, we have witnessed various new paradigms, including, for instance, federated learning, life-long learning, meta learning, etc. We strongly believe that by further encoding sparsity-awareness through Bayesian sparse learning strategies, these emerging learning paradigms can further improve their learning efficiency over the learning models that were introduced in this article. 
\end{itemize}

\bibliographystyle{IEEEtran}
\bibliography{spm_ref_new_0426}

\end{document}